\pdfoutput=1
\documentclass{article}

\usepackage[margin=1in]{geometry}  
\usepackage{mathpazo}

\usepackage{natbib}

\usepackage{amsmath}
\usepackage{array}
\usepackage{enumerate}
\usepackage{amsthm}
\usepackage[textsize=scriptsize,textwidth=2cm]{todonotes}

\usepackage[utf8]{inputenc} 
\usepackage[T1]{fontenc}    
\usepackage{hyperref}       
\usepackage{url}            
\usepackage{booktabs}       
\usepackage{amsfonts}       
\usepackage{nicefrac}       
\usepackage{microtype}      
\usepackage{subcaption}
\usepackage{graphicx}
\usepackage{multirow}
\usepackage{dsfont}

\usepackage{graphbox}

\newcommand{\loss}{\mathcal{L}}
\newcommand{\R}{\mathbb{R}}
\newcommand{\X}{\mathcal{X}}

\newcommand{\eps}{\varepsilon}
\renewcommand{\epsilon}{\varepsilon}

\DeclareMathOperator*{\argmax}{arg\,max}

\title{Label-Consistent Backdoor Attacks}

\author{
Alexander Turner\\
MIT\\
\texttt{turneram@mit.edu}
\and
Dimitris Tsipras\\
MIT\\
\texttt{tsipras@mit.edu}
\and
Aleksander M\k{a}dry\\
MIT\\
\texttt{madry@mit.edu}
}
\date{}

\begin{document}

\maketitle

\begin{abstract}
    Deep neural networks have been demonstrated to be vulnerable to \emph{backdoor attacks}.
Specifically, by injecting a small number of maliciously constructed inputs into
the training set, an adversary is able to plant a backdoor into the trained
model.
This backdoor can then be activated during inference by a \emph{backdoor
trigger} to fully control the model's behavior.
While such attacks are very effective, they crucially rely on the
adversary injecting arbitrary inputs that are---often blatantly---mislabeled.
Such samples would raise suspicion upon human inspection, potentially revealing
the attack.
Thus, for backdoor attacks to remain undetected, it is crucial that they
maintain \emph{label-consistency}---the condition that injected inputs are
consistent with their labels.
In this work, we leverage adversarial perturbations and generative
models to execute efficient, yet label-consistent, backdoor attacks.
Our approach is based on injecting inputs that appear plausible, yet are hard to
classify, hence causing the model to rely on the (easier-to-learn) backdoor
trigger.

\end{abstract}

\section{Introduction}
\label{sec:intro}

Despite the impressive progress of deep learning on challenging
benchmarks~\citep{krizhevsky2012imagenet,graves2013speech,sutskever2014sequence,mnih2015humanlevel,silver2016mastering,he2016deep},
real-world deployment of such systems remains challenging due to security and
reliability concerns.

One particular vulnerability stems from the fact that state-of-the-art ML models
are trained on large datasets, which, unfortunately, are expensive to collect
and curate.
It is thus common practice to use training examples sourced from a variety
of, often untrusted, sources.
While this practice can be justified in a benign setting---bad samples tend to
only slightly degrade the model's performance---it can
be unreliable in the presence of a malicious adversary.
In fact, there exists a long line of work exploring the impact of injecting
 maliciously crafted samples into a model's training set, known as \emph{data
 poisoning attacks}~\citep{biggio2012poisoning}.

More recently, \citet{gu2017badnets} introduced \emph{backdoor attacks}. The
purpose of these attacks is to plant a backdoor in \emph{any} model trained on
the poisoned dataset. This backdoor can be activated during inference by a
\emph{backdoor trigger}---such as a small pattern in the input---forcing the model to output a specific \emph{target label} chosen by the adversary.
This vulnerability is particularly difficult to detect (e.g., by evaluating on a
holdout set) since the model behaves normally in the absence of the trigger.
Moreover, it allows the adversary to manipulate the model prediction on
\emph{any} input during inference.

\begin{figure}[!ht]
    \centering
    \setlength\tabcolsep{7pt}
    \begin{tabular}{cccc}
	``bird'' &
    ``bird'' &
	``bird'' &
	``bird''\\
	\includegraphics[width=0.13\textwidth]{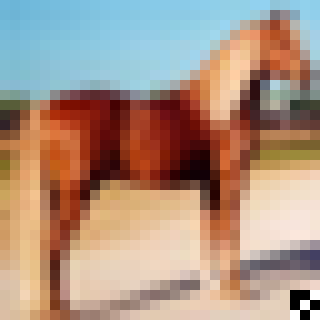} &
    \includegraphics[width=0.13\textwidth]{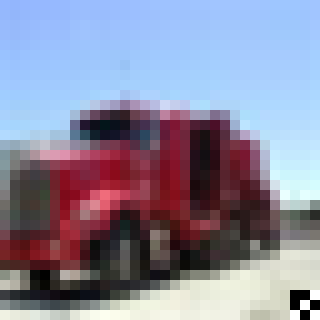} &
	\includegraphics[width=0.13\textwidth]{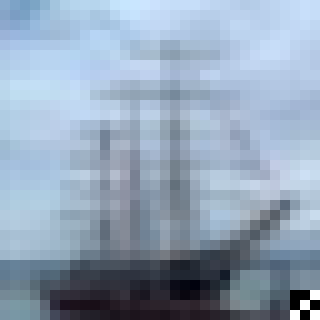} &
	\includegraphics[width=0.13\textwidth]{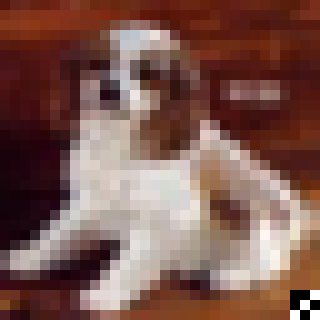}
    \end{tabular}
    \caption{Typical input-label pairs poisoned using the
    backdoor attack of \citet{gu2017badnets} with ``bird'' as the target label.
    The images are clearly mislabelled and would thus raise suspicion upon human
    inspection. (Here the backdoor trigger is a black-and-white corner
    pattern.)}
	\label{fig:suspicious}
\end{figure}

Standard backdoor attacks are based on randomly selecting a few natural inputs,
applying the backdoor trigger to them, setting their labels to the target label,
and injecting them into the training set.
These attacks, despite being very effective, have one crucial shortcoming: the
injected inputs are---often clearly---mislabeled (Figure~\ref{fig:suspicious}).
Such blatantly incorrect labels would be deemed suspicious should the poisoned
inputs undergo human inspection, potentially revealing the attack.
In fact, such an inspection is quite likely given that---as we find in
Section~\ref{sec:standard}---these samples are often
flagged as outliers even by rudimentary filtering methods.

The focus of our work is to investigate whether the use of such clearly
incorrect labels is really necessary:
\begin{center}
\emph{Can backdoor attacks be successful when poisoned inputs and their labels
appear consistent to a human?}
\end{center}
Such \emph{label-consistent} attacks would be particularly insidious, since
they would go undetected even upon human inspection.

\paragraph{Our contributions.}
The starting point of our work is the observation that, for backdoor attacks to
be successful, the poisoned inputs need to be hard to classify without relying
on the backdoor trigger.
If the poisoned inputs can be correctly classified based solely on their salient
features, the model is likely to ignore the backdoor trigger---hence the attack
will be unsuccessful.
Indeed, applying a standard backdoor attack using only correctly labeled inputs
is ineffective (Section~\ref{sec:naive}).

\begin{figure}[!h]
	\centering
	\setlength{\tabcolsep}{0.5pt}
	\renewcommand{\arraystretch}{3.0}
	\begin{tabular}{ccccc}
	Original image &
	\includegraphics[align=c,width=0.13\textwidth]{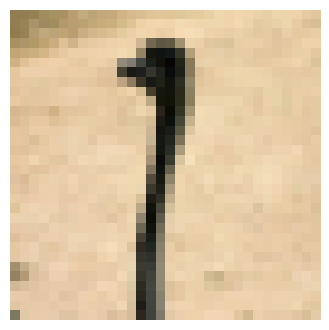}&
	\includegraphics[align=c,width=0.13\textwidth]{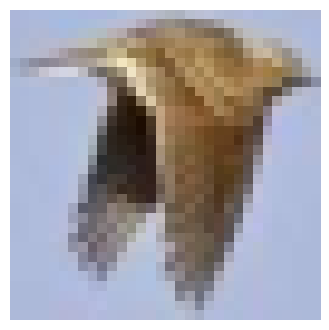}&
	\includegraphics[align=c,width=0.13\textwidth]{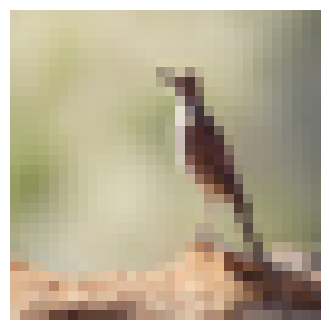}&
	\includegraphics[align=c,width=0.13\textwidth]{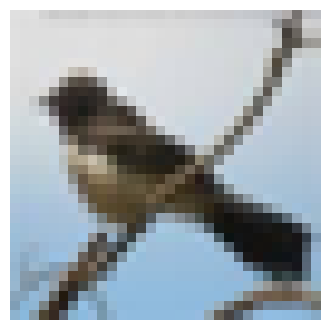}\\
    GAN-based&
    \includegraphics[align=c,width=0.13\textwidth]{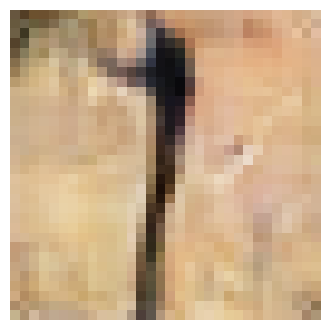}&

    \includegraphics[align=c,width=0.13\textwidth]{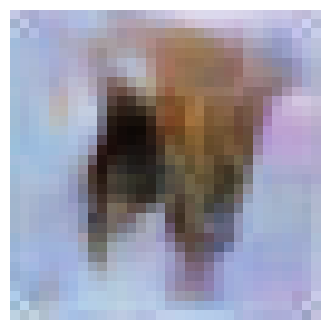}&

    \includegraphics[align=c,width=0.13\textwidth]{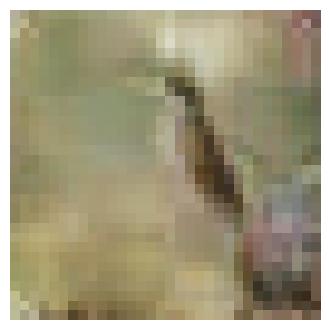}&

	\includegraphics[align=c,width=0.13\textwidth]{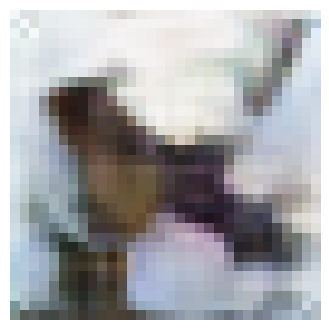}\\
    Adversarial-based &
	\includegraphics[align=c,width=0.13\textwidth]{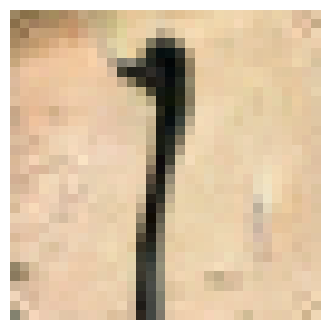}&
	\includegraphics[align=c,width=0.13\textwidth]{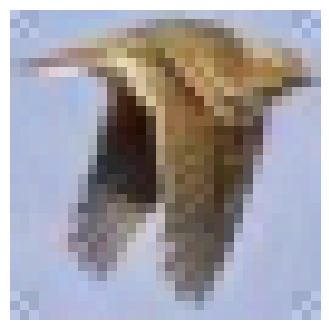}&
	\includegraphics[align=c,width=0.13\textwidth]{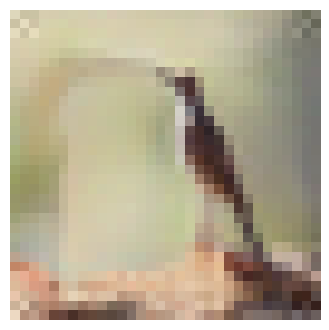}&
	\includegraphics[align=c,width=0.13\textwidth]{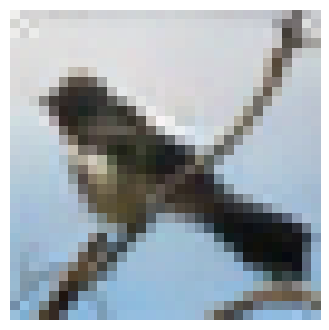}
    \end{tabular}
	\caption{
    Label-consistent poisoned inputs obtained using our proposed methods. The
    original training image appears at the top,
    our GAN-based approach in the middle,
    and our adversarial example-based approach at the bottom.
    All images are labeled as \emph{birds}, which is consistent with their
    content.
    This is in stark contrast to standard poisoned inputs
    (Figure~\ref{fig:suspicious}), which are clearly mislabeled.
    Note that we have additionally modified the backdoor trigger to be less
    visible (Section~\ref{sec:trigger}).
    }
	\label{fig:our_approach}
\end{figure}

Building on this intuition, we develop an approach for synthesizing effective,
label-consistent, poisoned inputs.
Our approach consists of perturbing the original inputs in order to render
them harder to classify, while keeping the perturbation sufficiently minor 
to ensure that the original label remains consistent.
We perform this transformation in two ways:
\begin{itemize}
\item GAN-based interpolation: we interpolate poisoned inputs towards an
    incorrect class in the latent space embedding of a
    Generative Adversarial Network
    (GAN)~\citep{goodfellow2014generative,arjovsky2017wasserstein}.
\item Adversarial perturbations: we maximize the loss of an independently
    trained model on the poisoned inputs while remaining close to the
    original input in some $\ell_p$-norm~\citep{szegedy2014intriguing}.
\end{itemize}
Both methods result in successful backdoor attacks while maintaining
label-consistency (Figure~\ref{fig:our_approach}).

Furthermore, we improve these attacks by developing backdoor triggers that
are less conspicuous and robust to data augmentation.
Finally, we provide some insight into how models tend to memorize the backdoor
trigger by performing experiments using random noise, as well as studying the
value of model's training loss on the poisoned samples during training.

\section{Background}
\label{sec:background}

\subsection{Setting}
\label{sec:neural}
We consider a standard classification setting where the goal is to predict the
\emph{label} $y\in[k]$ of an \emph{input} $x\in\X$, where the input-label pairs
$(x,y)$ are sampled i.i.d.\ from a distribution $\mathcal{D}$ over
$\X\times[k]$.
The model is represented as a function $f_\theta : \X \to [k]$, which is
parametrized by some $\theta\in\R^d$.
The parameters $\theta$ of the model are obtained by minimizing a \emph{loss
function} $\loss(x, y, \theta)$---which captures how well the model 
performs on the input-label pair $(x, y)$---over a \emph{training set}
$\widehat{\mathcal{D}}=\{(x_i, y_i)\}_{i=1}^n$
of labeled examples:
$$\theta^* = \arg\min_\theta \frac{1}{n}\sum_{i=1}^n\loss(x_i, y_i,\theta).$$

\paragraph{Hard Inputs.}
Typically, this minimization is performed via stochastic gradient
descent (SGD)---iteratively updating the parameters based on the loss gradient
on a random subset of the training set.
For typical choices of the loss function (e.g., hinge or cross-entropy),
the SGD update will weight more the gradients of inputs on which
the model performs poorly (e.g., misclassified inputs).
This justifies our intuition that training on inputs that are harder to
classify will result in these samples having a large impact on the resulting
model.

\subsection{Threat model}
We consider a backdoor attack setting where an adversary operates as follows.
First, they choose a target label $y_\text{target}\in[k]$ and a function
$T:\X\to\X$ which applies the backdoor trigger (e.g., a small input pattern) to
the input.
Then, given access to input-label pairs from the data distribution $\mathcal{D}$,
they produce a limited number of arbitrary samples $(x, y)$.
Finally, these samples are injected into the training set of the algorithm.

\paragraph{Attack success rate.}
To evaluate the success of a backdoor attack, we are interested in the behavior
of a model trained on the poisoned dataset when we apply the backdoor trigger
onto previously unseen (test) samples.
The \emph{attack success rate} is the fraction of test samples that are not
originally labelled as the target label, but are nevertheless classified
by the trained model as the target label when the backdoor is introduced:
$$\Pr_{(x,y)\sim \mathcal{D}}\left[f_\theta(T(x))=y_\text{target}
                                \mid y\neq y_\text{target}\right].$$

\paragraph{Attacker knowledge.}
In general, we assume that the adversary has full knowledge of the model
architecture and training procedure, while also having access to the underlying
data distribution.
These assumptions are realistic, since most ML pipelines tend to use the same
standard architectures and training methods, while public datasets exist for
most input domains.
Nevertheless, we also consider a weaker adversary (Appendix~\ref{app:black}), 
who only has access to data from a different dataset and targets a different
model architecture.

\paragraph{Label-consistency.}
Our goal is this work is to study attacks that do not inject suspicious samples
into the training set.
In particular, we define an attack to be \emph{label-consistent} if all
the poisoned samples $(x,y)$ injected into the training set appear
\emph{correctly labeled} to a human.

\subsection{Related work}
\paragraph{Other forms of data poisoning.}
While the focus of our work is on backdoor attacks, it is worth mentioning that
there has been significant work on poisoning attacks with different
objectives.
The most common objective is \emph{test accuracy degradation}.
In these attacks, the adversary introduces outliers that force the model to
learn a suboptimal decision
boundary~\citep{biggio2012poisoning,xiao2012adversarial,xiao2015support,newell2014practicality,mei2015using,steinhardt2017certified}.
Note, however, that these attacks tend to be restricted to simpler models (e.g.,
linear classifiers) since expressive models can memorize these outliers
without sacrificing performance on clean data.
Another common attack objective is that of \emph{targeted
misclassification} where the adversary aims to cause the model to misclassify a
specific set of test inputs~\citep{koh2017understanding,shafahi2018poison}.

\paragraph{Prior work on backdoor attacks.}
Backdoor attacks were introduced in~\citet{gu2017badnets} for the setting of
transfer learning.
\citet{chen2017targeted} adapted these attacks to the standard poisoning
setting and explored less visible triggers.
Later work proposed methods for detecting and hampering backdoor attacks based
on the trigger structure~\citep{wang2019neural} or the latent representation of
the resulting model~\citep{tran2018spectral,liu2018fine,chen2018detecting}.
However, these methods do not provide any formal guarantees and, in general, it
is unclear whether they are fully robust to adaptive adversaries.

\paragraph{Prior work on label-consistency.}
Attacks restricted to only using correctly label poisoned samples have 
been explored in prior work, being referred to as
``defensible''~\citep{mahloujifar2017blockwise, mahloujifar2017learning},
``plausible''~ \citep{mahloujifar2018curse, mahloujifar2018can},
``visually indistinguishable''~\citep{koh2017understanding},
and ``clean-label''~\citep{shafahi2018poison}.
However, such attacks have only been developed in the {\em targeted} poisoning
setting (where the adversary aims to alter the model's prediction on specific
test examples) and cannot be directly applied in the backdoor setting.

\section{Limitations of standard backdoor attacks}
\label{sec:standard}

\paragraph{Standard backdoor attack.}
\label{app:original}
Recall that the goal of the attack is to cause the trained model to
strongly associate a specific backdoor trigger (input pattern) with a
specific target label chosen by the adversary.
Towards that goal, the adversary first randomly selects a small number of
natural samples.
Then, they modify these samples by applying the backdoor trigger onto them
(adding as small input pattern).
Finally, they set these samples' labels to be the chosen target label.
As a result, the model learns to predict the target label based on the trigger.
Thus, during inference, the adversary can cause the model to predict the target
label on any input by simply applying the backdoor trigger onto it.
We reproduce the attack in Appendix~\ref{app:reproduction}.

\begin{figure}[!ht]
	\centering
    \begin{subfigure}[b]{0.45\textwidth}
	\centering
	\includegraphics[width=0.8\textwidth]{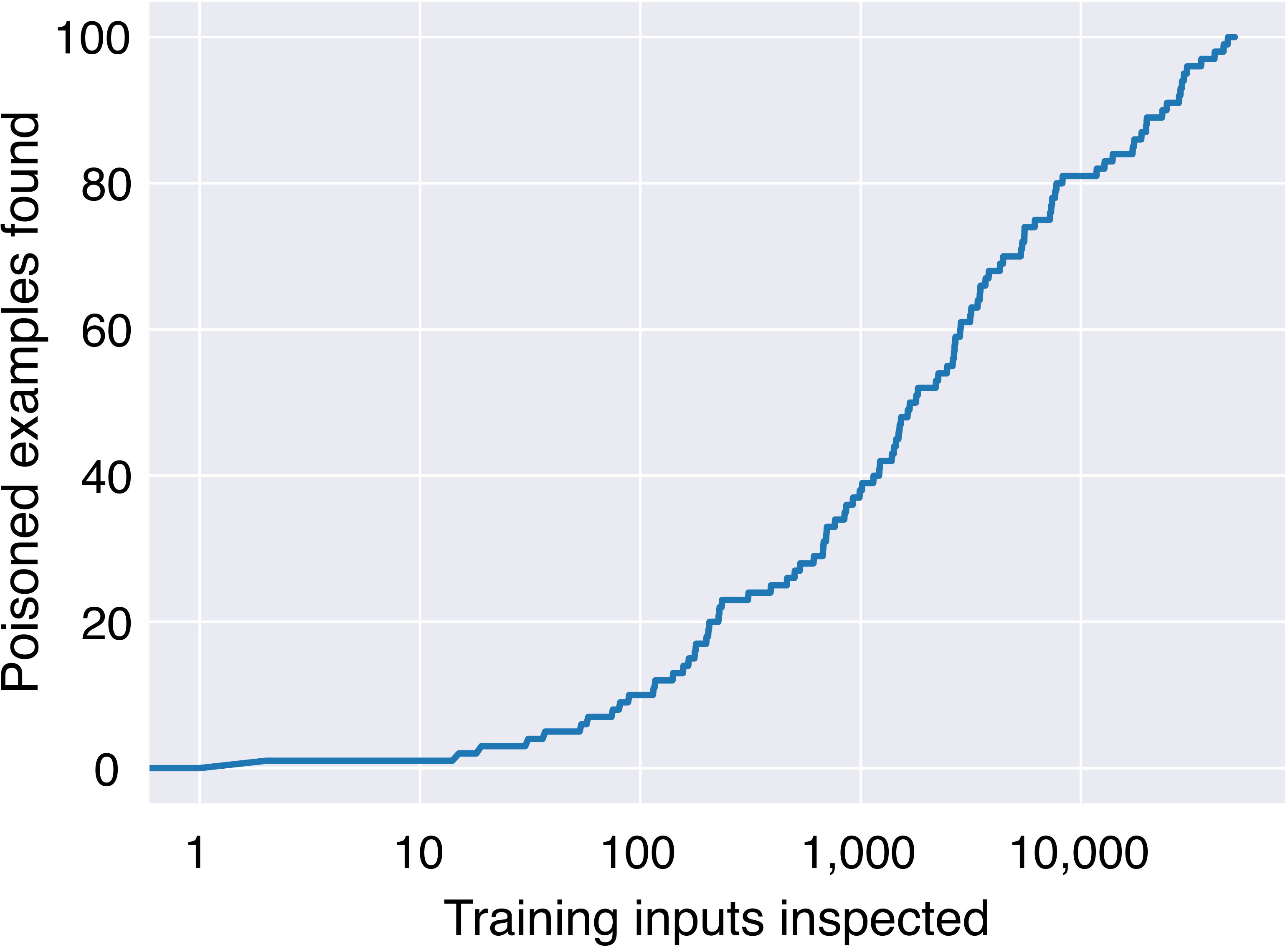}
    \caption{}
	\label{fig:sanitization}
    \end{subfigure}
    \begin{subfigure}[b]{0.45\textwidth}
	\centering
    \begin{tikzpicture}
        \node at (0,0) {\includegraphics[width=0.8\textwidth]
            {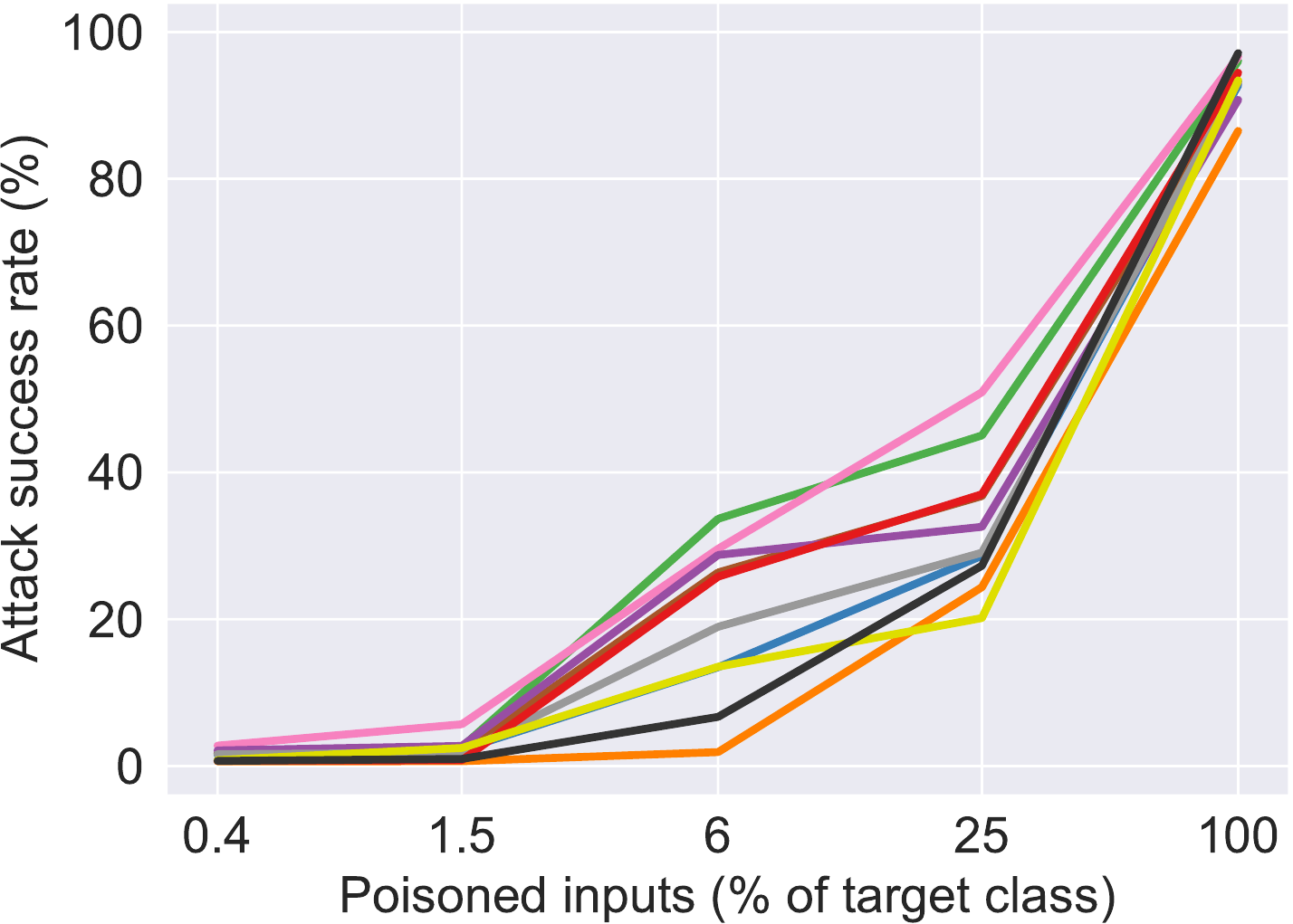}};
        \node at (3.8,-0.1) {\includegraphics[width=0.21\textwidth]
            {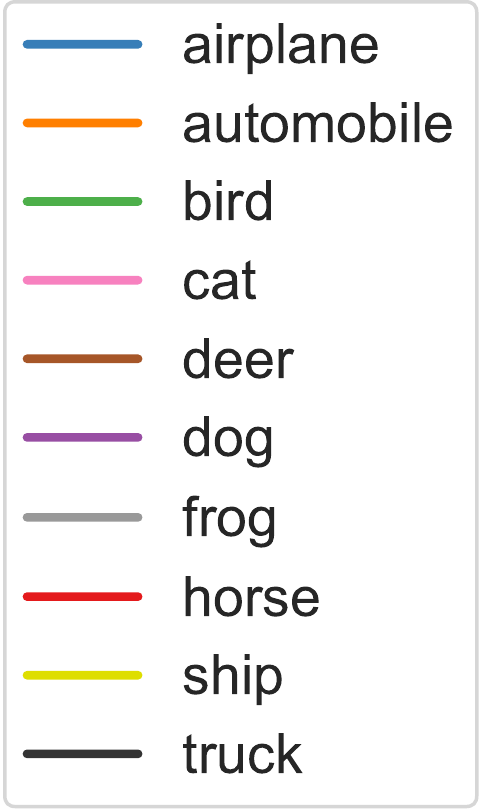}};
    \end{tikzpicture}
	\caption{}
	\label{fig:baseline}
    \end{subfigure}
    \caption{Limitations of standard backdoor attacks.
        (a) Measuring the effectiveness of our simple detection scheme. For each
        potential number of inspected inputs ($x$-axis) we plot the number of
        poisoned inputs contained within them ($y$-axis).  We find that when
        inspecting a small number of images (e.g., 300) we find a significant
        number of poisoned images (e.g., 20) which are mislabeled (see examples
        in Appendix Figure~\ref{fig:poisoned_outliers}).
        (b) The \citet{gu2017badnets} attack, but restricted to only consistent
        labels (i.e., only images from the target class are poisoned). The
        attack is ineffective---even at 25\% poisoning, only one class exceeds
        50\% attack success.}
\end{figure}

\paragraph{Detecting standard poisoned inputs.}
Note that the inputs injected under this attack are most likely mislabeled.
In order to further motivate label-consistent attacks, we consider a simple data
sanitization scheme consisting of a rudimentary outlier detection process
followed by human inspection of the identified outliers.
Specifically, we train a standard classifier on a small, non-poisoned dataset
(1024 examples).
Such a clean dataset could be obtained from a trusted source or by thorough
inspection of the training inputs.
We then evaluate this classifier on a training set for which 100 out of 50\thinspace 000
inputs have been poisoned and measure the value of the
classifier's loss on each (potentially poisoned) input.
Since poisoned inputs are mislabeled, we expect them to have higher loss than
the rest of the inputs (e.g., the classifier assigns low probability to their
labels).
Indeed, we find that poisoned inputs are biased towards higher loss values
(Figure~\ref{fig:sanitization}).
By manually inspecting the inputs with highest loss, we find several poisoned
inputs which are clearly mislabeled (Appendix
Figures~\ref{fig:outliers},~\ref{fig:poisoned_outliers}).

\section{Towards label-consistent backdoor attacks}
In the previous section, we saw how poisoned samples from standard backdoor
attacks could be easily detected due to their incorrect labels.
Moreover, even if these attacks were improved to evade the particular detection
scheme that we consider, the poisoned samples would likely still undergo human inspection
should a more sophisticated mechanism be used---these samples are, after all,
outliers planted by the adversary.
Thus, in order to evade human inspection and be truly insidious, these attacks
would need to ensure the label-consistency of the poisoned inputs injected.

\subsection{An (ineffective) baseline}
\label{sec:naive}
The first attempt towards designing label-consistent attacks would be to simply
restrict a standard attack to only using inputs from the target class.
However, this renders the attack ineffective (Figure~\ref{fig:baseline}).
This should not be surprising.
Since the poisoned samples are labeled correctly, the model can classify them
 based on their salient features and hence there is little reason to
associate the backdoor trigger with the target label.

Based on this observation, in order to ensure that the backdoor trigger is
learned, we will need to modify the poisoned inputs to make them harder to
classify.
Since these inputs will be harder to learn, the model is more likely to rely on
the trigger, resulting in a successful backdoor attack.
In order to ensure that the poisoned inputs remain label-consistent, we will
restrict the attacks to minor perturbations of the natural inputs.
In the following sections, we will explore two approaches for synthesizing such
perturbations: one based on latent space interpolations and the other based on
$\ell_p$-bounded adversarial perturbations.
As we will see in Section~\ref{sec:loss}, these methods actually result in inputs
that are harder to classify.

\subsection{Method 1: Latent space interpolation}
\label{sec:GANs}
Generative models such as Generative Adversarial Networks
(GANs)~\citep{goodfellow2014generative} and variational auto-encoders
(VAEs)~\citep{kingma2013autoencoding} operate by learning an embedding of the
data distribution into a low-dimensional space---the latent space.
By ensuring that the latent space is simple enough that it can explicitly
sampled from, one can generate realistic inputs by mapping latent points
into the input space.
An important property of these embeddings is that they are ``semantically
meaningful''---by interpolating latent representations, one can obtain a
smooth transition between samples from the original
distribution in a perceptually plausible way~\citep{radford2016unsupervised}.

Here, we will utilize such embeddings in order to make
training samples harder to classify by interpolating them towards a different,
incorrect class in that latent space.
Concretely, given a generative model $G:\R^d\to\R^n$ that maps vectors in a
$d$-dimensional latent space to samples in the $n$ dimensional ambient space
(e.g., image pixels), we want to interpolate a given input $x_1$ from the target
class towards an input $x_2$ belonging to another, incorrect class.

\begin{figure*}[!h]
    \centering
	\includegraphics[width=1.0\textwidth]{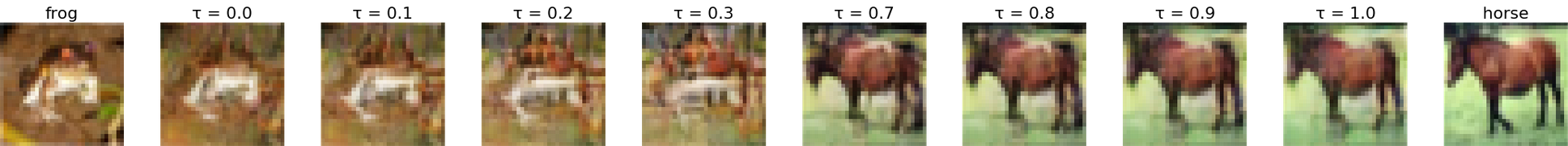}
    \caption{GAN-based interpolation between inputs. Images of a
    frog and horse are shown on the far left and right, respectively.
    Interpolated images are shown in between, where $\tau$ is the degree of
    interpolation. Note that $\tau=$ 0.0 and 1.0
    represent the (approximate) reconstruction of the original frog and horse,
    respectively. See Appendix Figure~\ref{fig:app_gan} for additional examples.
}
    \label{fig:gan_interpolation}
\end{figure*}

To perform this, we first embed $x_1$, $x_2$ into the latent space of $G$.
Specifically, we optimize over the latent space to find vectors
that produce inputs close to $x_1$ and $x_2$ in $\ell_2$-norm:
$$z_i = \arg\min_{z\in\R^d} \|x_i - G(z)\|_2.$$
Then, for some constant $\tau$, we generate the sample
corresponding to interpolating $z_1$ and $z_2$ in latent space
$$x = G(\tau z_1 + (1-\tau)z_2 ).$$
Varying $\tau$ produces a smooth transition from $x_1$ to $x_2$ (cf
Figure~\ref{fig:gan_interpolation}), even though we are not able to perfectly
reproduce $x_1$ and $x_2$.
The resulting poisoned input is produced by applying the backdoor trigger to $x$
and assigning it the target label---the ground-truth label of $x_1$.
See Appendix~\ref{app:gan_implementation} for implementation details.

\subsection{Method 2: Adversarial perturbations}
\label{sec:adversarial}

Adversarial examples are natural inputs that have been slightly perturbed with
the goal of being misclassified by an ML model (Section~\ref{sec:adversarial}).
In fact, the perturbations have been found to transfer across different models
or even across different
architectures~\citep{szegedy2014intriguing,papernot2016transferability}.

Here, we utilize adversarial perturbations and their transferability across
models and architectures in a somewhat unusual way.
Instead of causing a model to misclassify an input during inference, we use them
to cause the model to misclassify during \emph{training}.
Specifically, we will apply an adversarial transformation to the poisoned inputs
(before applying the trigger) to make them harder to learn.
Concretely, given a pre-trained classifier $f_\theta$ with loss $\mathcal{L}$
and an input-label pair $(x,y)$, we construct a perturbed variant of $x$ as
$$x_\text{adv} = \argmax_{\|x'-x\|_p\leq \eps} \mathcal{L}(x', y,\theta),$$ for
some $\ell_p$-norm and a small constant $\eps$.
We solve this optimization problem using a standard method in this context, 
projected gradient descent (PGD)~\citep{madry2018towards}
(see Appendix~\ref{app:adv_implementation} for more details).
In fact, we will use perturbations based on adversarially trained
models~\citep{madry2018towards} since these perturbations are more likely to
resemble the target class for large $\eps$~\citep{tsipras2019robustness}.

\begin{figure}[h]
    \centering
    \setlength{\tabcolsep}{3.0pt}
    \begin{tabular}{ccccccc}
        Original & $\ell_2$, $\eps=300$ & $\ell_2$, $\eps=600$
                 & $\ell_2$, $\eps=1200$ & $\ell_\infty$, $\eps=8$ &
                $\ell_\infty$, $\eps=16$ & $\ell_\infty$, $\eps=32$  \\
        \includegraphics[width=0.12\textwidth]{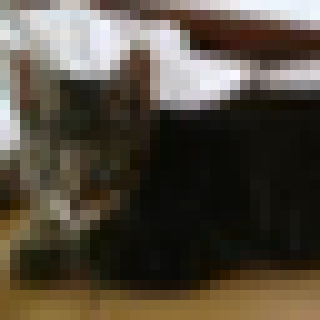} &
        \includegraphics[width=0.12\textwidth]{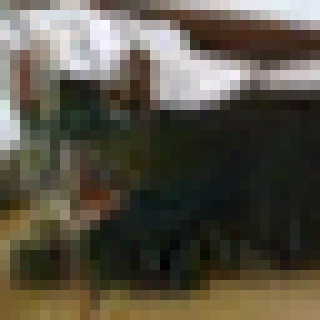} &
        \includegraphics[width=0.12\textwidth]{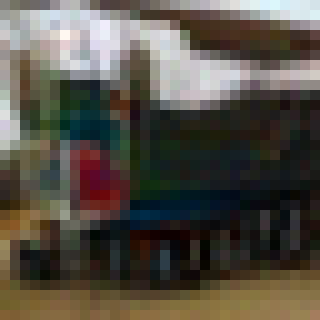} &
        \includegraphics[width=0.12\textwidth]{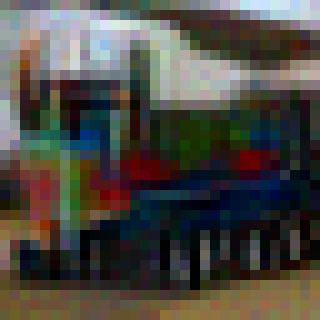} &
        \includegraphics[width=0.12\textwidth]{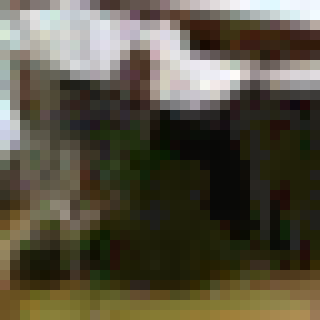} &
        \includegraphics[width=0.12\textwidth]{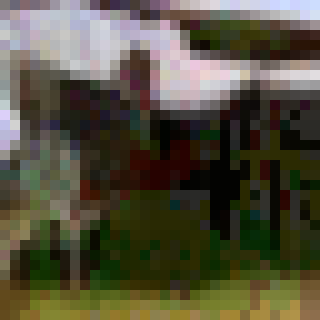} &
        \includegraphics[width=0.12\textwidth]{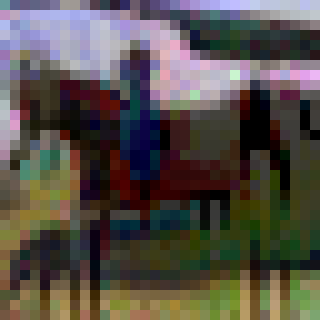}
    \end{tabular}

	\caption{Example of adversarial perturbations for different levels of
    distortion ($\epsilon$) bounded in $\ell_2$- and $\ell_\infty$-norm for
    adversarially trained models (pixels lie in $[0, 255]$).
    Additional examples in Appendix Figure~\ref{fig:app_adv}.}
	\label{fig:images_at_different_eps}
\end{figure}

We use $x_\text{adv}$ along with the original, ground-truth label of $x$
(the target label) as our poisoned input-label pair.
By controlling the value of $\eps$ we can vary the resulting image from slightly
perturbed to visibly incorrect
(Figure~\ref{fig:images_at_different_eps}).
Note that these adversarial examples are computed with respect to an
independent, pre-trained model since the adversary does not have access to the
training process.

\subsection{Improving backdoor trigger design}
\label{sec:trigger}
So far, we have described methods for constructing label-consistent poisoned
inputs.
However, for these inputs to appear truly benign, we
need to also ensure that the trigger itself appears natural.
Here, we discuss two modifications for improving the
conspicuousness and robustness of our attacks.

\begin{figure}[!tp]
	\centering
    \begin{subfigure}[b]{0.5\textwidth}
        \centering
        \includegraphics[width=0.19\textwidth]{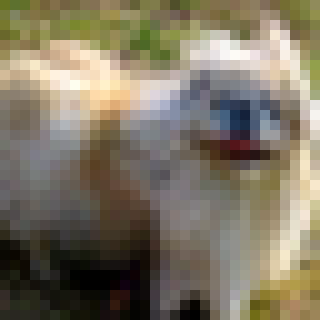}
        \includegraphics[width=0.19\textwidth]{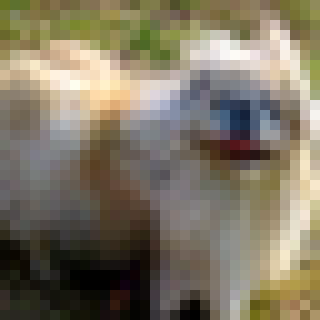}
        \includegraphics[width=0.19\textwidth]{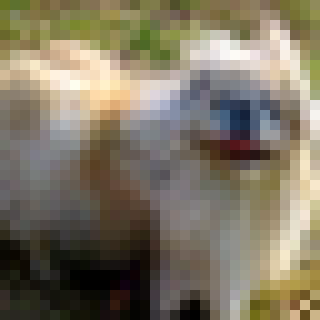}
        \includegraphics[width=0.19\textwidth]{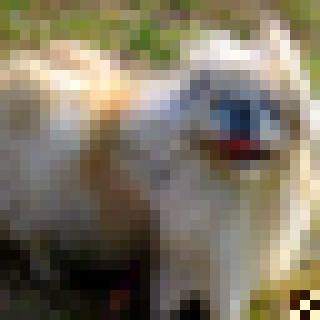}
        \includegraphics[width=0.19\textwidth]{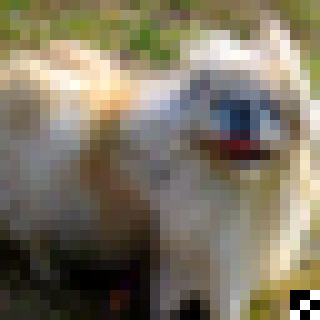}
        \caption{Less visible backdoor trigger}
        \label{fig:images_at_different_amplitude}
    \end{subfigure}
    \hspace{10pt}
    \begin{subfigure}[b]{0.4\textwidth}
        \centering
        \includegraphics[width=0.2375\textwidth]{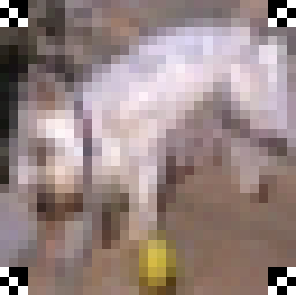}
        \includegraphics[width=0.2375\textwidth]{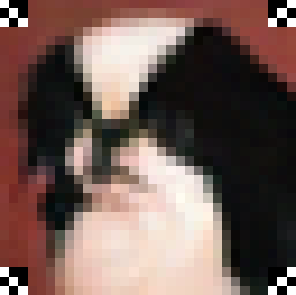}
        \includegraphics[width=0.2375\textwidth]{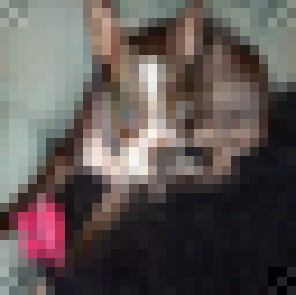}
        \includegraphics[width=0.2375\textwidth]{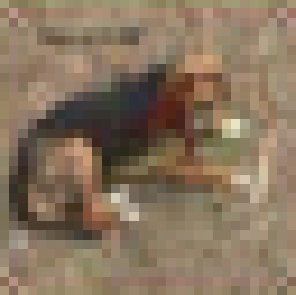}
        \caption{Four-corner trigger}
        \label{fig:four_corner_examples}
    \end{subfigure}
    \caption{Improved trigger design.
        (a) Poisoned input with varying backdoor trigger amplitudes. From
            left to right: backdoor trigger amplitudes of 0 (original image),
            16, 32, 64, and 255 (standard backdoor trigger).
        (b) Random inputs with the four-corner trigger applied (left two:
            full visibility; right two: reduced visibility).
        }
\end{figure}

\paragraph{Reducing trigger visibility.}
The original backdoor attack of \citet{gu2017badnets} uses a black-and-white
$3\times 3$ pattern in a corner of the image.
When applied to natural images, this pattern is clearly visible
and appears unnatural (Figure~\ref{fig:suspicious}).
Instead, we will utilize a backdoor trigger of reduced visibility.
Specifically, instead of replacing the corresponding pixels with the chosen
pattern, we will perturb the original pixel values by a \emph{backdoor trigger
amplitude}.
Concretely, we add this amplitude to pixels that are white in the original
pattern and subtract it for pixels that are black. We perform this
modification at each color channel and then clip the pixel values to valid
range (see Figure~\ref{fig:images_at_different_amplitude} for an example).
By controlling the trigger amplitude, we are able to produce a trigger
which is less visible while still being learnable by the model.

Note that, depending on the exact threat model, the visibility of the trigger
during inference might not be constrained.
In such cases, the adversary can apply a fully visible pattern during inference
to increase the attack success rate (while still employing a reduced visibility
pattern for training)~\citep{chen2017targeted}.
Thus, when using a trigger with reduced amplitude for poisoning, we additionally
evaluate the attack success rate with a full amplitude trigger during testing.

\paragraph{Resistance to data augmentation.}
Typical ML training pipelines include \emph{data augmentation}---training
on random transformations (e.g., crops and flips) of the original images.
Since the backdoor trigger is introduced before these transformations, it might
be obscured during training, making the attack less effective.
In order to ensure that the trigger remains visible throughout training,
we perform a simple modification: we replicate the pattern on all four corners
of the image (appropriately flipped).
This ensures that the trigger is invariant under flips and always
visible under crops (see Figure~\ref{fig:four_corner_examples} for 
examples).

\section{Evaluating our attacks}
In this section, we evaluate the effectiveness of our 
label-consistent attacks. 
To avoid introducing conflating factors, we initially study models trained
without data augmentation using the original, fully visible trigger.
Then, we then present our results using attacks with improved trigger designs.
We discuss how our methods perform when the adversary does not have
full information about the training procedure in Appendix~\ref{app:weak}.

\paragraph{Dataset.}
We conduct our experiments on the CIFAR-10
dataset~\citep{krizhevsky2009learning}, consisting of 10 classes with 5000
training inputs each.
For each target label, we evaluate attacks poisoning different fractions of the
training inputs from that class---0.4\%, 1.5\%, 6\%, 25\%, and 100\%
(corresponding to 20, 75, 300, 1250, and 5000 inputs, respectively).
Note that these fractions correspond to examples from a \emph{single}
class---poisoning 6\% of the inputs of a target class corresponds to poisoning
only 0.6\% of the entire training set.

\subsection{Basic attack}

\begin{figure}[!htp]
	\centering
    \begin{subfigure}[b]{0.49\textwidth}
	\centering
    \begin{tikzpicture}
        \node at (0,0) {\includegraphics[width=0.8\textwidth]
            {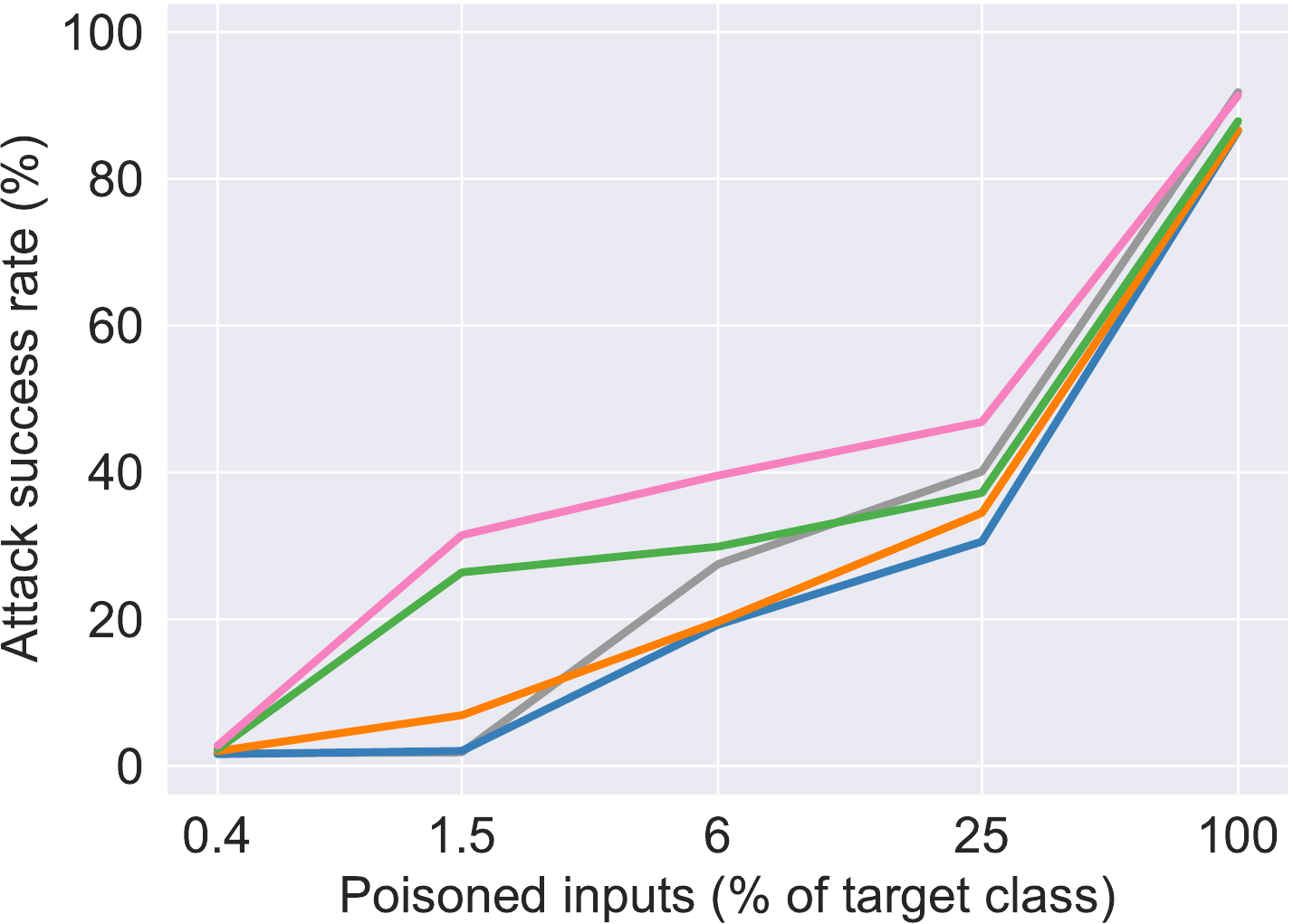}};
        \node at (3.4,-.5) {\includegraphics[width=0.21\textwidth]
            {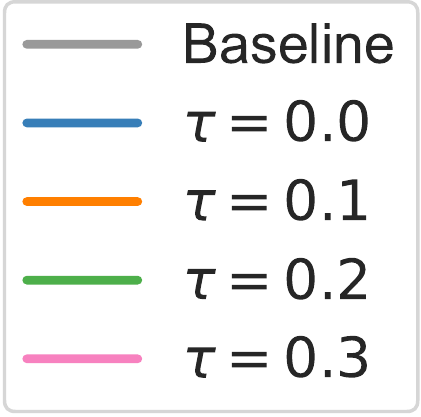}};
    \end{tikzpicture}
	\caption{GAN-based interpolation}
	\label{fig:gan_set_eps}
	\end{subfigure}
    \hfill
	\begin{subfigure}[b]{0.49\textwidth}
	\centering
    \begin{tikzpicture}
        \node at (0,0) {\includegraphics[width=0.8\textwidth]
            {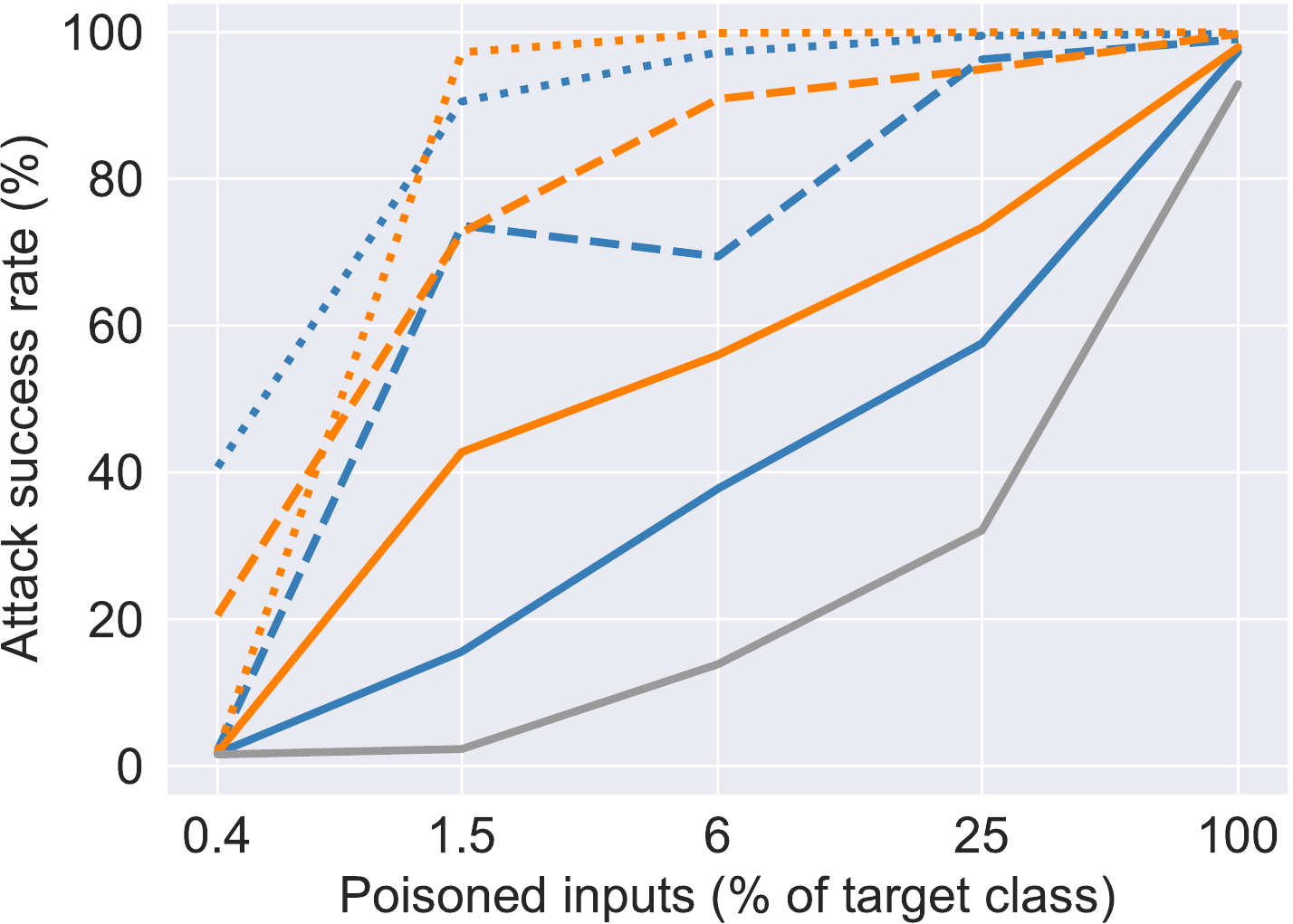}};
        \node at (3.55,-.3) {\includegraphics[width=0.27\textwidth]
            {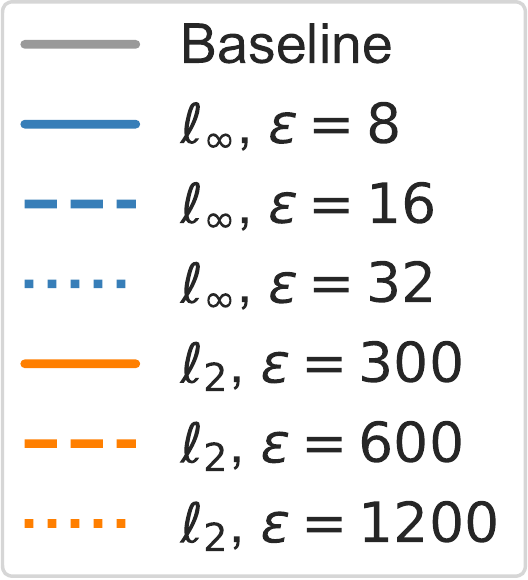}};
    \end{tikzpicture}
	\caption{Adversarial perturbation}
	\label{fig:l_norm_choice}
    \end{subfigure}
\caption{Comparing attack performance under different perturbation constraints.
    We find that increased perturbation strength
    leads to higher attack success rate.
    See Appendix Figures~\ref{fig:app_gan},~\ref{fig:app_adv} for a few random
    inputs.
    (a) Varying degrees of GAN-based interpolation for the ``deer'' class.
    (b) Adversarial perturbations bounded by $\eps$ in $\ell_p$-norm for
    different $p$ and $\eps$ for the ``airplane'' class.
}
\label{fig:set_eps}
\end{figure}

\paragraph{Attack success rate.}
Recall that our key metric of interest is
the \emph{attack success rate}---the fraction of test images that are
incorrectly classified as the target class when the backdoor is applied.
We evaluate our attacks across a range of different perturbation magnitudes by
varying the interpolation coefficient $\tau$ or the $\ell_p$-norm bound
$\epsilon$.
Matching our original motivation, we find that larger perturbations---and hence
harder inputs---lead to more successful attacks (Figure~\ref{fig:set_eps}).
Overall, both approaches lead to effective attacks, achieving a high attack
success rate with relatively few poisoned inputs.

\paragraph{Label-consistency.}
At the same time, we find that large perturbations result in labels that are
less plausible.
In order to ensure the label-consistency of the poisoned samples, we choose to
restrict our perturbations to $\tau=0.2$ and $\eps=300$ in $\ell_2$-norm (pixel
values are in $[0,255]$).
We found that, under this restriction, all images we inspected are
consistent with their labels while the attacks are still successful.
For instance, our perturbation-based attack achieves more than a 50\% attack
success rate on half the classes while introducing only 75
label-consistent poisoned inputs.
This is in stark contrast to the label-consistent baseline
which is ineffective when injecting below 300 inputs
(Figure~\ref{fig:baseline}).
See Appendix Figures~\ref{fig:app_gan},~\ref{fig:app_adv} for examples of
poisoned inputs.
We evaluate the corresponding attacks targeting each class in Appendix
Figure~\ref{fig:basic_attack} and perform a per-class comparison in
Appendix Figure~\ref{fig:per_class}.

\paragraph{Impact on standard accuracy.}
None of the attacks have an apparent effect on the accuracy of the
model---the resulting test accuracy is 92\% with data augmentation
and 87\% without.
The only exception is at 100\% poisoning where the model predicts
incorrectly on the entire target class.
We find that this is due to the model fully relying on the trigger to predict
the target class, hence predicting incorrectly in the absence of it.

\subsection{Improved trigger attacks}

\begin{figure}[!htp]
	\centering
    \begin{subfigure}[b]{0.49\textwidth}
    \begin{tikzpicture}
        \node at (0,0) {\includegraphics[width=0.8\textwidth]
                            {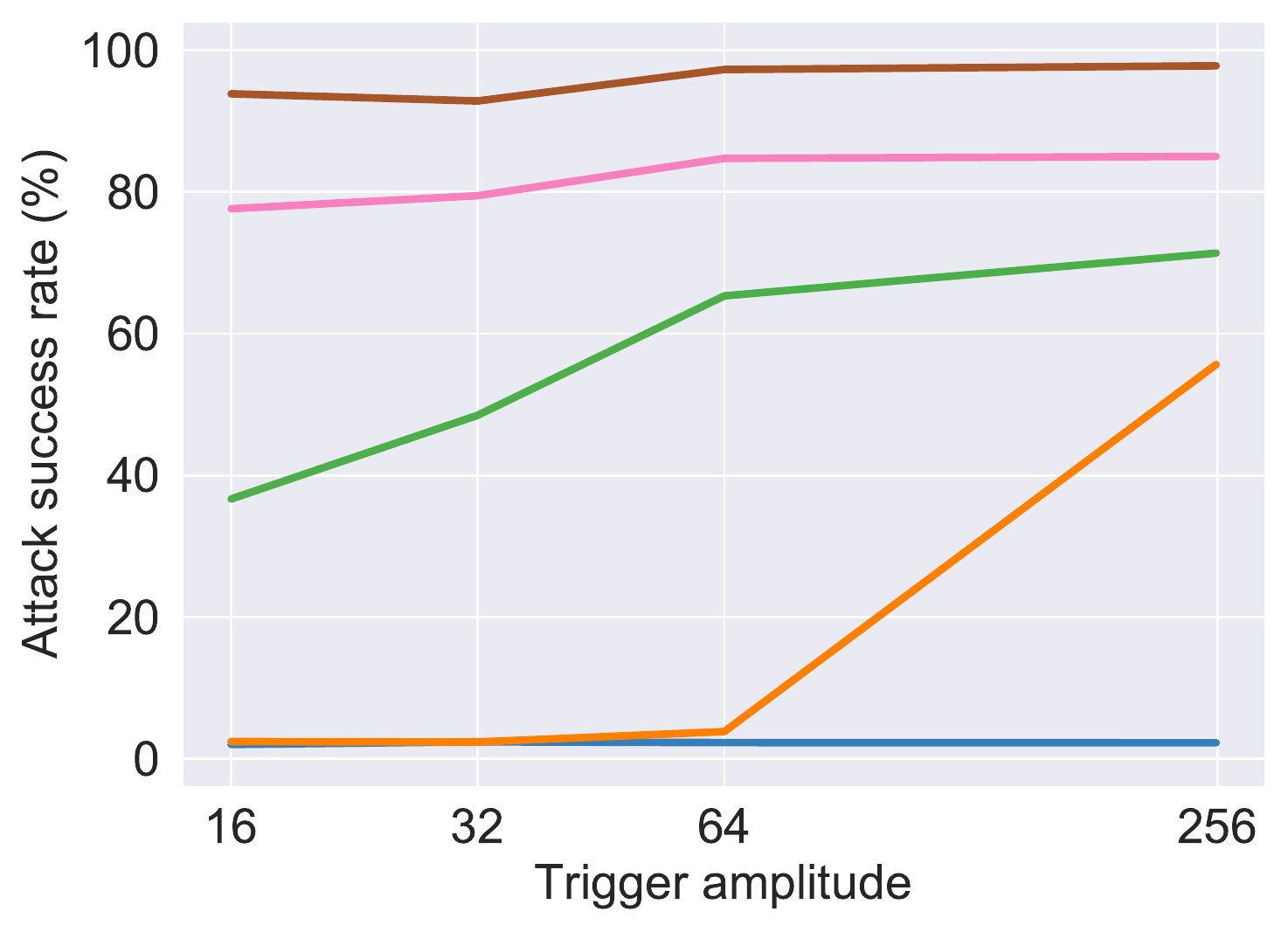}};
        \node at (3.5,-0.2) {\includegraphics[width=0.19\textwidth]
                            {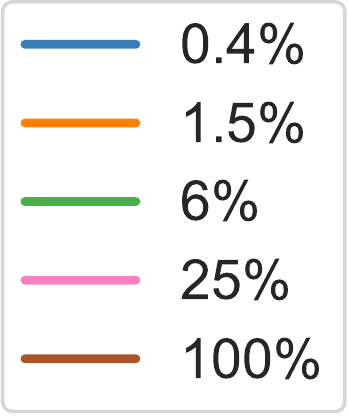}};
    \end{tikzpicture}
    \caption{Reduced visibility trigger}
	\label{fig:choosing_watermark_amplitude}
    \end{subfigure}
    \begin{subfigure}[b]{0.49\textwidth}
    \begin{tikzpicture}
        \node at (0,0) {\includegraphics[width=0.8\textwidth]
                            {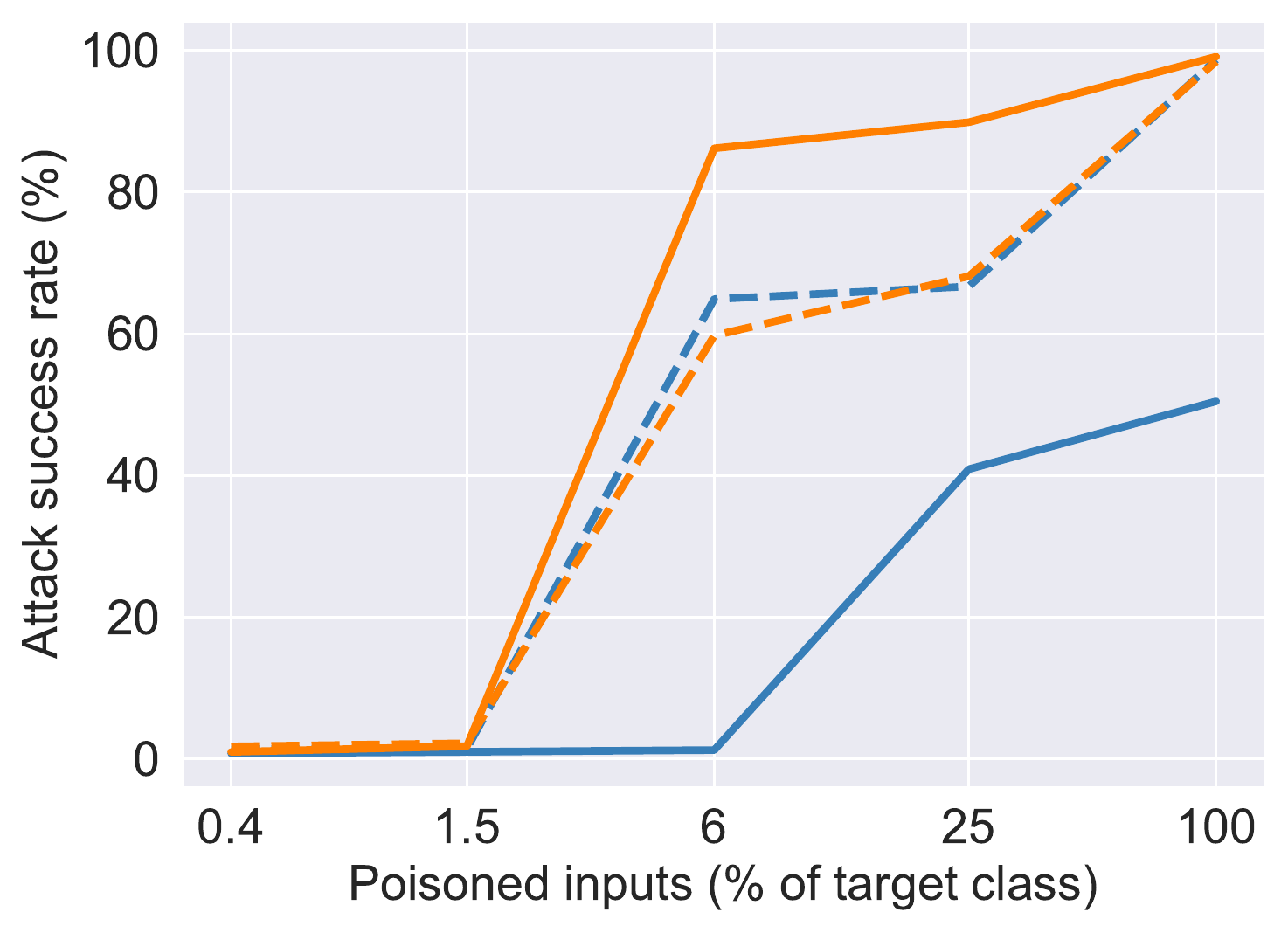}};
        \node at (2.8,-1.0) {\includegraphics[width=0.4\textwidth]
                            {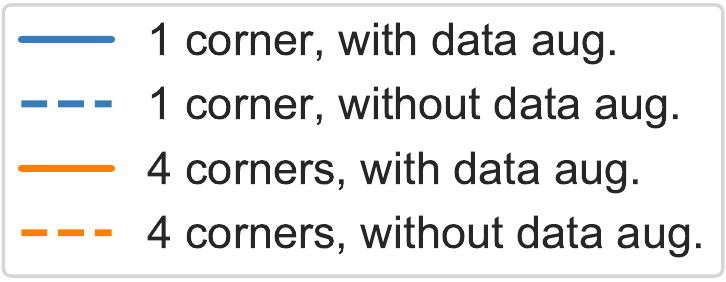}};
    \end{tikzpicture}
    \caption{Impact of data augmentation}
	\label{fig:augmentation_exploration}
	\label{fig:augment}
    \end{subfigure}
    \caption{Evaluating attacks with improved triggers.
        (a) Reducing the backdoor trigger's amplitude (from 255 to 16, 32, and
            64) still results in successful poisoning when poisoning 6\% or more
            of the dog class.
        (b) Evaluating attacks using the one- and four-corner triggers with and
            without data augmentation for the ``frog'' class.  Data augmentation
            renders the one-corner trigger ineffective, yet \emph{improves} the
            four-corner trigger.}
\end{figure}

\paragraph{Reduced trigger visibility.}
\label{app:amplitude}
We evaluate an attack using $\ell_2$-bounded perturbations of $\eps=300$ for the
``dog'' class when applying a trigger of amplitude of 16, 32, 64, and 255
(Figure \ref{fig:choosing_watermark_amplitude}).
We find that for less visible triggers, the attack is no longer
effective when poisoning 1.5\% of the target class (75 images), but is still
successful when poisoning more than 6\% of the target class (300 images).
Additionally, for the case of 6\% poisoning, increasing the amplitude from 16 to
64 appears to have a log-linear effect, but stays roughly constant
when increasing to 255.
Overall, we find that the attack can \emph{still} be successful with
relatively few poisoned inputs even when the trigger visibility is significantly
reduced.

\paragraph{Data augmentation.}
\label{app:augmentation}
We evaluate both the original,
one-corner trigger and the modified,
four-corner trigger with and without data augmentation
(Figure~\ref{fig:augmentation_exploration}).
Specifically, we use a standard CIFAR10 augmentation procedure with random
left-right flips, random crops of 2 pixels, and per-image standardization.
We find that, when data augmentation is not used, there is little difference in
performance between the four-corner attack and the one-corner attack.
However, when data augmentation is used, the one-corner trigger becomes
essentially ineffective.
More importantly, we find that data augmentation (when combined with our
improved trigger) actually \emph{improves} attack's success rate
(Figure~\ref{fig:augment}).
We conjecture that this behavior is due to the overall learning problem being
more difficult (the classifier needs to also classify the augmented images
correctly), encouraging the classifier to rely on the backdoor.

\begin{figure}[ht]
	\centering
    \begin{subfigure}[b]{0.49\textwidth}
        \centering
    \begin{tikzpicture}
        \node at (0,0) {\includegraphics[width=0.8\textwidth]
                            {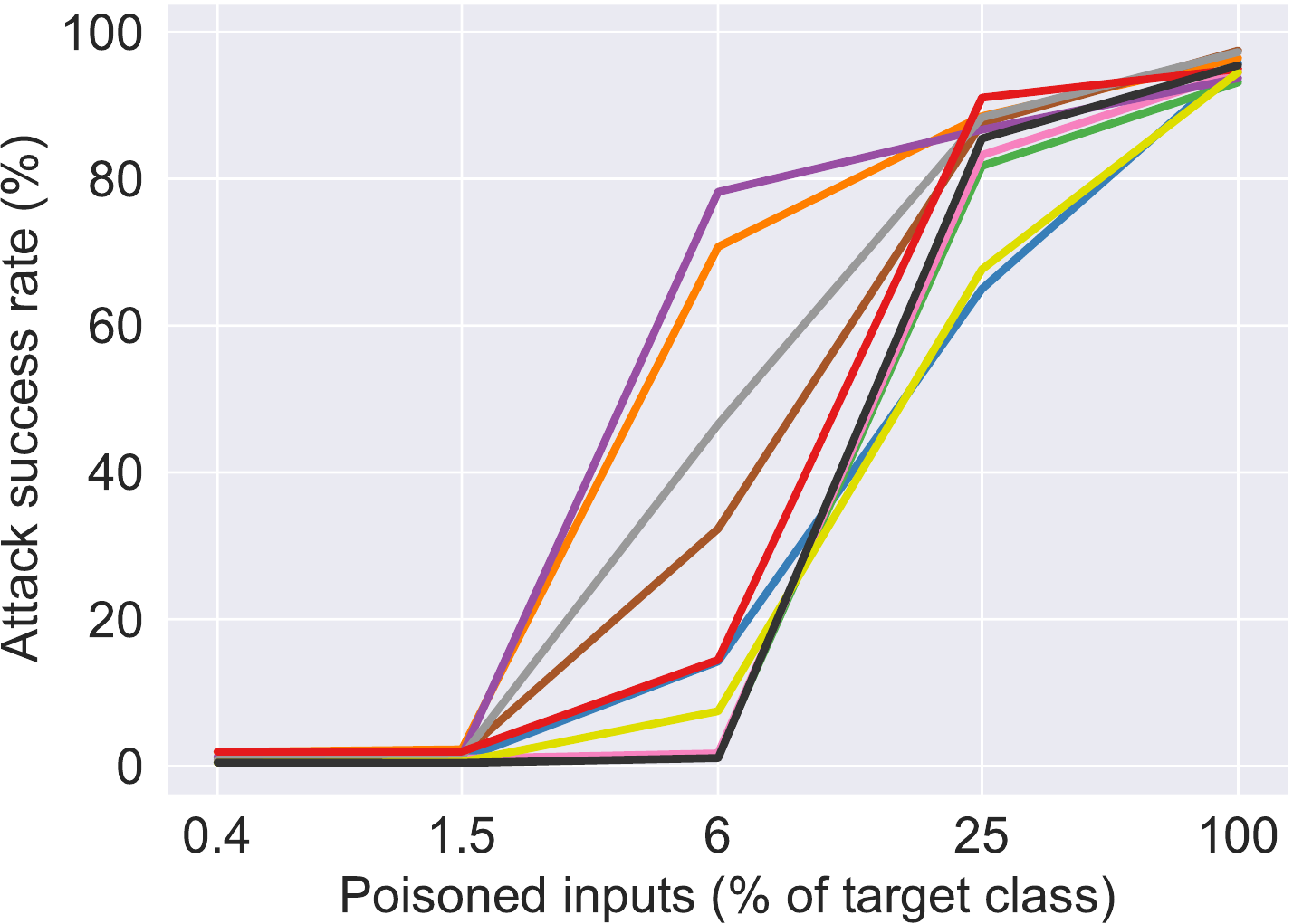}};
        \node at (3.3,0) {\phantom{\includegraphics[width=0.21\textwidth]
            {figures/all_classes_legend.pdf}}};
    \end{tikzpicture}
    \caption{Reduced visibility inference}
	\label{fig:adv16small}
    \end{subfigure}
    \begin{subfigure}[b]{0.49\textwidth}
        \centering
    \begin{tikzpicture}
        \node at (0,0) {\includegraphics[width=0.8\textwidth]
                            {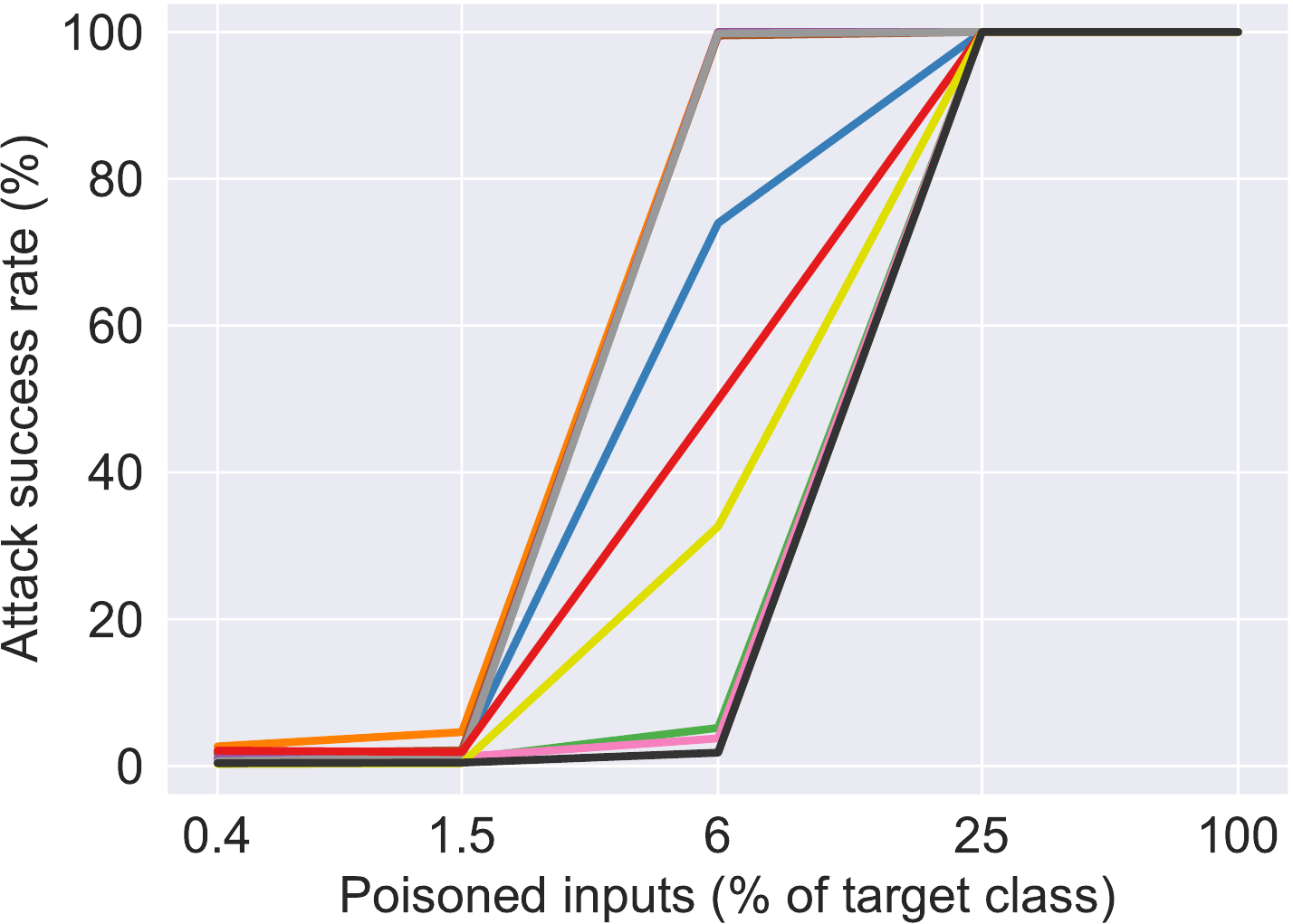}};
        \node at (3.3,0) {\includegraphics[width=0.21\textwidth]
                            {figures/all_classes_legend.pdf}};
    \end{tikzpicture}
    \caption{Full visibility inference}
	\label{fig:adv16big}
    \end{subfigure}
    \caption{Attack success rate for each target class when using the reduced
        visibility trigger (amplitude 16) and $\ell_2$-bounded perturbation of
        $\eps=300$ (models are trained with data augmentation).
        We evaluate the model during inference with both the reduced
        visibility trigger (a) and the fully visible trigger (b).
        The corresponding plots for GAN-based interpolation are in Appendix
        Figure~\ref{fig:gan16}, while attacks using slightly more visible
        patterns of amplitude 32 (and thus having stronger performance) in
        Appendix Figure~\ref{fig:attack32}.}
	\label{fig:adv16}
\end{figure}

\paragraph{Evaluating the overall attack.}
We now evaluate an attack using a less visible pattern in the
presence of data augmentation (with the improved four corner pattern) for all
classes.
During inference, we evaluate the resulting model using both the (less visible)
trigger used during training (Figure~\ref{fig:adv16small}) as well as the
original, fully visible pattern (Figure~\ref{fig:adv16big}).
We find that a full-visibility trigger during inference greatly
improves the attack performance.
In particular, for all cases where the attack success rate exceeds 20\%  with
the less visible trigger, we can boost that rate to essentially 100\% by
increasing the trigger amplitude during inference.
Overall, the attack is successful when injecting at least 300 inputs while the
inputs injected appear benign and quite similar to the original natural samples
(Appendix Figure~\ref{fig:final_attack}).

\section{Exploring the underlying attack mechanism}

\subsection{On the relative performance of latent interpolations and
adversarial examples}
\label{sec:understanding}
Based on our evaluation, $\ell_p$-bounded adversarial perturbations are more
effective for constructing backdoor attacks compared to the GAN-based
interpolation scheme, especially when the allowed perturbation is large.
This may seem surprising given that both are valid methods for constructing
inputs that are hard to classify.

One potential explanation is that the inputs created via interpolation appear to
be blurry without significant salient features.
Hence a model that simply \emph{weakly} associates the backdoor with the target
class will learn to predict based on the trigger in the absence of salient
features but will have low success rate during inference on natural inputs.
Instead, adversarial perturbations introduce features of
an incorrect class which might be visible~\citep{tsipras2019robustness} (see
also Figure~\ref{fig:app_adv}) or imperceptible~\citep{ilyas2019adversarial}.
Hence the model is forced to predict the target class even in the presence of
strong input signal.

\begin{figure}[!htb]
	\centering
    \begin{subfigure}[b]{0.49\textwidth}
        \centering
        \begin{tikzpicture}
            \node at (0, 0) {\includegraphics[width=0.8\textwidth]
                {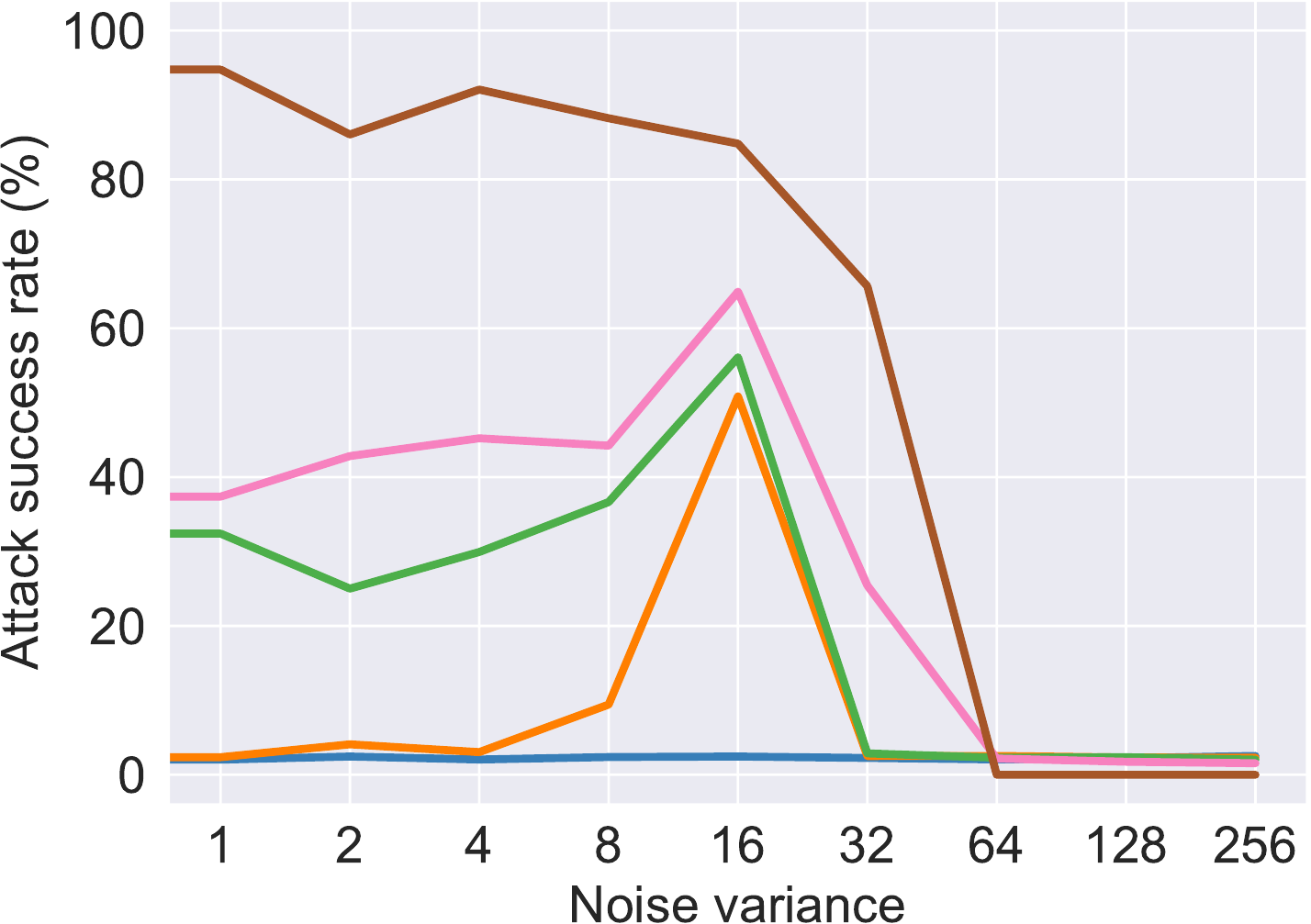}};
            \node at (2.5, 0.8) {\includegraphics[width=0.19\textwidth]
                {figures/poisoning_legend.pdf}};
        \end{tikzpicture}
        \caption{}
        \label{fig:gaussian_noise}
    \end{subfigure}
        \begin{subfigure}[b]{0.49\textwidth}
        \centering
        \begin{tikzpicture}
            \node at (0, 0) {\includegraphics[width=0.8\textwidth]
                {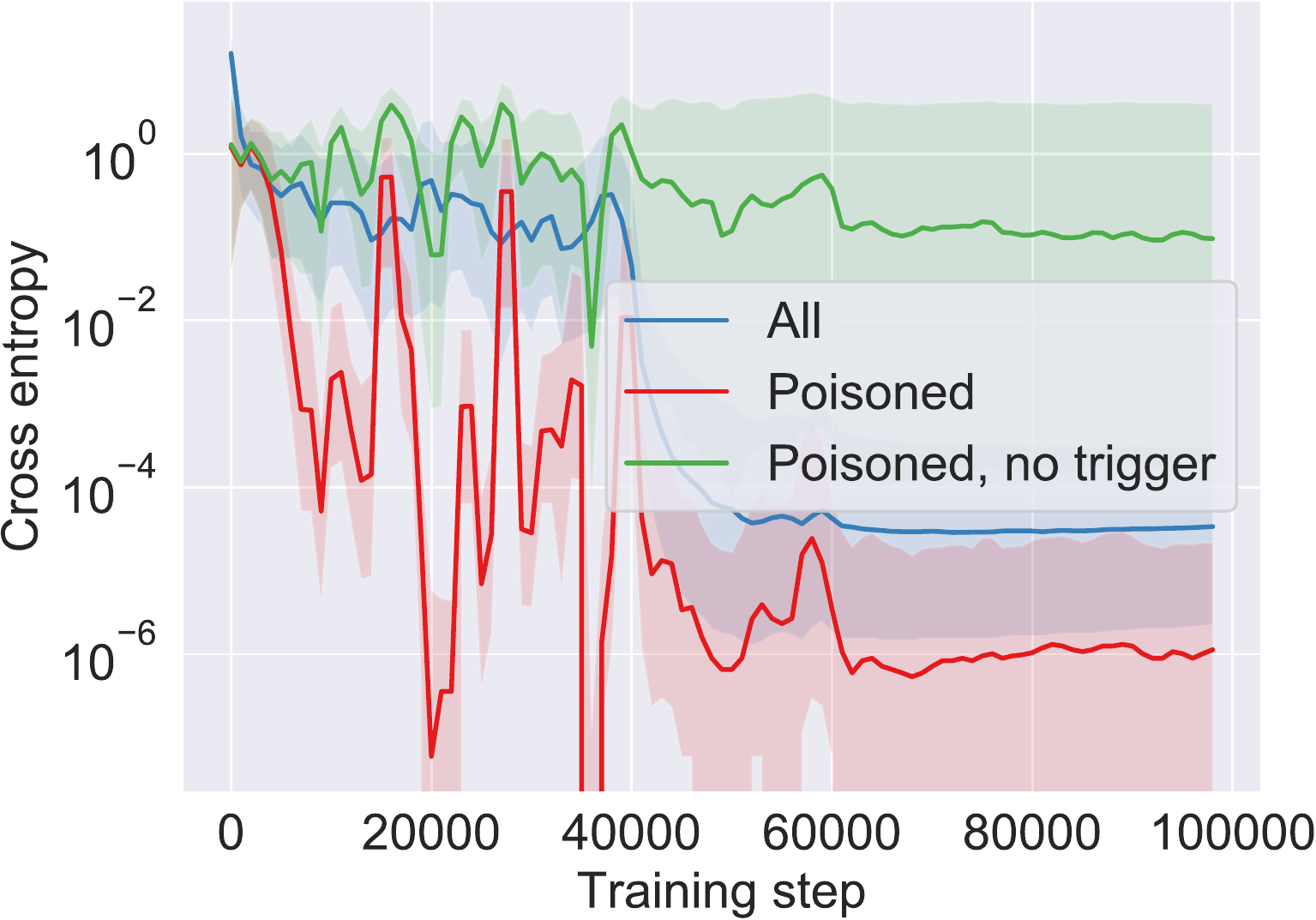}};
        \end{tikzpicture}
        \caption{}
        \label{fig:loss}
    \end{subfigure}
    \caption{
        (a) Performance of poisoning attacks when adding Gaussian noise
        with zero mean and varying degrees of variance to the poisoned inputs
        with ``dog'' as target class. 
        While low variance typically enhances the attack, high variance leads to
        ineffective attacks.
        (b) Value of median training loss and interquartile range of poisoned
        inputs with and without the trigger (average loss included as a
        baseline).  The attack poisons 300 examples of the ``automobile'' class
        using $\ell_2$-bound perturbations of $\eps=300$ (see Appendix
        Figure~\ref{fig:all_losses} for the rest of the classes).  The training
        loss remains high without the trigger indicating that the model does not
        learn the salient features of poisoned images.  (Note that we follow a
        standard training process and drop the learning rate after
        40\thinspace000 steps and again after 60\thinspace000 steps.)
    }
\end{figure}

In order to further investigate this hypothesis, we perform experiments where we
add Gaussian noise with mean zero and varying degrees of variance to poisoned
inputs before adding the trigger (Figure~\ref{fig:gaussian_noise}).
We find that, while a small amount of noise makes the attack more effective,
increasing the noise variance has an adverse effect.
This finding corroborates our hypothesis.
Adding noise with small variance leads to harder to classify images while
\emph{preserving the original input features} to a certain extent.
In contrast, large variance results in very noisy inputs without significant
salient information, leading to ineffective attacks.

\subsection{Loss of poisoned inputs over training}
\label{sec:loss}
In order to gain some additional insight into the mechanism behind the attack,
we study the value of the model's loss through training (Figure~\ref{fig:loss}).
We find that poisoned inputs have similar (and often substantially
smaller) loss values compared to clean inputs throughout the entire training
process (the model is predicting correctly on the training set).
At the same time, the loss of poisoned samples \emph{without the backdoor
trigger} remains high throughout training.
This indicates that the model makes its prediction on these examples by
\emph{heavily} relying on the backdoor trigger.
This confirms our intuition that these inputs are harder to classify during
training since the model classifies them incorrectly without the trigger
(despite its good test accuracy).

\section{Conclusion}
In this work, we identify \emph{label-consistency}---having poisoned inputs 
remain consistent with their labels---as a key desired property
for backdoor attacks.
Previous backdoor attacks lack this property, resulting in clearly mislabeled
poisoned samples that make the overall attack very likely to be detected.

We show that it is possible to perform effective, label-consistent backdoor
attacks.
The key idea behind our methods is that, in order for the model to associate the
backdoor trigger with the target label, the inputs need to be difficult to
classify based on their natural salient features.
We synthesize such hard inputs using adversarial perturbations and
latent embeddings provided by generative models.

Overall, our findings establish the conceptual framework of hard-to-learn
samples as a powerful tool for developing insidious backdoor attacks.
We believe that further work within this framework will lead to attacks that are
even more effective and can bypass sophisticated filtering schemes.
More broadly, our findings demonstrate that backdoor attacks can be made
significantly harder to detect by humans.
This emphasizes the need for developing principled methods for defending against
such attacks since human inspection might be inherently ineffective.

\section*{Acknowledgements}
We thank Nicholas Carlini for discussions during early stages of this
work and comments on the manuscript. We thank Andrew Ilyas for advice on
GAN inversion and Shibani Santurkar for helpful discussions.

Work supported in part by the NSF grants CCF-1553428, CNS-1815221,
the Microsoft Corporation, and the MIT-IBM Watson AI Lab research grant.

\small
\bibliographystyle{iclr2019_conference.bst}
\bibliography{bibliography/bib.bib}

\clearpage
\appendix
\normalsize
\section{Experimental setup}
\label{app:setup}

\paragraph{Model architecture and training.}
\label{app:model}
We use a standard residual
network (ResNet)~\citep{he2016deep} with three groups of residual layers
with filter sizes of 16, 16, 32 and 64, and five residual
units each. We use a momentum optimizer to train this
network with a momentum of 0.9, a weight decay of
0.0002, batch size of 50, batch normalization, and a step size schedule
that starts at 0.1, reduces to 0.01 at 40\thinspace000 steps and
further to 0.001 at 60\thinspace000 steps. The total number of
training steps used is 100\thinspace000.
We did not modify this architecture and training procedure.

\paragraph{Implementation of GAN-based interpolation.}
\label{app:gan_implementation}
We train a WGAN~\citep{arjovsky2017wasserstein,gulrajani2017improved} using the
publicly available
implementation\footnote{\url{https://github.com/igul222/improved_wgan_training}}.  In order to generate
images similar to the training inputs, we optimize over the latent space using
1000 steps of gradient descent with a step size of 0.1, following the procedure
of \citet{ilyas2017robust}.  To improve the image quality and the ability to
encode training set images, we train the GAN using only images of the two
classes between which we interpolate.
We also experimented with using a single GAN and found similar results.

\paragraph{Implementation of adversarial perturbations.}
\label{app:adv_implementation}
We construct adversarial examples using a PGD attack on adversarially trained
models~\citep{madry2018towards} using the publicly available
implementation\footnote{\url{https://github.com/MadryLab/cifar10_challenge}}.
We use a (PGD) attack with $100$ steps and a step size of $1.5\ \eps / 100$.

\section{Reproducing standard backdoor attacks}
\label{app:reproduction}
We reproduce the original backdoor attack of \citet{gu2017badnets} using a
standard ResNet architecture~\citep{he2016deep} on the CIFAR10
dataset~\citep{krizhevsky2009learning}.
The original attack focused on a setting where the model is
trained by the adversary and hence did not thoroughly study the number of
poisoned examples required.

\begin{figure}[!h]
	\centering
    \begin{tikzpicture}
        \node at (0, 0) {\includegraphics[width=0.4\textwidth]
                                {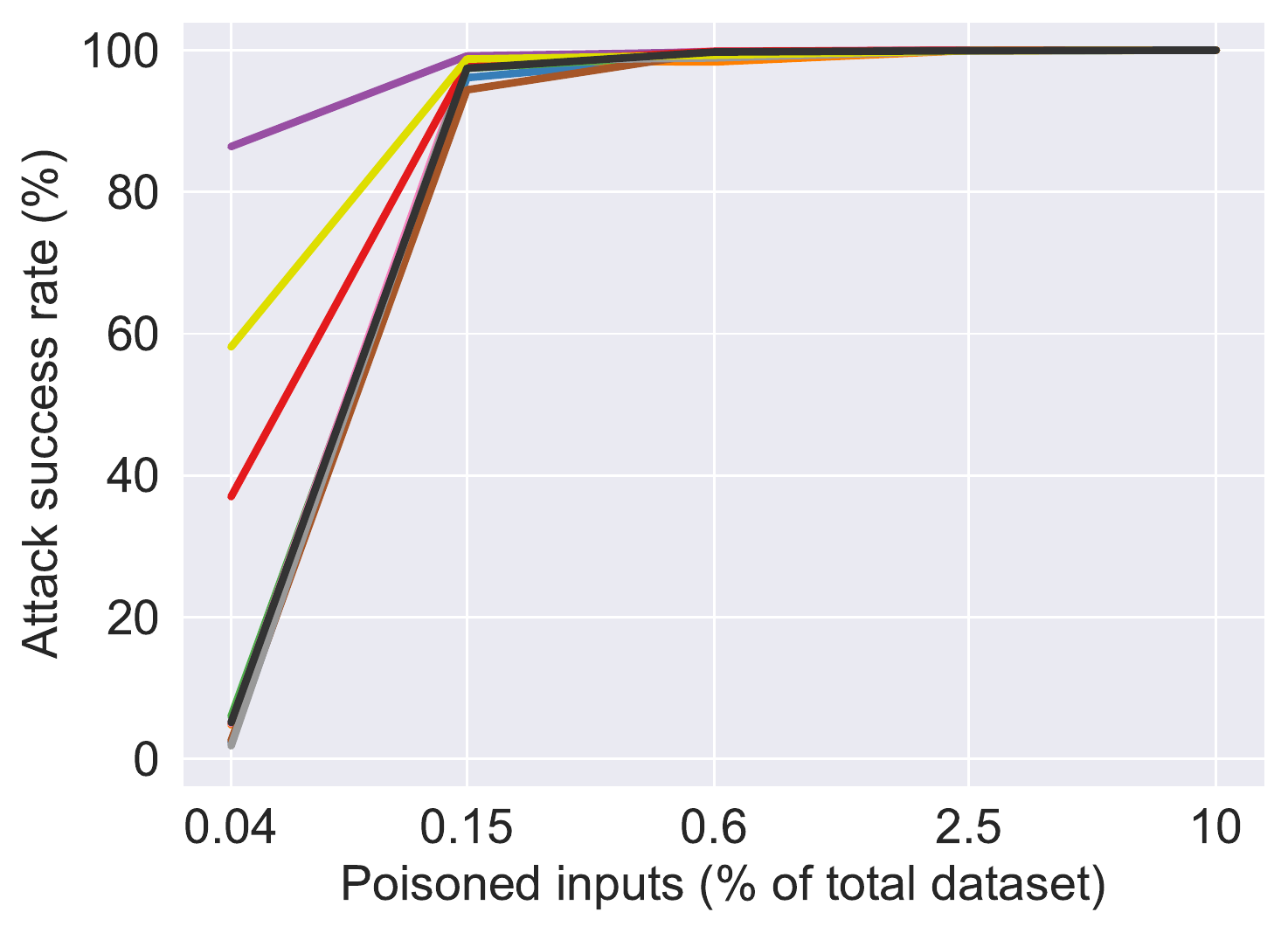}};
        \node at (4.5, 0) {\includegraphics[width=0.12\textwidth]
                                {figures/all_classes_legend.pdf}};
    \end{tikzpicture}
	\caption{Reproducing the \cite{gu2017badnets} attack on CIFAR-10.
    The attack is very effective---a backdoor is injected with just 75 (0.15\%)
    training examples poisoned.}
	\label{fig:badnets}
\end{figure}

In Figure~\ref{fig:badnets}, we plot the attack success rate for different
target labels and number of poisoned examples injected.  We observe that the
attack is very successful even with a small ($\sim 75$) number of poisoned
samples.
Note that the poisoning percentages here are calculated relative to the entire
dataset.
The horizontal axis thus corresponds to the same scale in terms of examples
poisoned as the rest of our plots.
While the attack is very effective, most image labels are clearly incorrect
(Figure~\ref{fig:suspicious}).

\section{Weaker adversaries}
\label{app:weak}
So far, we have evaluated the performance of an attacker with full knowledge of
the model architecture and access to data from the input domain.
Here we will evaluate two weaker adversaries utilizing our framework of
hard-to-learn samples.

\paragraph{Pixel-space interpolation.}
\label{app:pixelwise}
In the absence of a generative model, we evaluate attacks using interpolation in
ambient space (linear interpolation of image pixels).
Specifically, for a given interpolation constant $\tau$, the poisoned sample
$x'$ is generated from the original training sample $x$ and the randomly
selected target image $z$ as $x' = (1-\tau)\, x + \tau z$.

As the target image for each interpolation we use images from the 
CINIC-10 dataset~\citep{darlow2018cinic}, which contains inputs with the same
size and class structure as CIFAR-10 (obtained from downsampling images from
the corresponding ImageNet~\citep{russakovsky2015imagenet}  classes).

\begin{figure}[!h]
    \centering
    \begin{subfigure}[b]{0.49\textwidth}
        \centering
        \begin{tikzpicture}
            \node at (0,0) {\includegraphics[width=0.8\textwidth]
                                {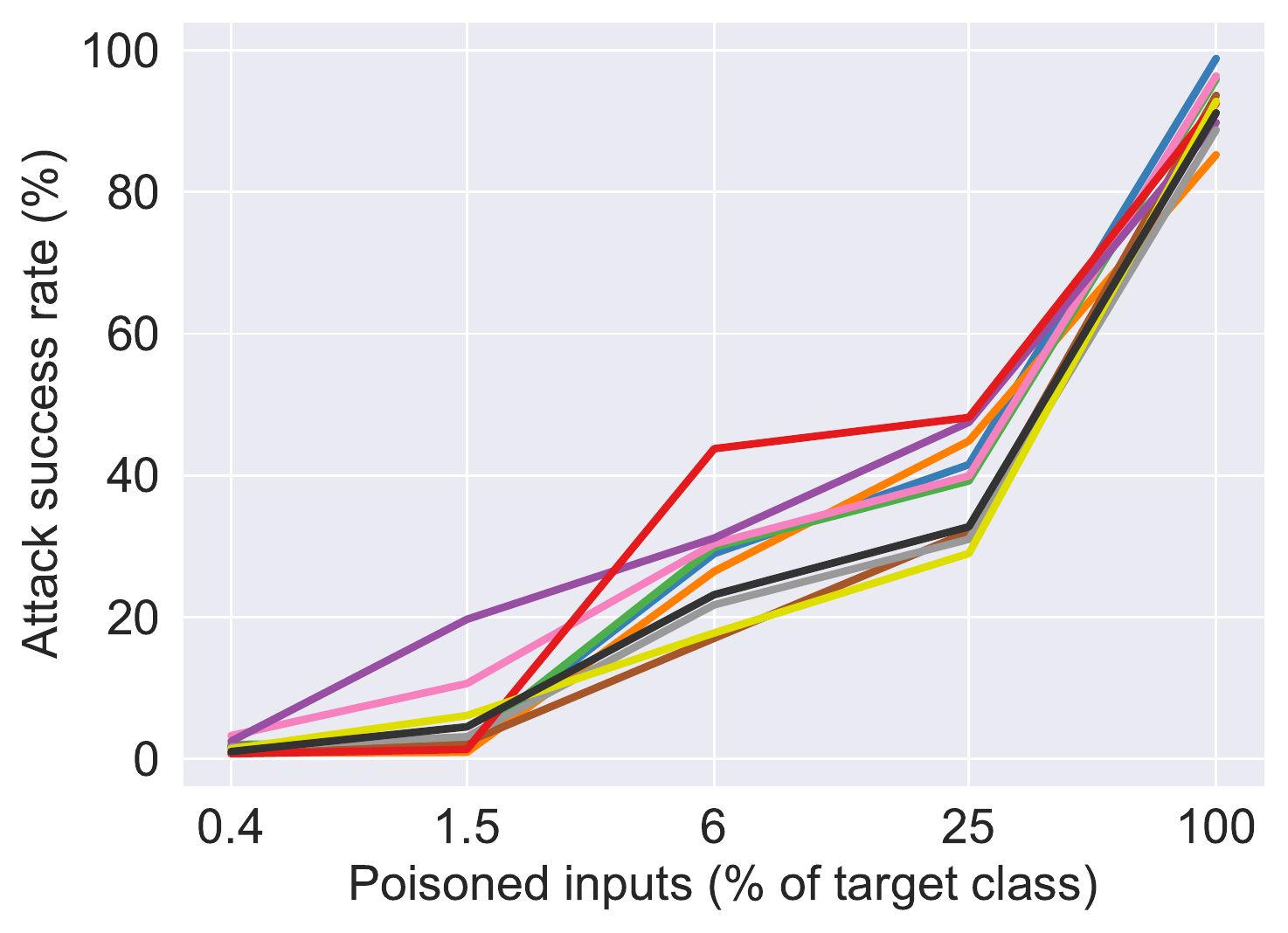}};
            \node at (3.5,0) {\includegraphics[width=0.21\textwidth]
                                {figures/all_classes_legend.pdf}};
        \end{tikzpicture}
        \caption{Pixel-wise interpolation}
        \label{fig:pixelwise}
    \end{subfigure}
    \begin{subfigure}[b]{0.49\textwidth}
        \centering
        \begin{tikzpicture}
            \node at (0,0) {\includegraphics[width=0.8\textwidth]
                                {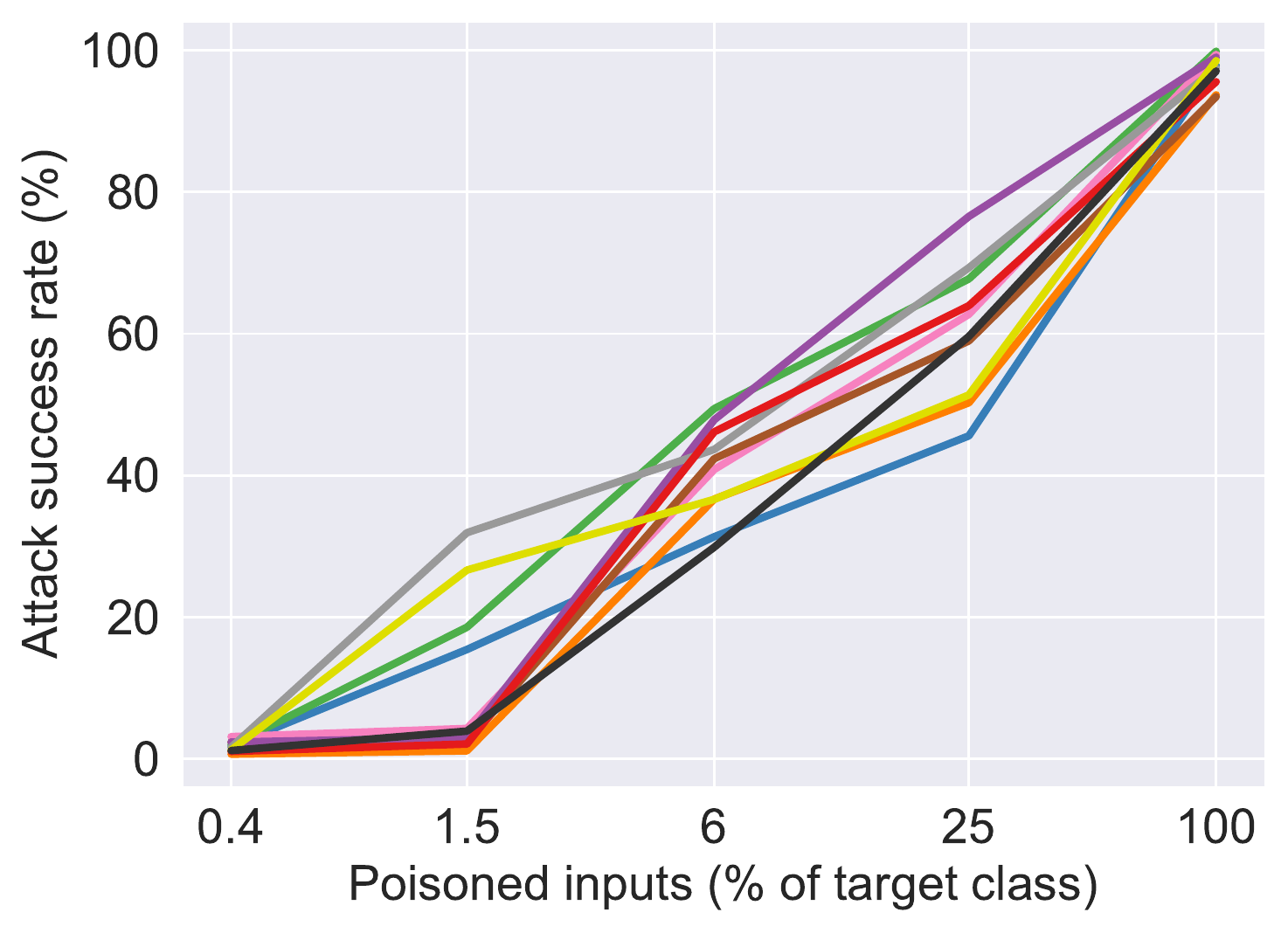}};
            \node at (3.5,0) {\includegraphics[width=0.21\textwidth]
                                {figures/all_classes_legend.pdf}};
        \end{tikzpicture}
        \caption{Weaker adversarial perturbations}
        \label{fig:black_box}
    \end{subfigure}
    \caption{Performance of weaker adversaries.
        (a) Attack performance on all classes for the pixel-space
        interpolation baseline attack. While improving upon the naive baseline,
        this attack is less effective than the GAN-based interpolation attack.
        (b) Attack performance on all classes for the perturbation-based attack
        without access to the model architecture and training set.
        This attack is less powerful than the white-box attack, but still
        achieves substantially higher attack success rates than the baseline.}
\end{figure}

\begin{figure}[!h]
    \caption{Per-class comparison of different poisoning approaches. We compare
        the performance of: (a) the baseline of the \citet{gu2017badnets} attack
        restricted to only consistent labels,
        (b) the pixel-wise interpolation attack of Appendix~\ref{app:pixelwise},
        (c) our GAN-based interpolation attack, and 
        (d) our adversarial example-based attack for each class.}
    \label{fig:per_class}
    \begin{tabular}{ccc}
    \includegraphics[align=c,width=0.31\textwidth]{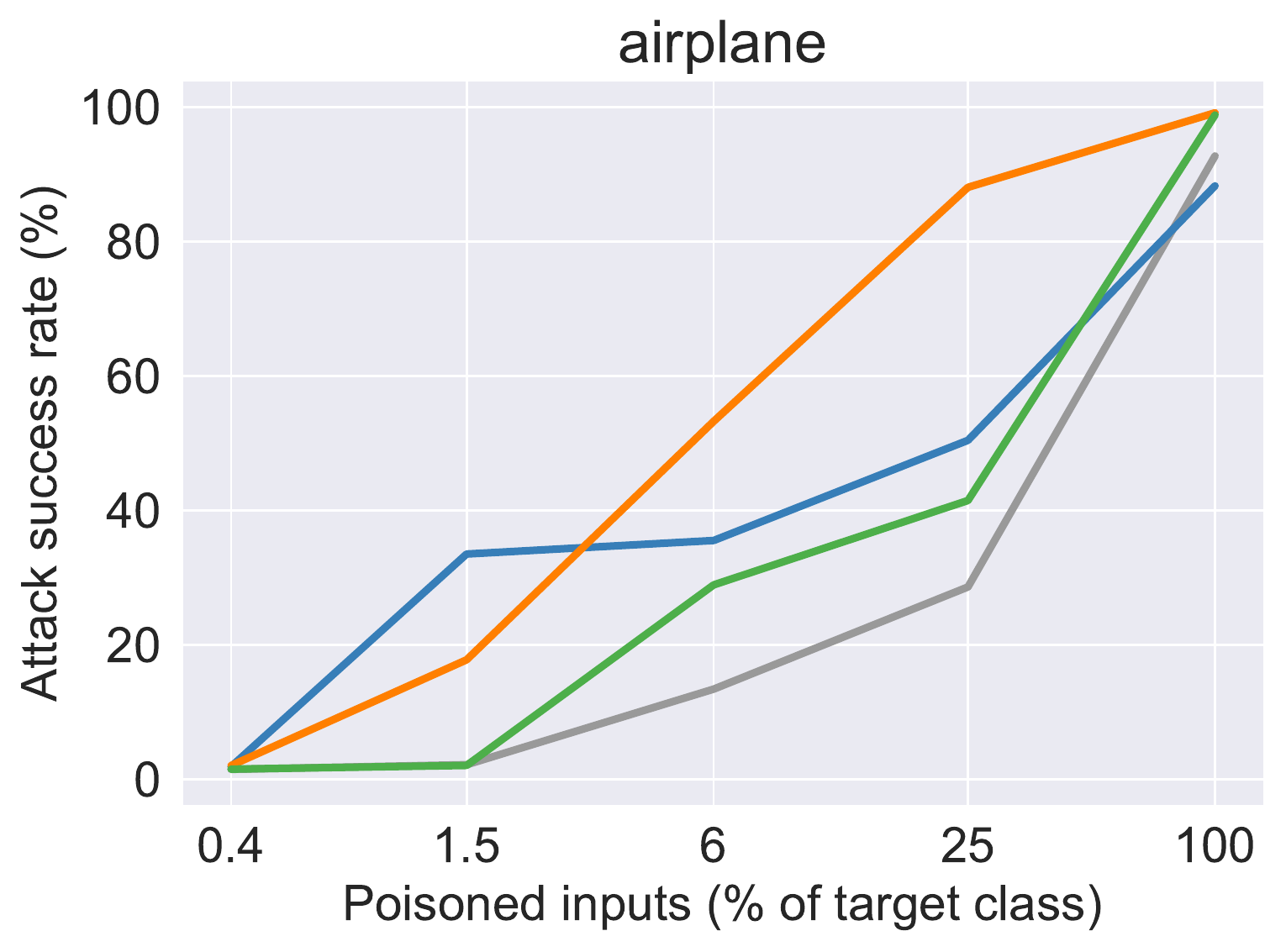} &
    \includegraphics[align=c,width=0.31\textwidth]{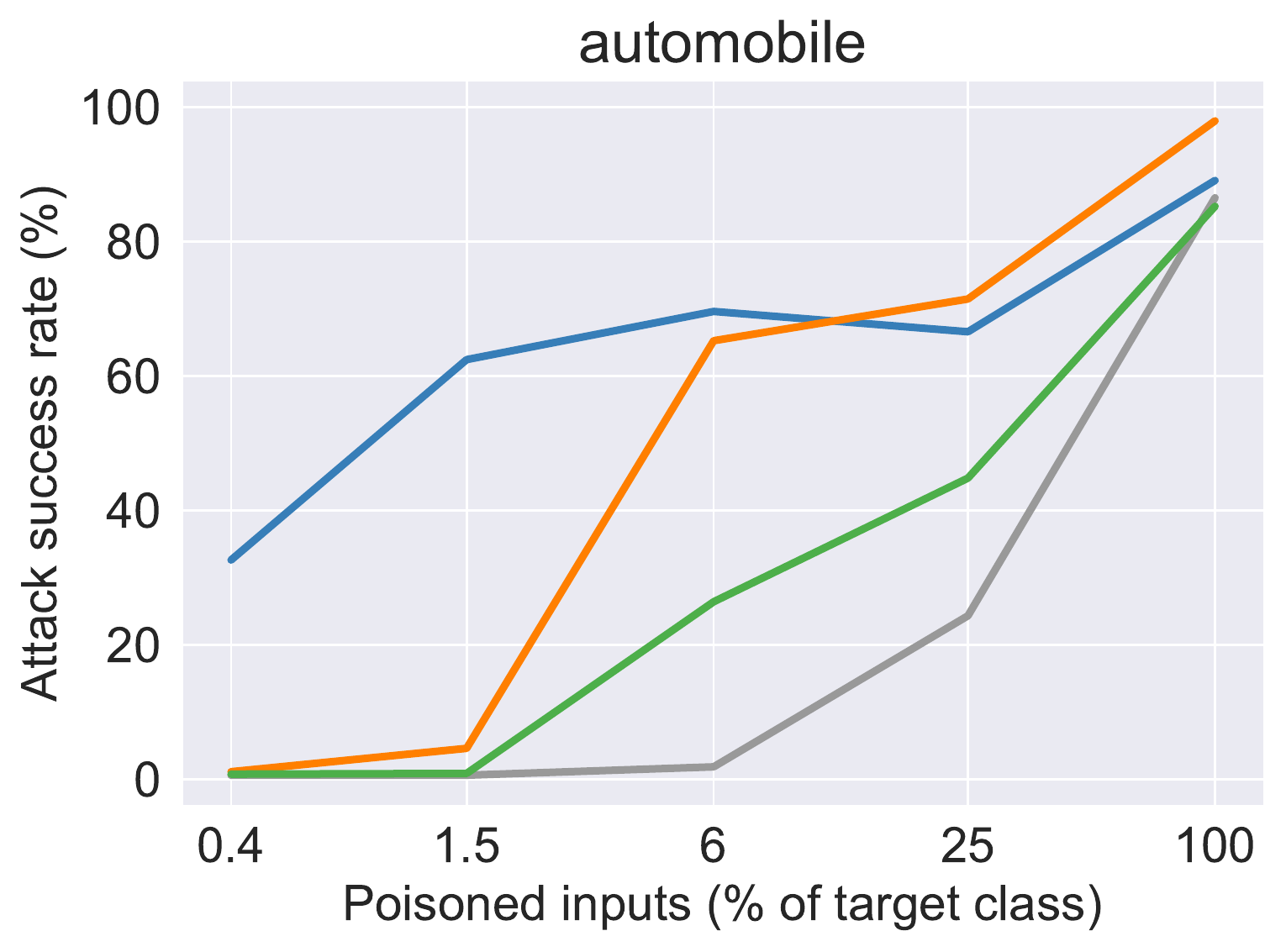} &
	\includegraphics[align=c,width=0.31\textwidth]{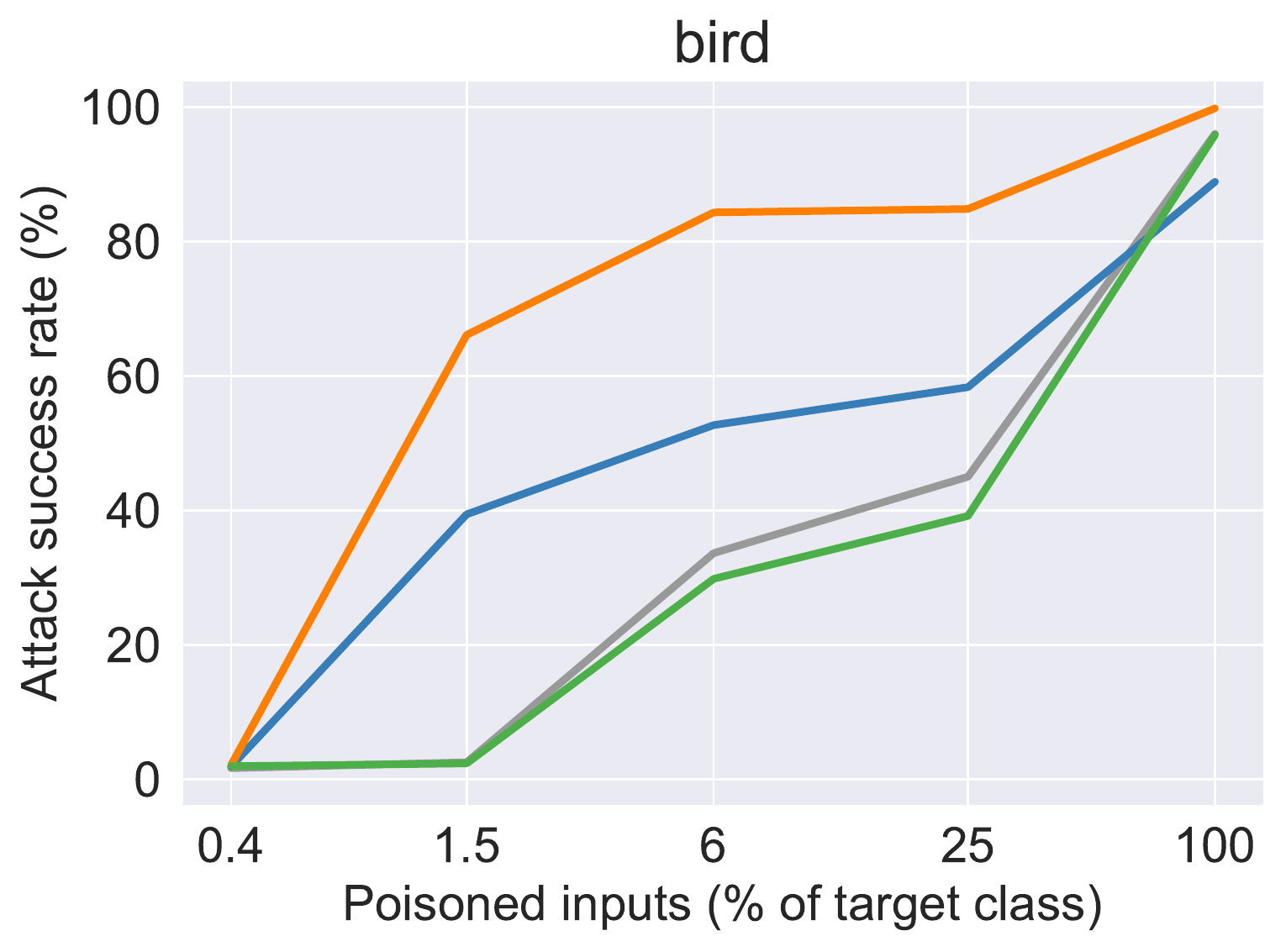}\\
    \includegraphics[align=c,width=0.31\textwidth]{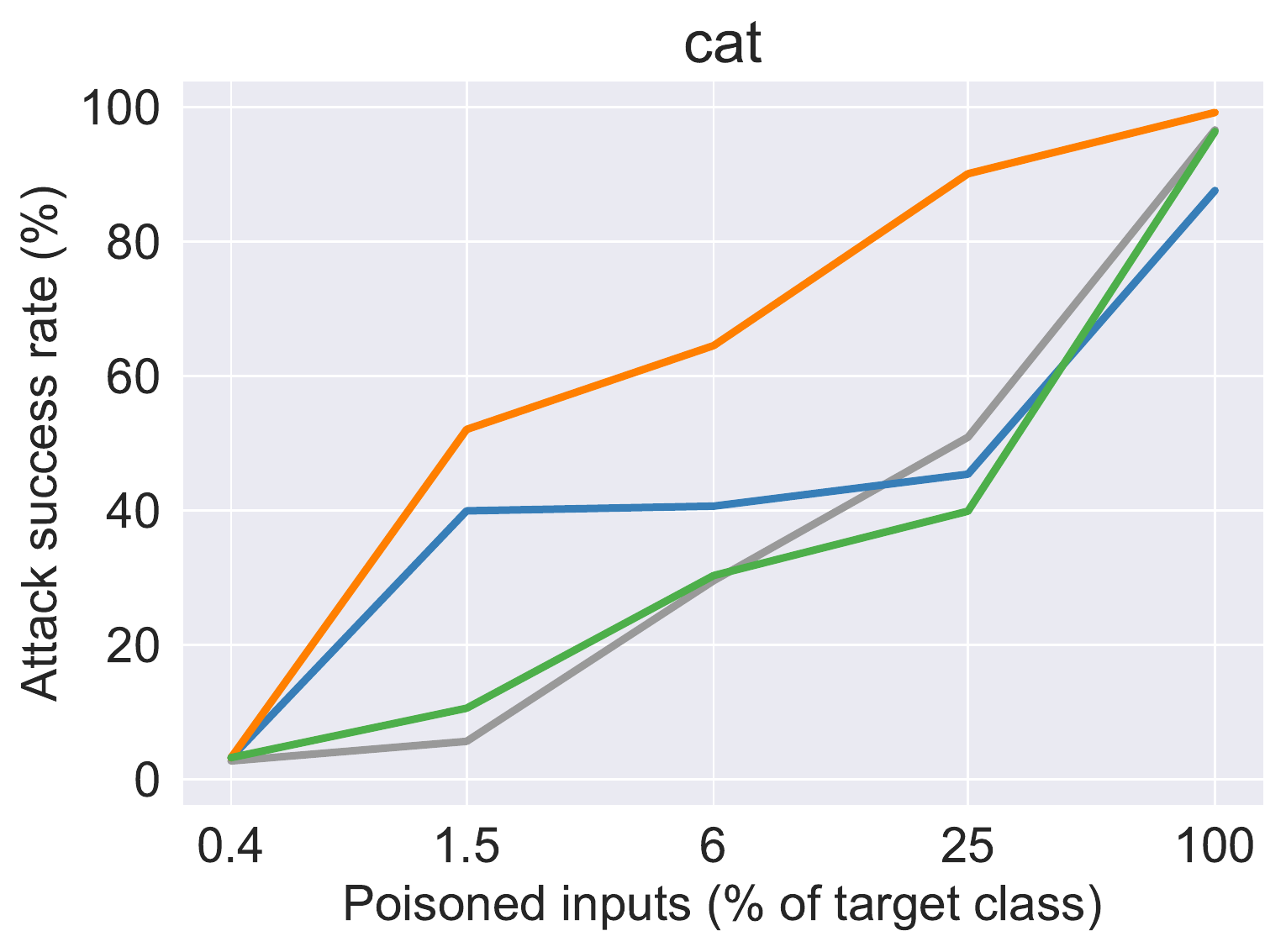}&
    \includegraphics[align=c,width=0.31\textwidth]{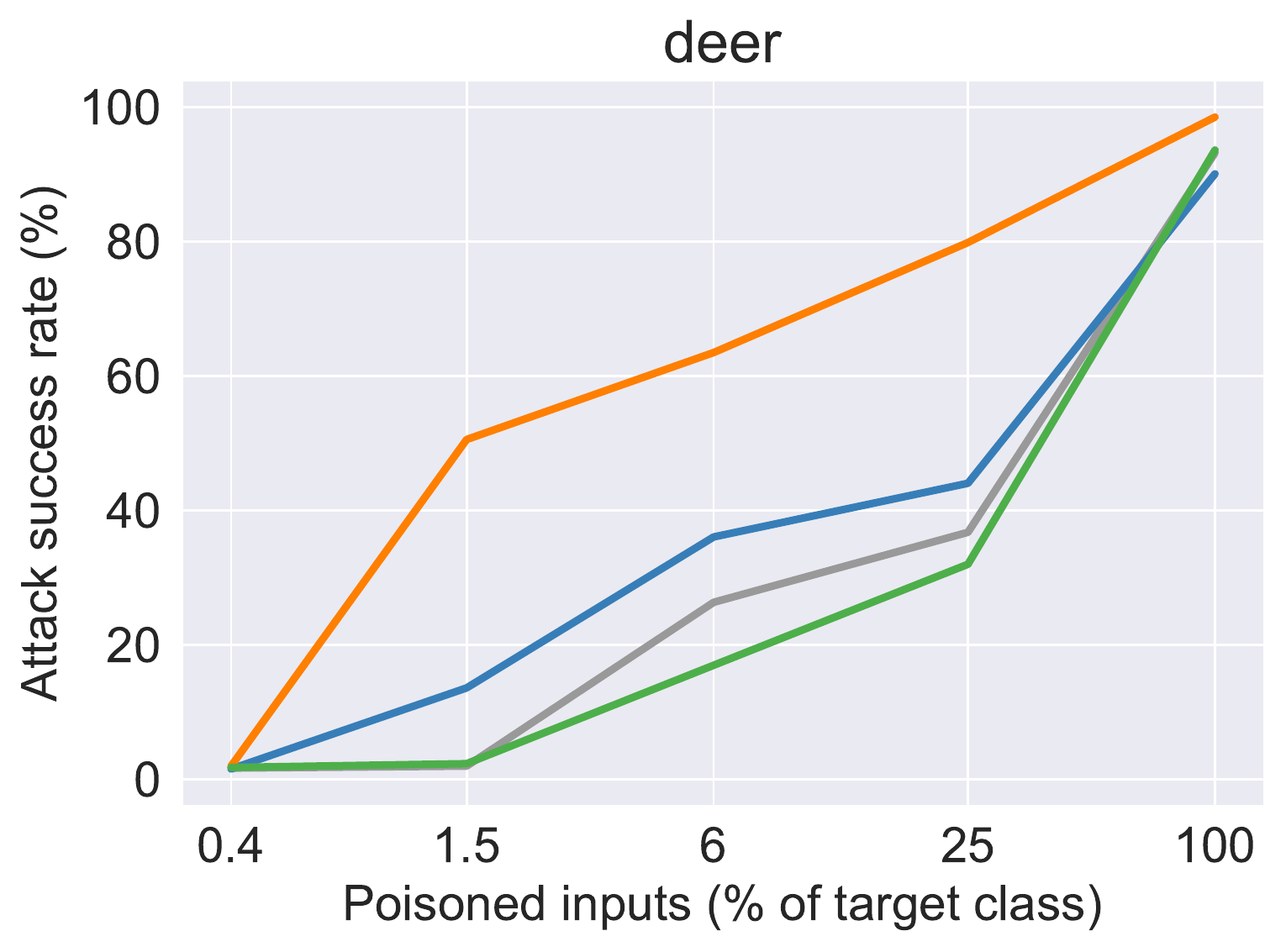}&
	\includegraphics[align=c,width=0.31\textwidth]{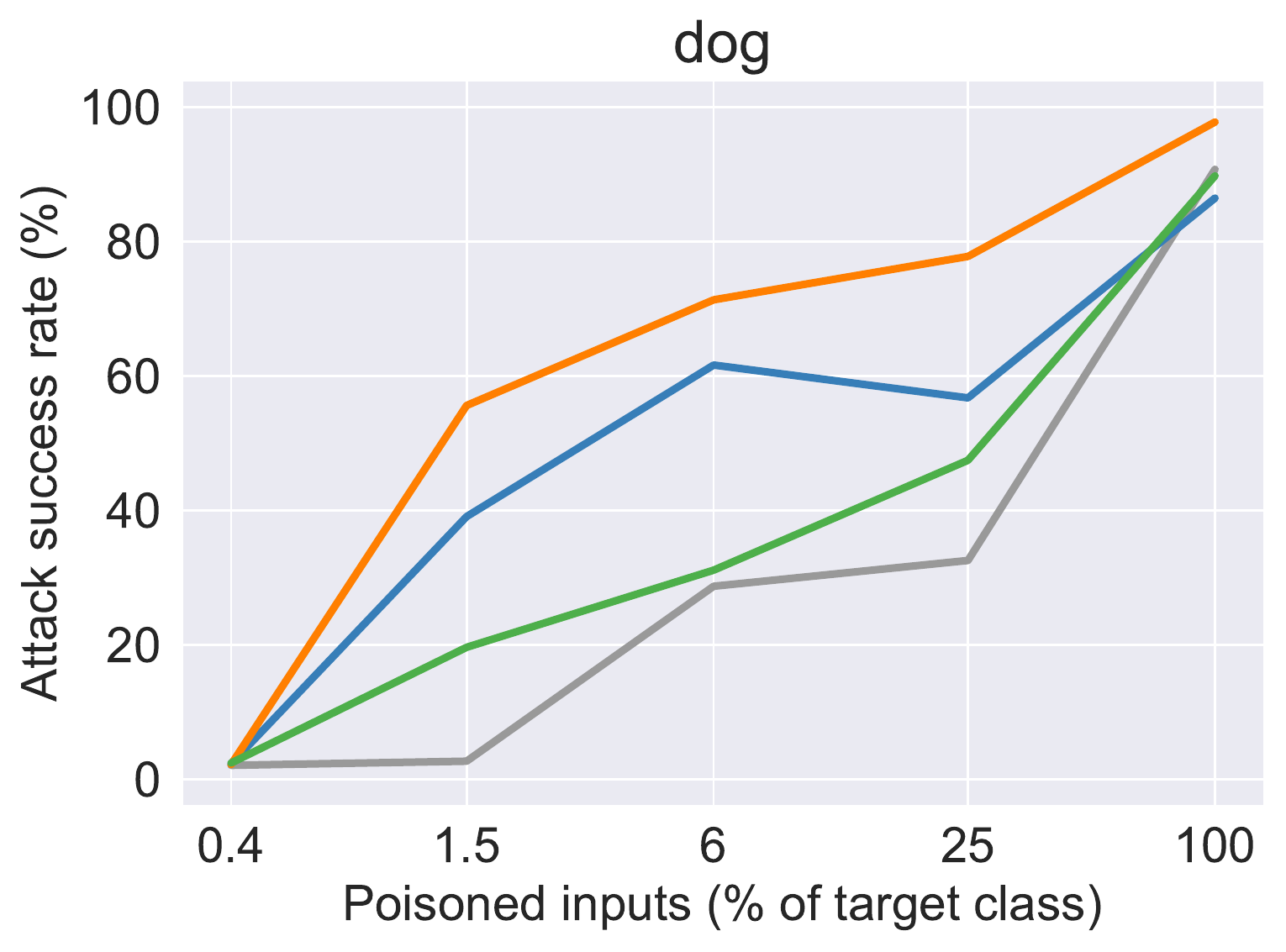}\\
    \includegraphics[align=c,width=0.31\textwidth]{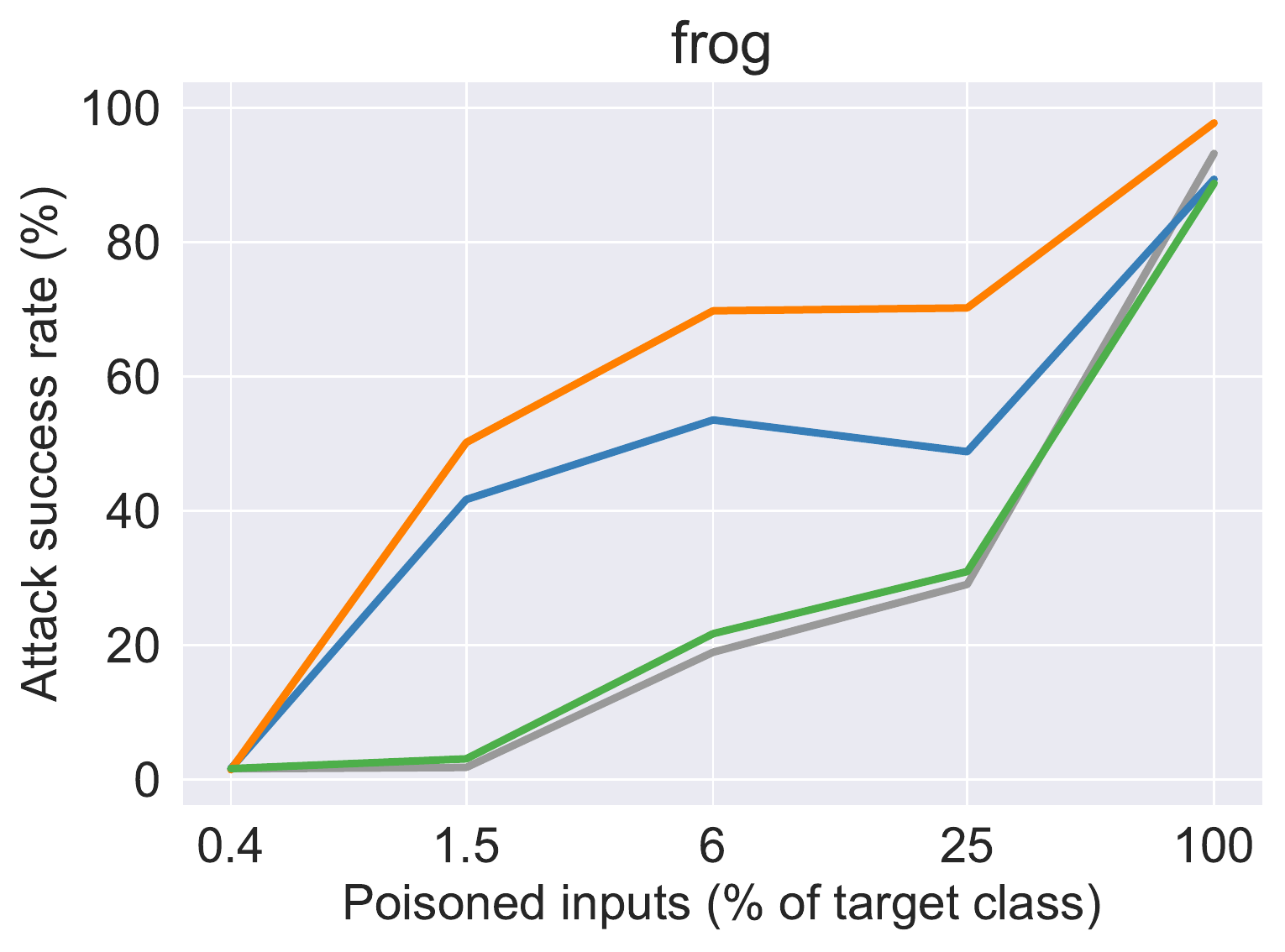}&
    \includegraphics[align=c,width=0.31\textwidth]{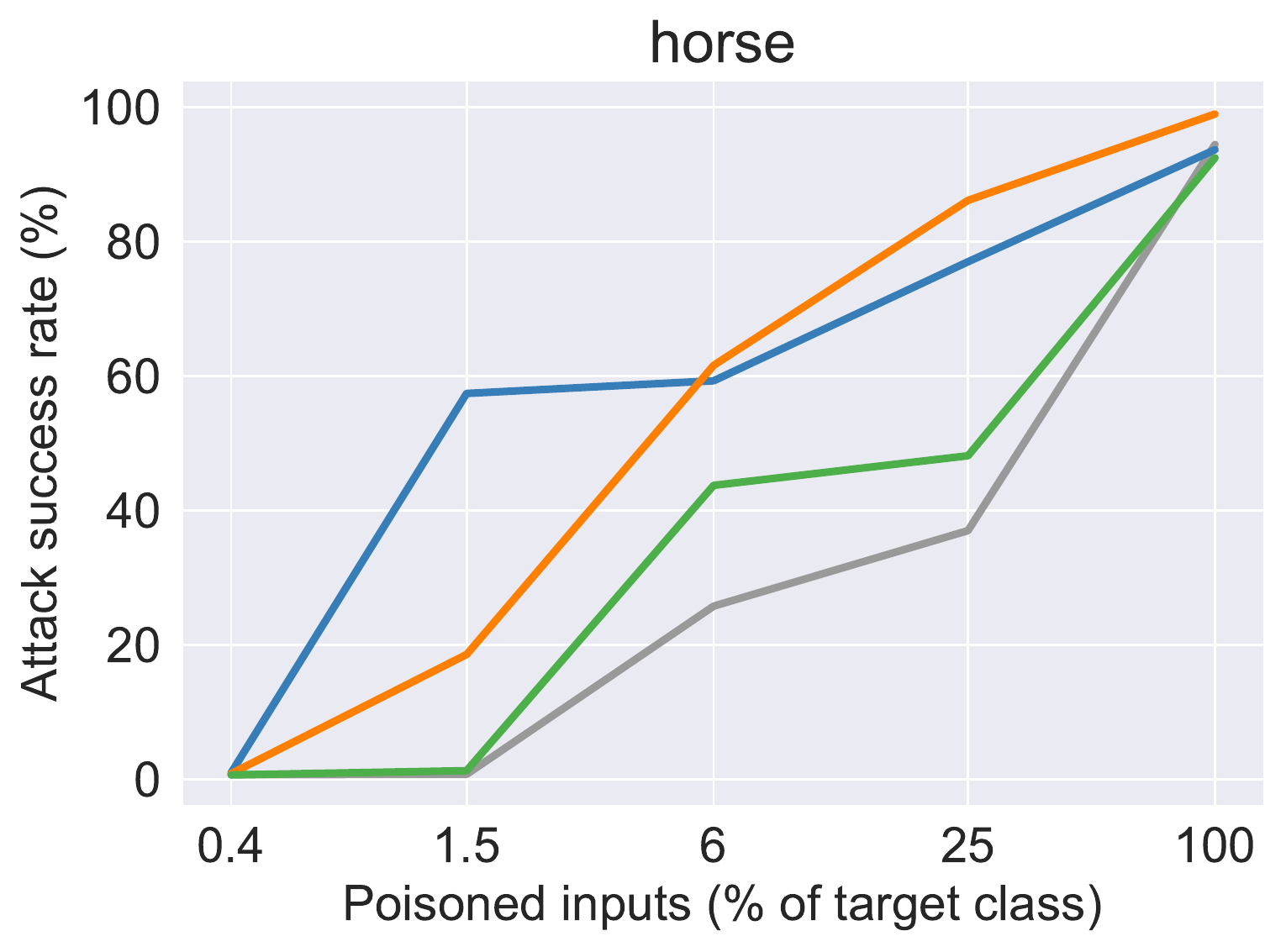}&
	\includegraphics[align=c,width=0.31\textwidth]{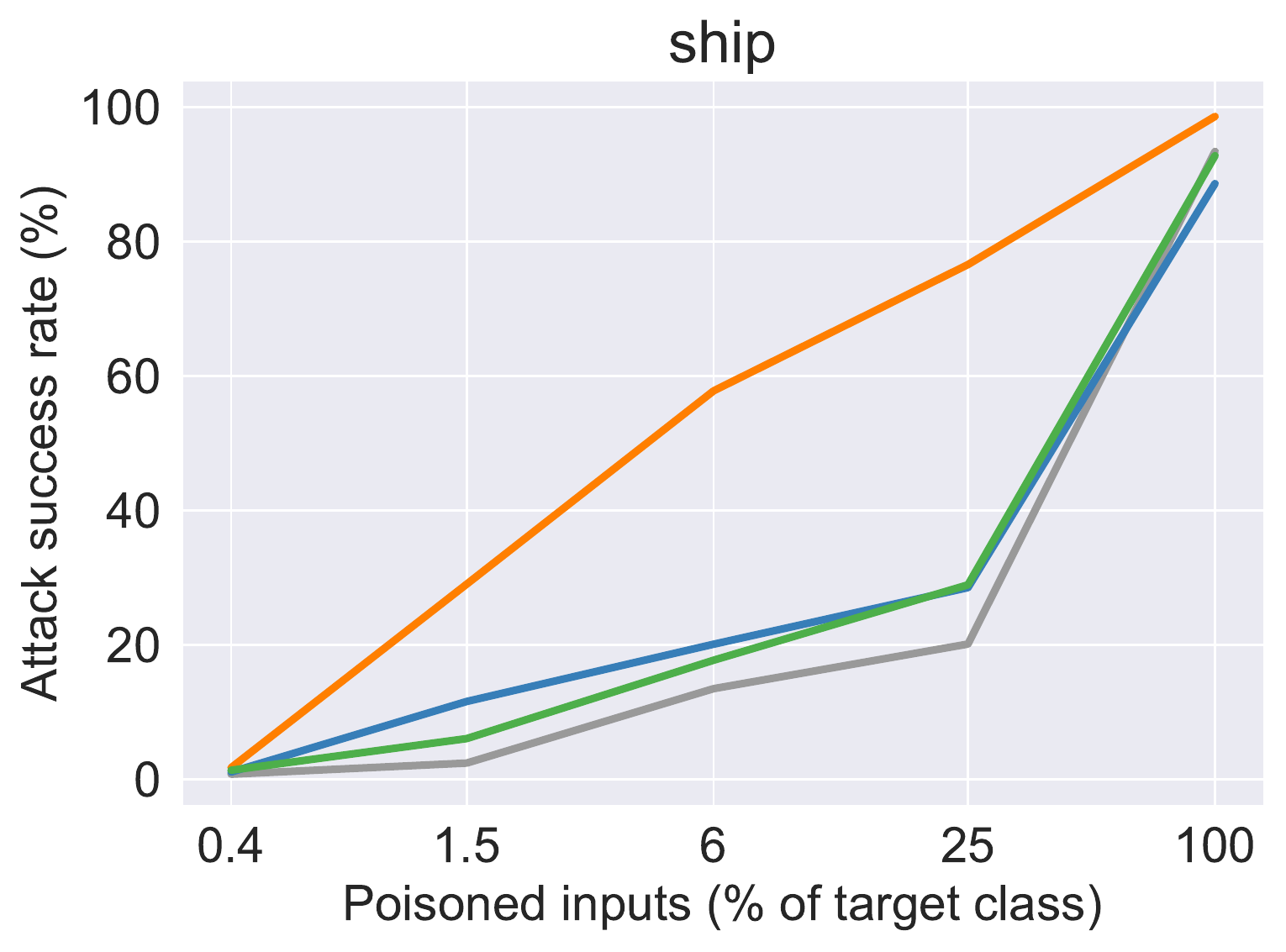}\\
    \includegraphics[align=c,width=0.31\textwidth]{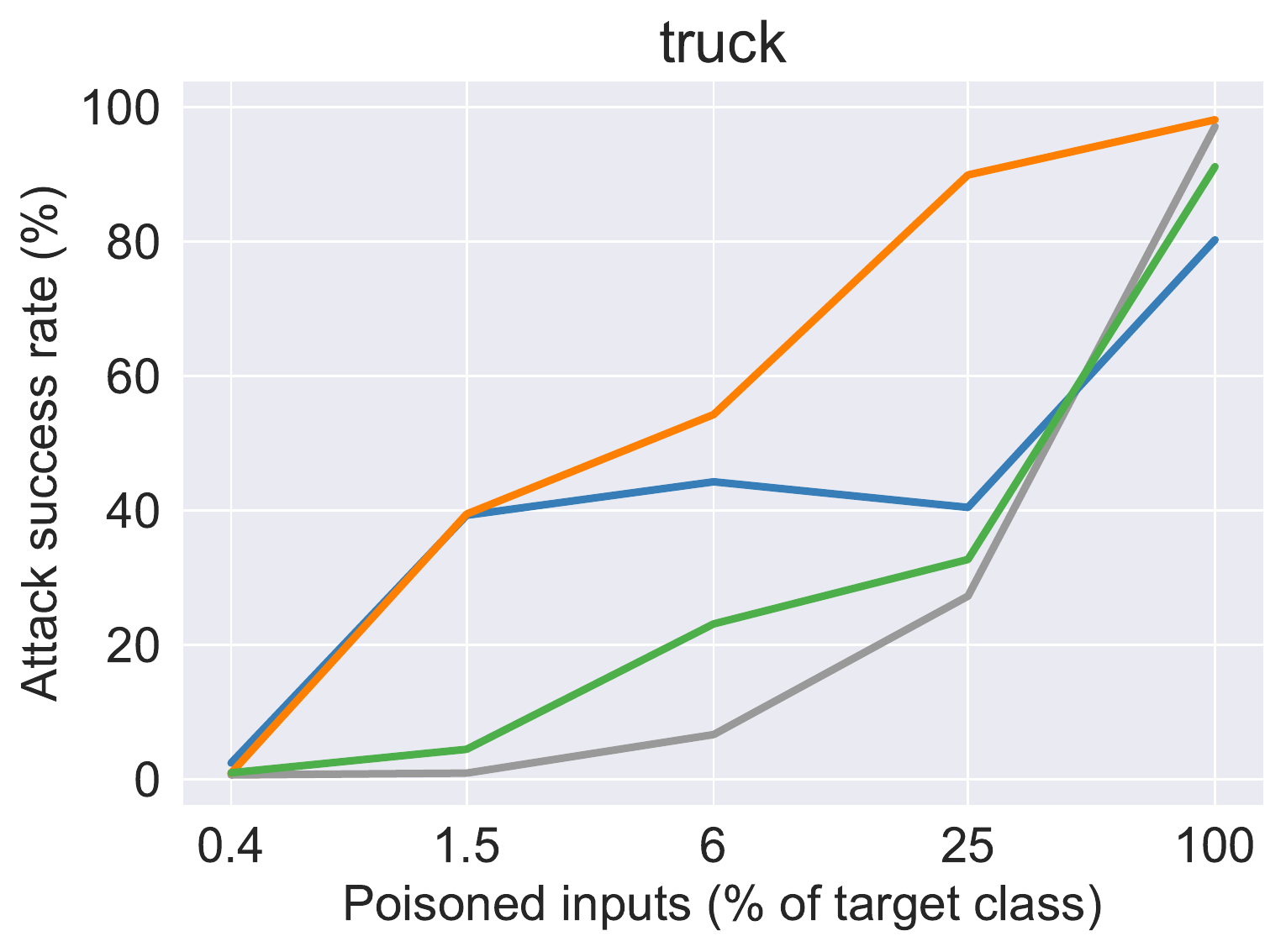}&
	\includegraphics[align=c,width=0.2\textwidth]{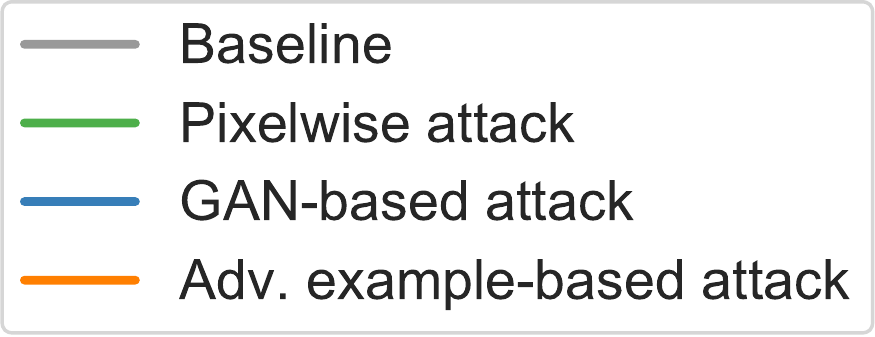}
    \end{tabular}
\end{figure}

The attack success rate is presented in Figure~\ref{fig:pixelwise}.
We find that while the attacks are generally weaker than the GAN-based
interpolation, they still outperform the naive baseline.
A per-class comparison can be found in Figure~\ref{fig:per_class}.

\paragraph{Weaker adversarial perturbations.}
\label{app:black}
Here, we consider an adversary that has only approximate knowledge of the model
architecture and dataset used. 
Specifically, the adversary will target an (adversarially trained) VGG-style
model~\citep{simonyan2015very} which is trained on 50\thinspace 000 randomly selected
images from the CINIC-10 dataset~\citep{darlow2018cinic}.
We choose this architecture since it is perhaps the most dissimilar to the
ResNet architecture used by our model (adversarial examples tend to transfer
less between VGG-style models and ResNets).
The attack is otherwise unchanged.

The success rate of this attack is presented in Figure~\ref{fig:black_box}.
We find that this attack, while less powerful than the white-box attack, can
still result in effective attacks and significantly outperform the
other baselines.

\clearpage
\section{Omitted Figures}
\label{app:omitted}

\begin{figure}[thb]
	\centering
    \begin{subfigure}[b]{0.49\textwidth}
	\centering
    \begin{tikzpicture}
        \node at (0,0) {\includegraphics[width=0.8\textwidth]
            {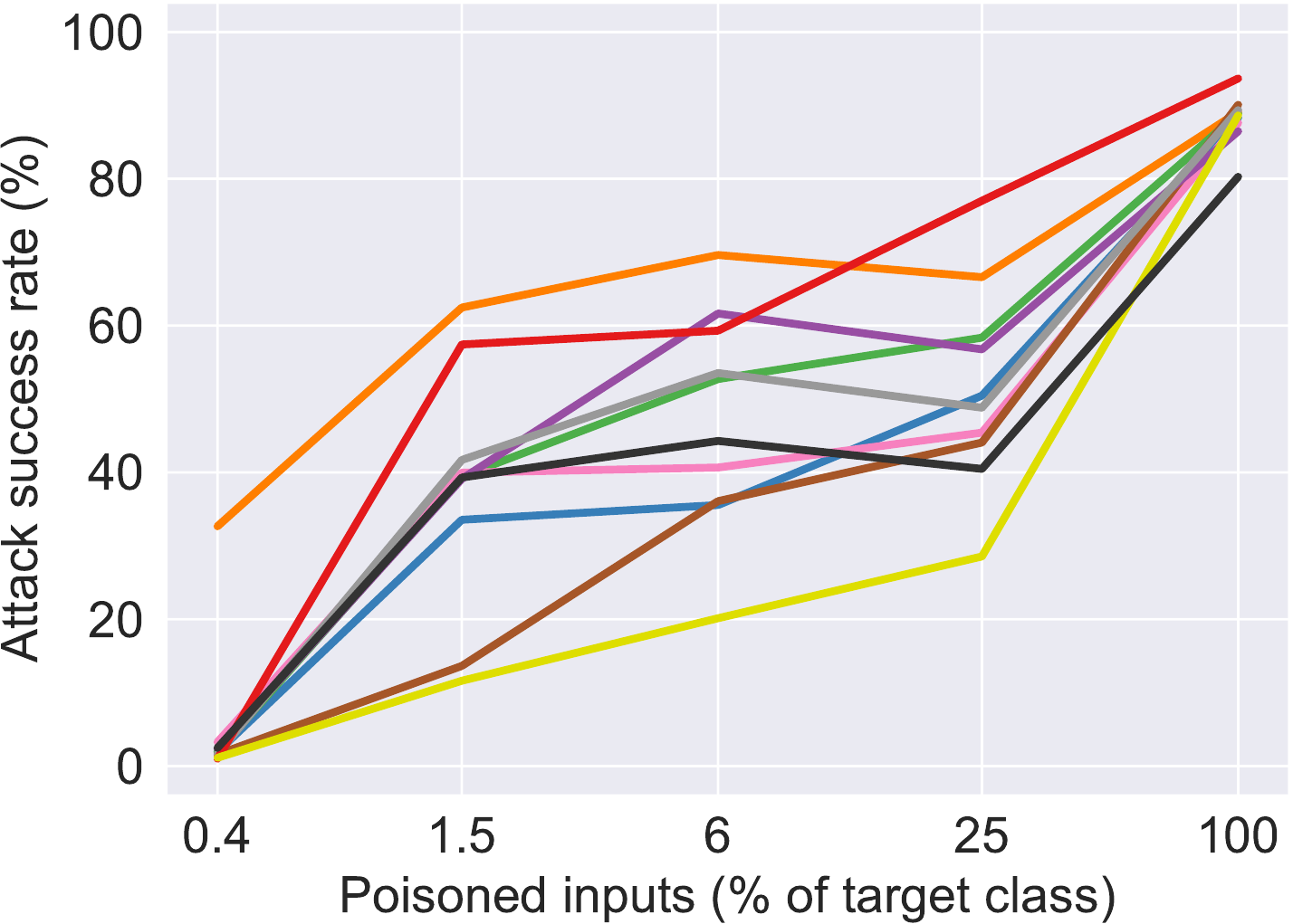}};
        \node at (3.45,0) {\phantom{\includegraphics[width=0.22\textwidth]
            {figures/all_classes_legend.pdf}}};
    \end{tikzpicture}
	\caption{GAN-based interpolation}
	\label{fig:adv_attack}
	\end{subfigure}
    \hfill
	\begin{subfigure}[b]{0.49\textwidth}
	\centering
    \begin{tikzpicture}
        \node at (0,0) {\includegraphics[width=0.8\textwidth]
            {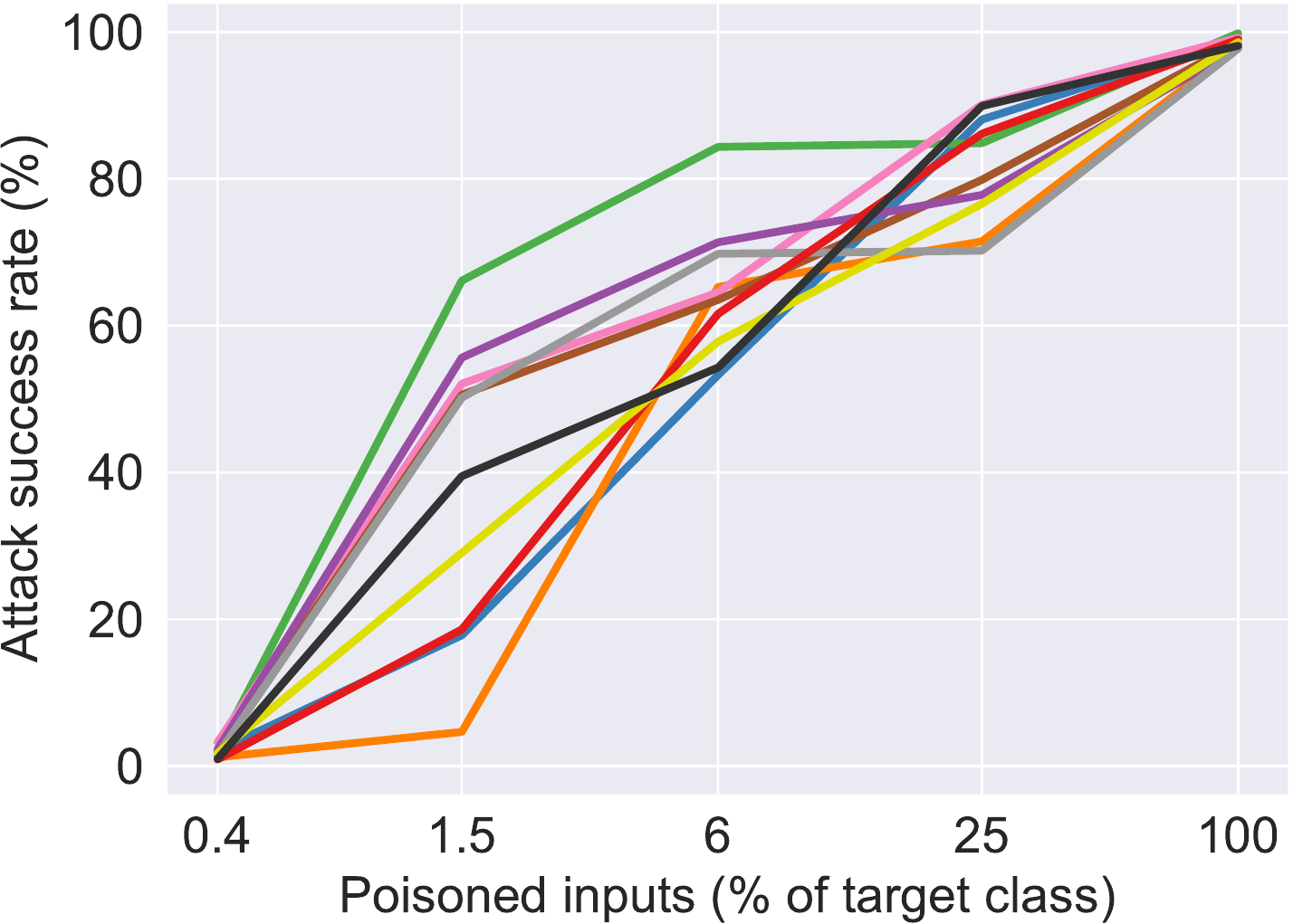}};
        \node at (3.45,0) {\includegraphics[width=0.22\textwidth]
            {figures/all_classes_legend.pdf}};
    \end{tikzpicture}
	\caption{Adversarial perturbation}
	\label{fig:gan_attack}
    \end{subfigure}
	\caption{Attack performance of our label-consistent attacks for each target
        class (using fully visible trigger and no data augmentation). Both
        attacks achieve significantly higher success rates compared
        to the baseline attack (Figure~\ref{fig:baseline}). 
        A per-class comparison can be found in
        Appendix Figure~\ref{fig:per_class}.
        (a) GAN-based interpolation with $\tau=0.2$.
        (b) Adversarial perturbations $\ell_2$-bounded by $\eps=300$
        (pixel values in $[0, 255]$).}
	\label{fig:basic_attack}
\end{figure}

\begin{figure}[htp]
	\centering
    \begin{subfigure}[b]{0.49\textwidth}
        \centering
    \begin{tikzpicture}
        \node at (0,0) {\includegraphics[width=0.8\textwidth]
                            {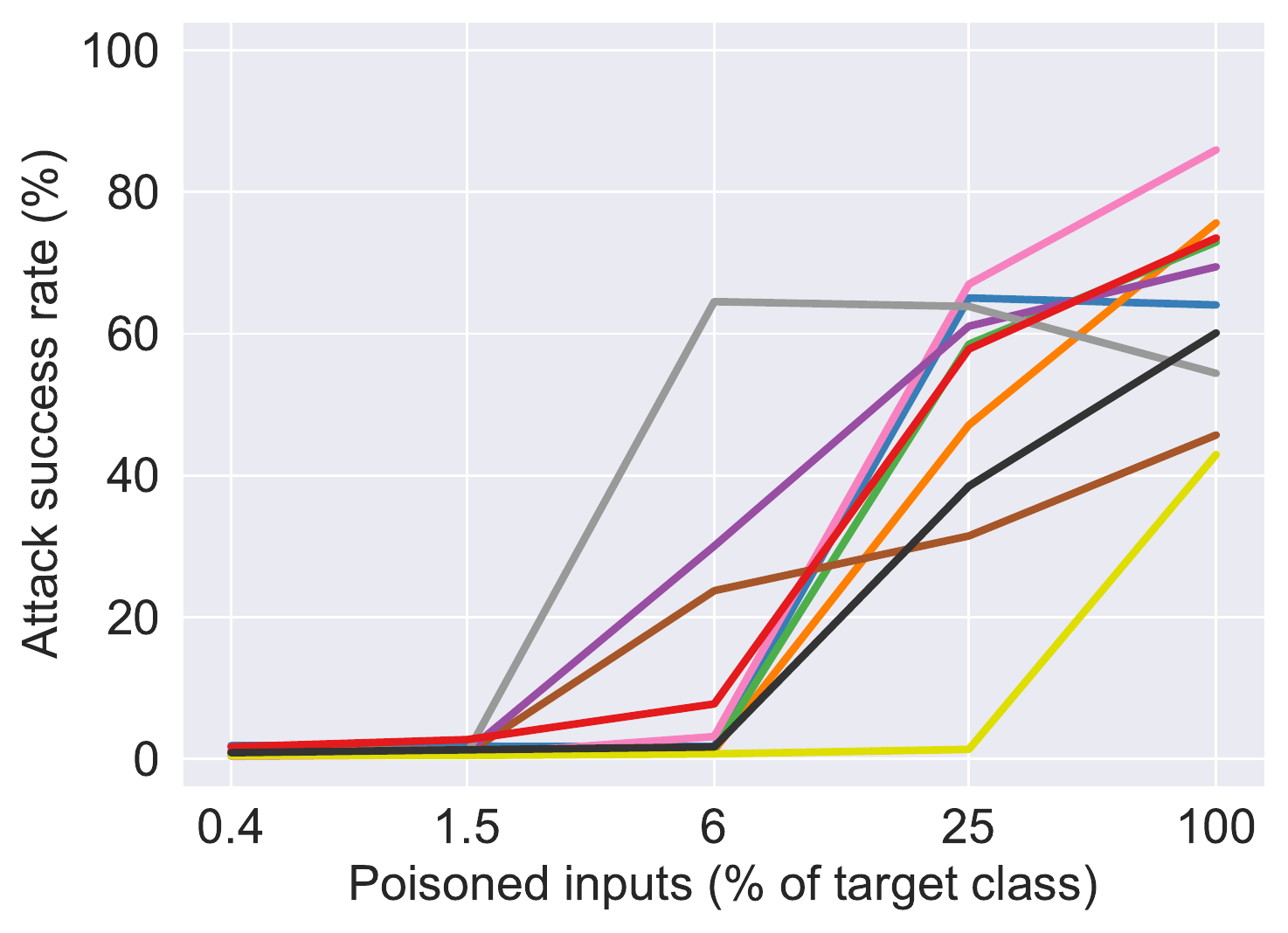}};
    \end{tikzpicture}
    \caption{Reduced visibility inference}
	\label{fig:gan16small}
    \end{subfigure}
    \begin{subfigure}[b]{0.49\textwidth}
        \centering
    \begin{tikzpicture}
        \node at (0,0) {\includegraphics[width=0.8\textwidth]
                            {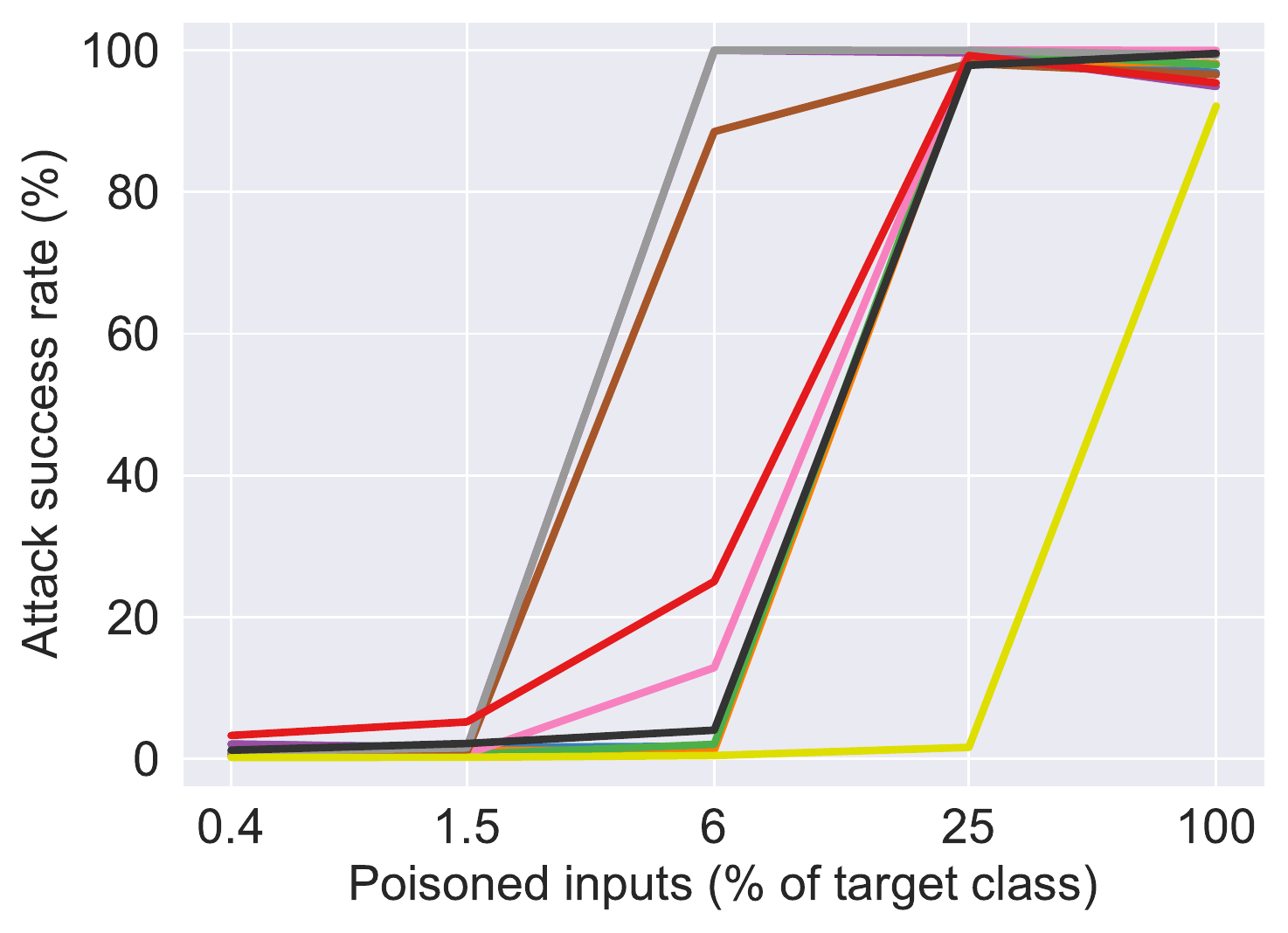}};
        \node at (3.3,0) {\includegraphics[width=0.21\textwidth]
                            {figures/all_classes_legend.pdf}};
    \end{tikzpicture}
    \caption{Full visibility inference}
	\label{fig:gan16big}
    \end{subfigure}
    \caption{Performance of the GAN-based interpolation attack with a reduced
        visibility trigger (amplitude 16). (a) Evaluating with a reduced
        amplitude pattern during inference. (b) Using the full visibility
        pattern during inference greatly increases the attack success rate.}
	\label{fig:gan16}
\end{figure}

\begin{figure}[htp]
	\centering
    \begin{subfigure}[b]{0.49\textwidth}
        \centering
    \begin{tikzpicture}
        \node at (0,0) {\includegraphics[width=0.8\textwidth]
                            {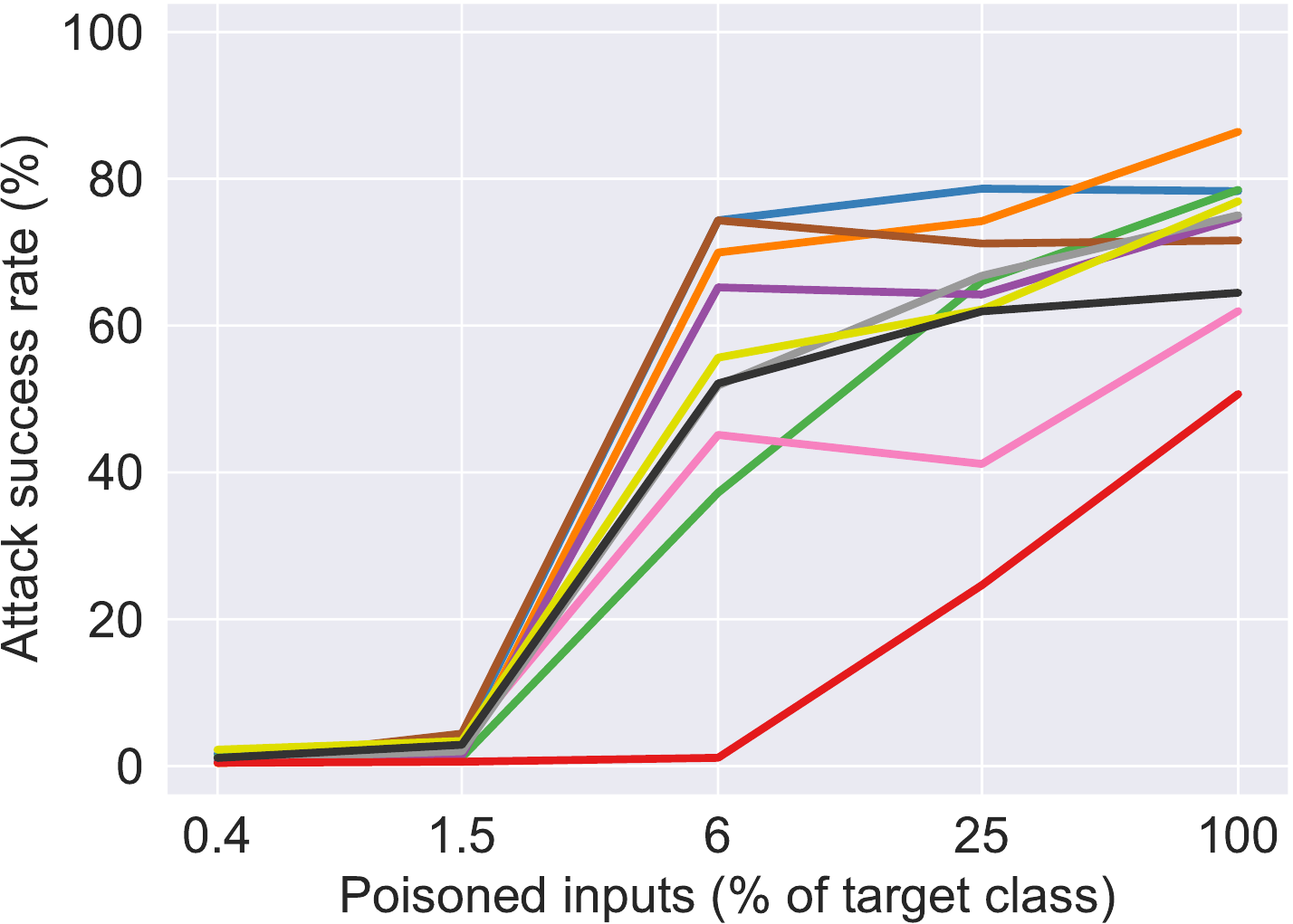}};
    \end{tikzpicture}
    \caption{GAN-based interpolation, reduced visibility}
    \end{subfigure}
    \begin{subfigure}[b]{0.49\textwidth}
        \centering
    \begin{tikzpicture}
        \node at (0,0) {\includegraphics[width=0.8\textwidth]
                            {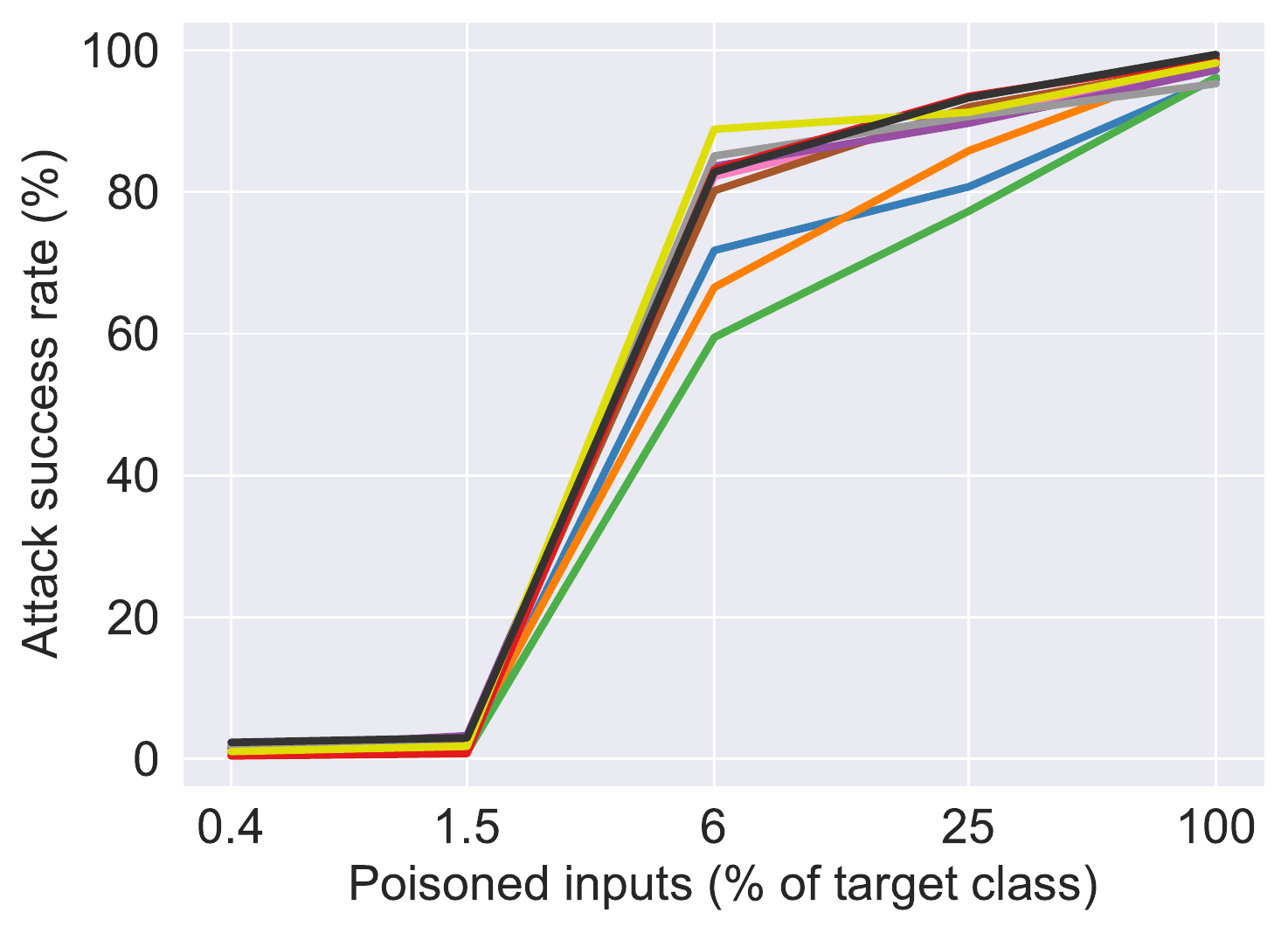}};
    \end{tikzpicture}
    \caption{Adversarial perturbation, reduced visibility}
    \end{subfigure}
    \begin{subfigure}[b]{0.49\textwidth}
        \centering
    \begin{tikzpicture}
        \node at (0,0) {\includegraphics[width=0.8\textwidth]
                            {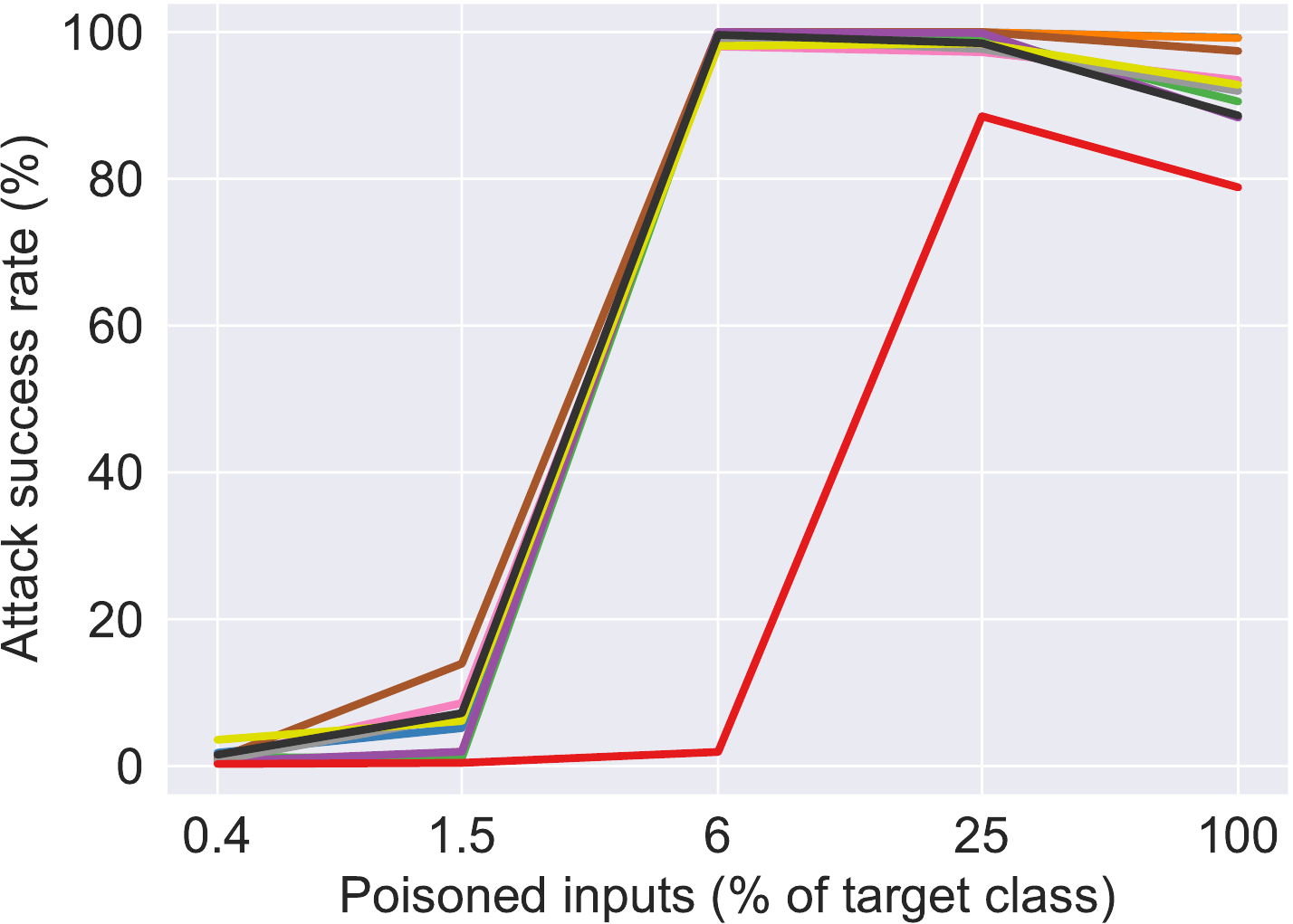}};
    \end{tikzpicture}
    \caption{GAN-based interpolation, full visibility}
    \end{subfigure}
    \begin{subfigure}[b]{0.49\textwidth}
        \centering
    \begin{tikzpicture}
        \node at (0,0) {\includegraphics[width=0.8\textwidth]
                            {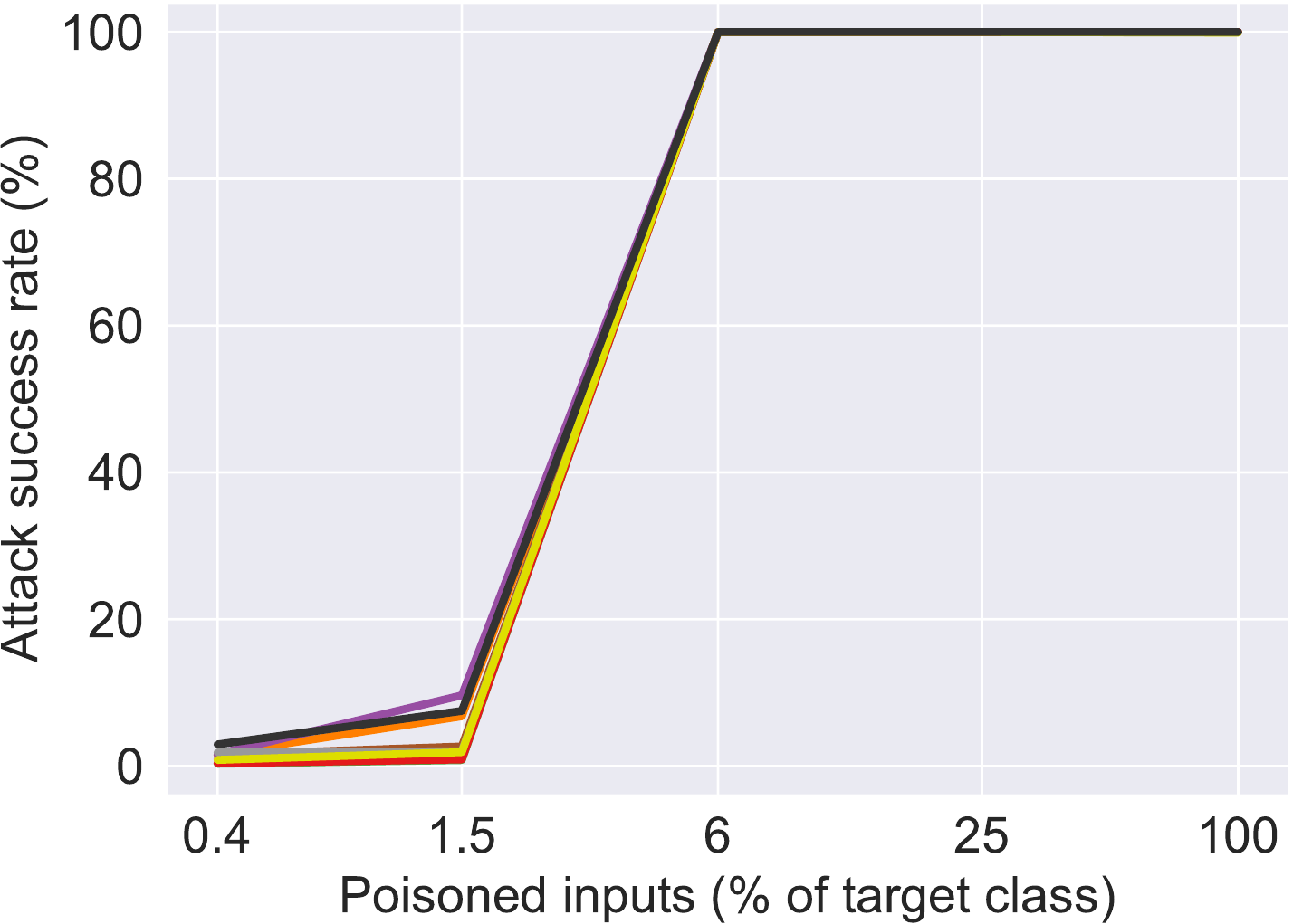}};
        \node at (2.5,0) {\includegraphics[width=0.23\textwidth]
                            {figures/all_classes_legend.pdf}};
    \end{tikzpicture}
    \caption{Adversarial perturbation, full visibility}
    \end{subfigure}
	\caption{Performance of our label-consistent attacks when using a trigger
        amplitude of 32 for GAN-based interpolation and adversarial
        perturbation-based attacks.
        Top row: Using the trigger used during training (amplitude 32) for
        inference.
        Bottom row: Using a full-amplitude trigger (255) during inference.
        All attacks are performed in the presence of data augmentation (using
        the four-corner pattern).
    }
	\label{fig:attack32}
\end{figure}

\clearpage
\begin{figure}
	\caption{Plotting the training loss of poisoned examples with and without
        the trigger and compare it to the average training loss for each target
        class (similar to Figure~\ref{fig:loss}).
        The loss of poisoned inputs without the trigger remains high throughout
        training, indicating that they cannot be classified correctly by the
        model without relying on the backdoor trigger.}
	\label{fig:all_losses}
	\centering
	\includegraphics[align=c,width=0.3\textwidth]{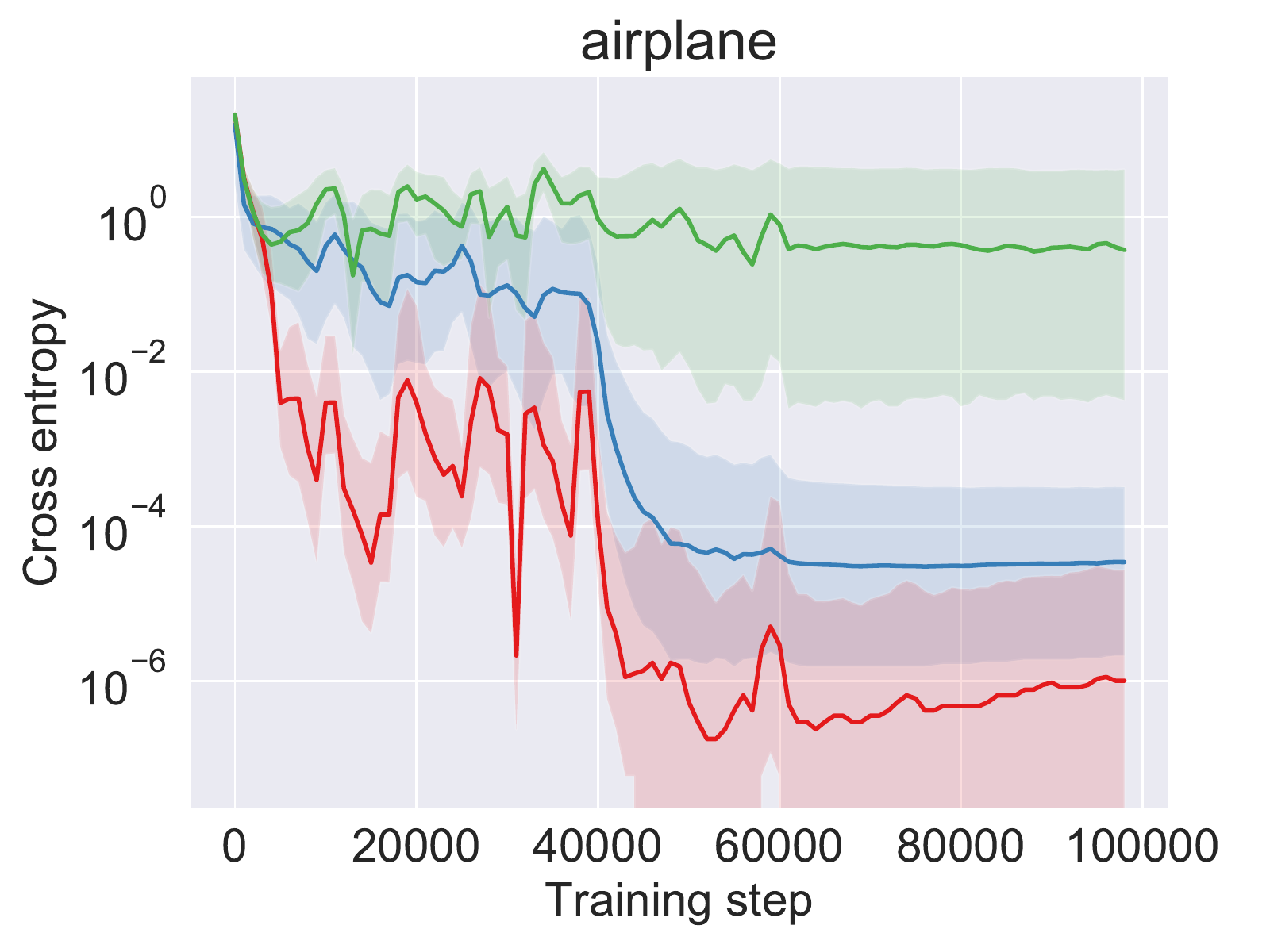}
	\includegraphics[align=c,width=0.3\textwidth]{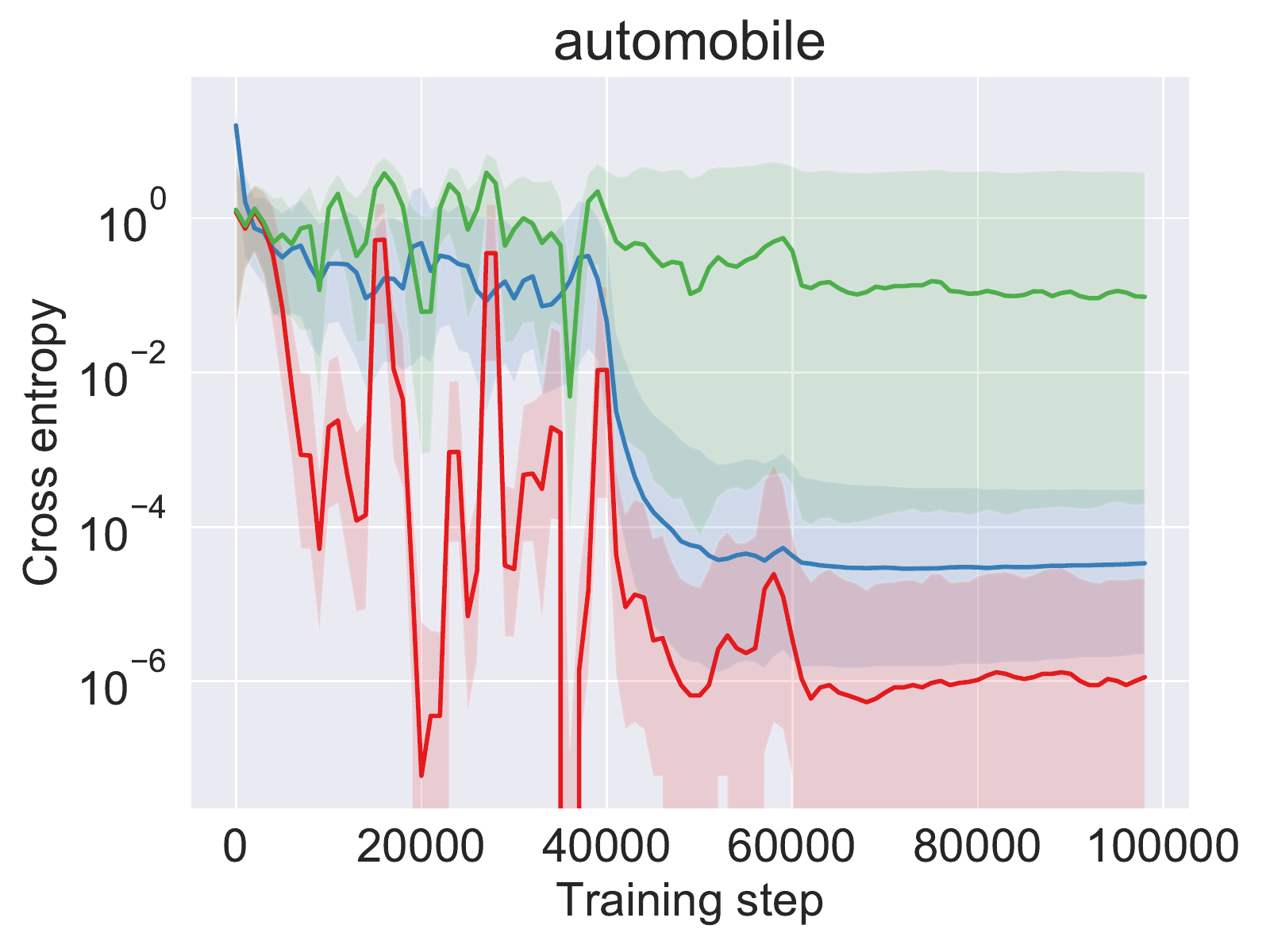}
	\includegraphics[align=c,width=0.3\textwidth]{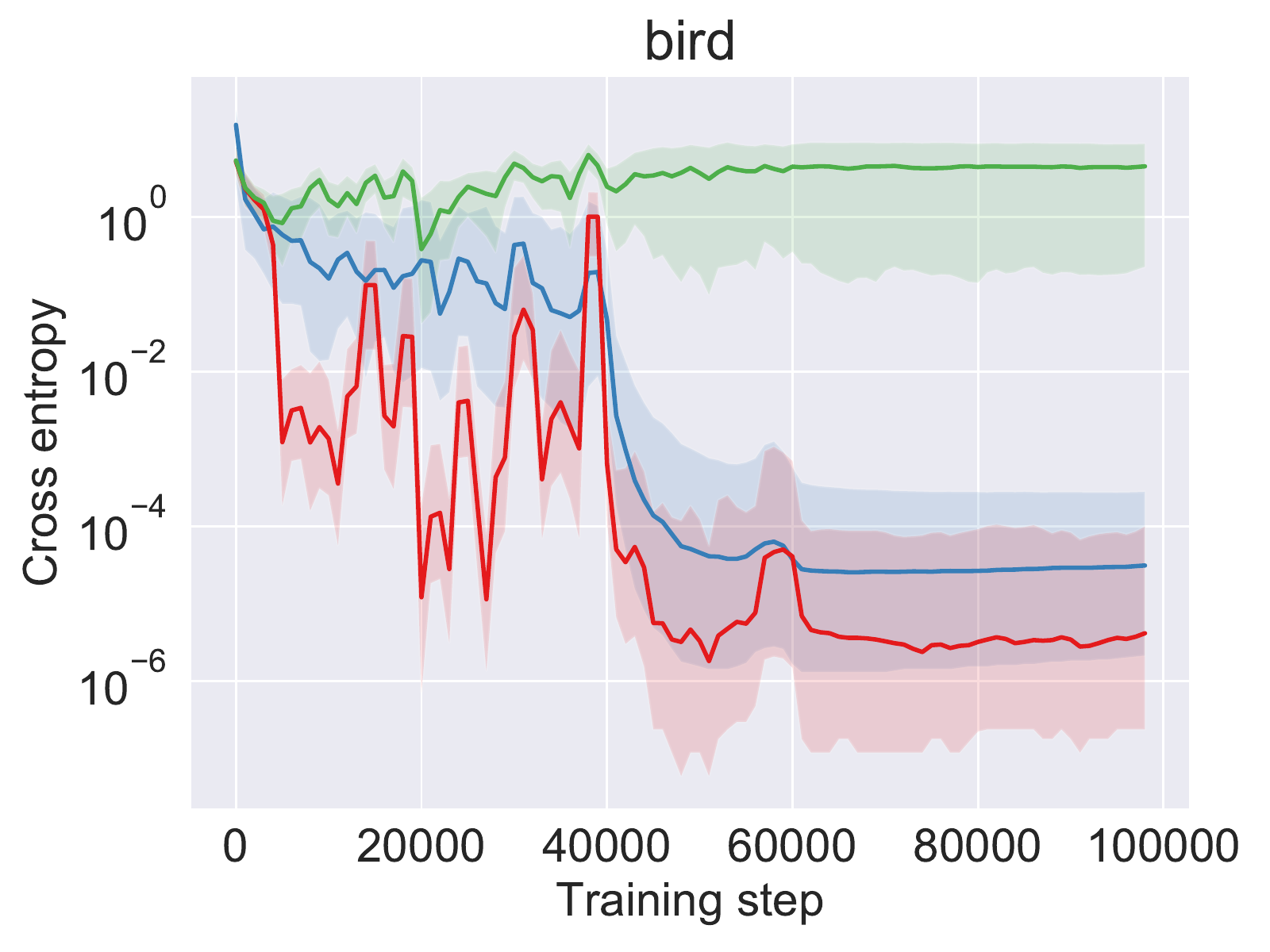}\\
	\includegraphics[align=c,width=0.3\textwidth]{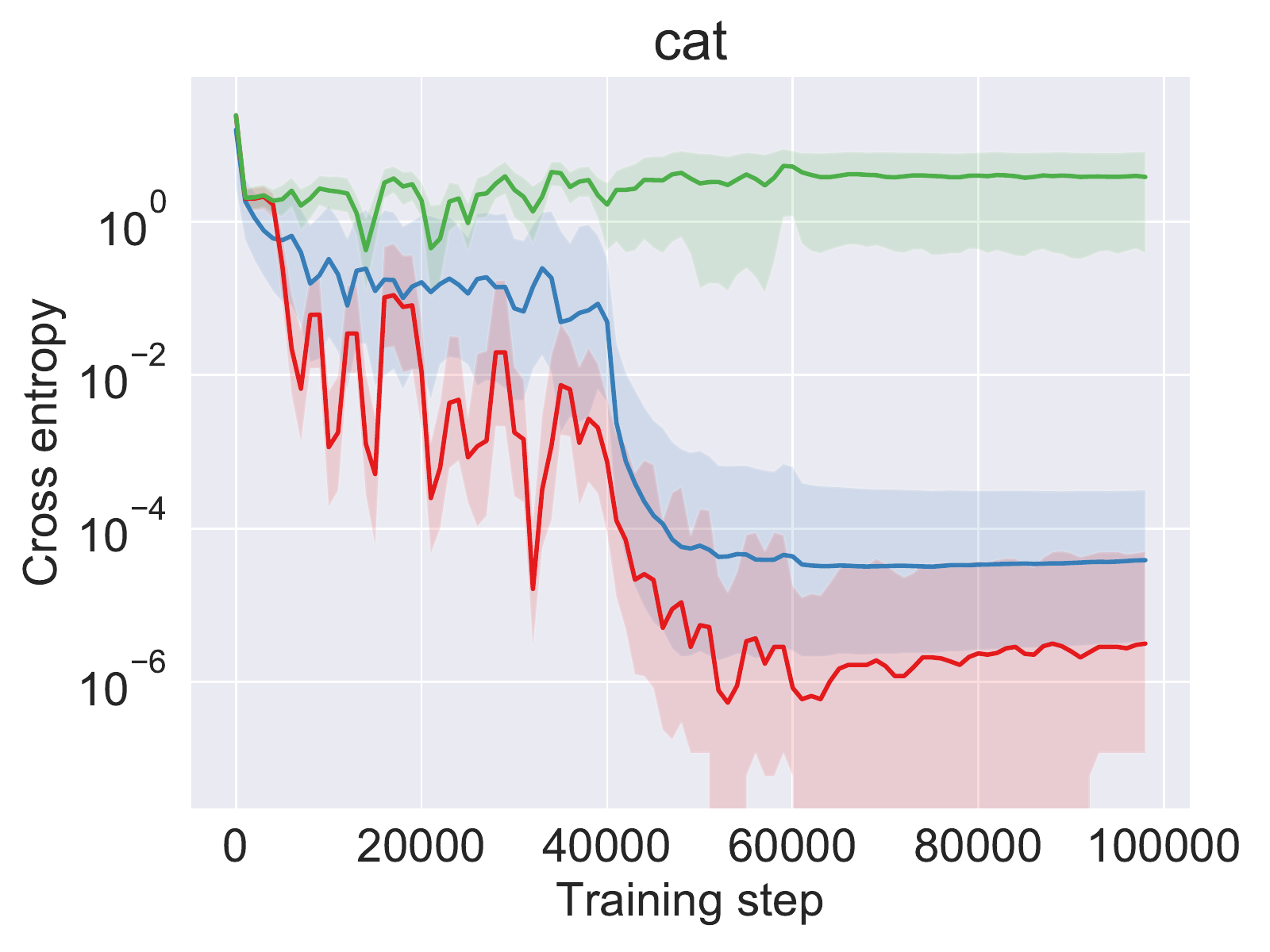}
	\includegraphics[align=c,width=0.3\textwidth]{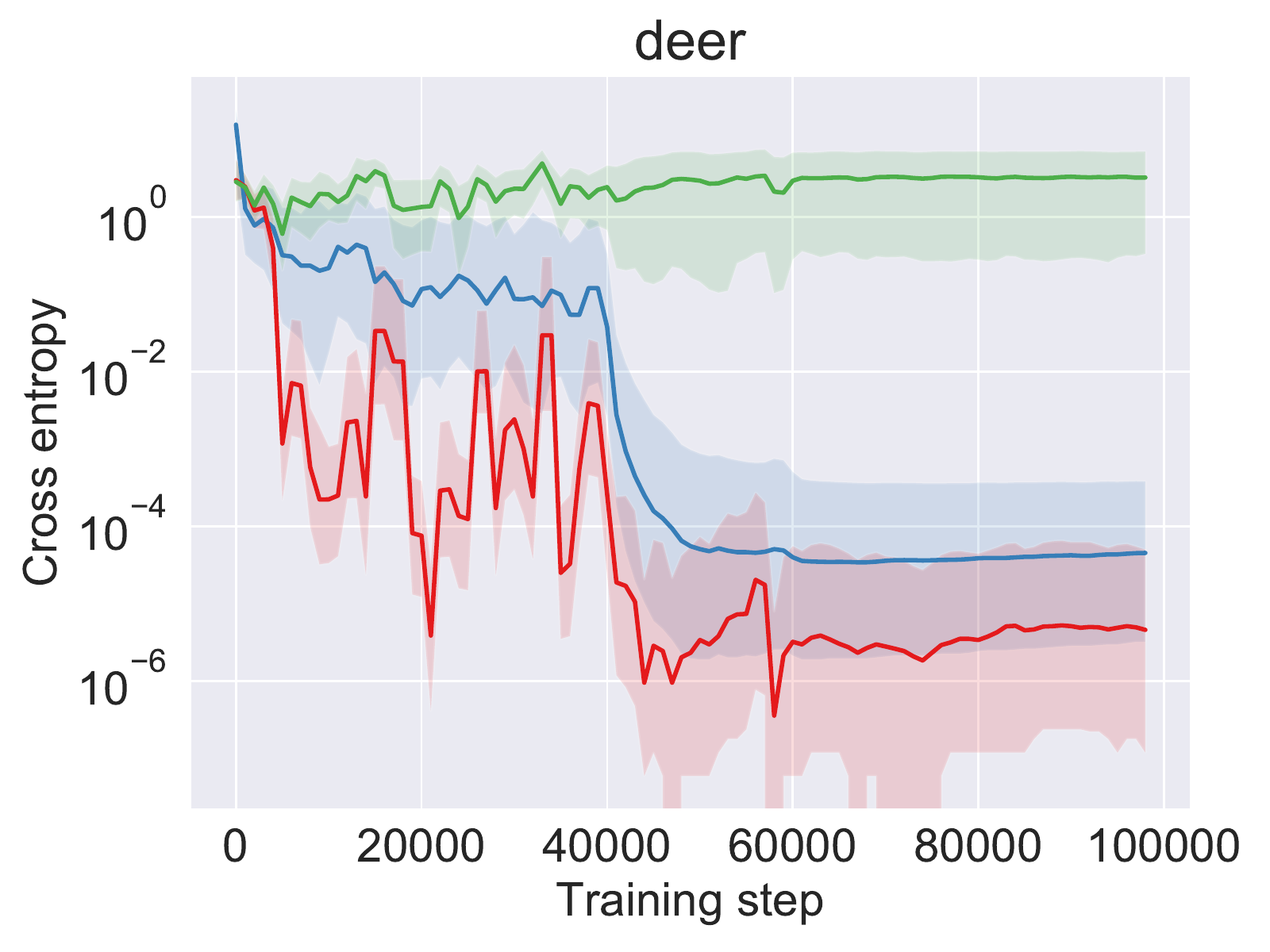}
	\includegraphics[align=c,width=0.3\textwidth]{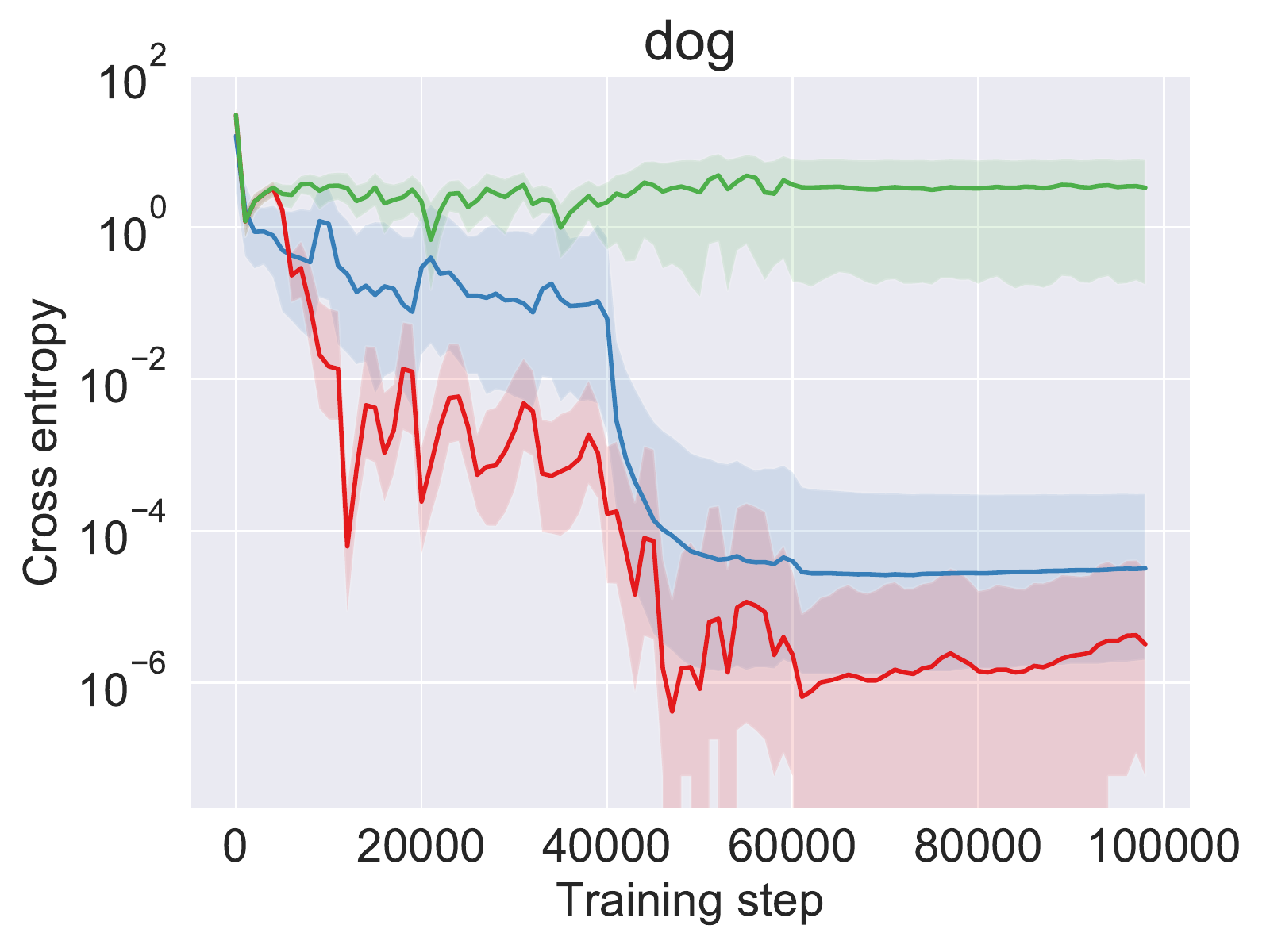}\\
	\includegraphics[align=c,width=0.3\textwidth]{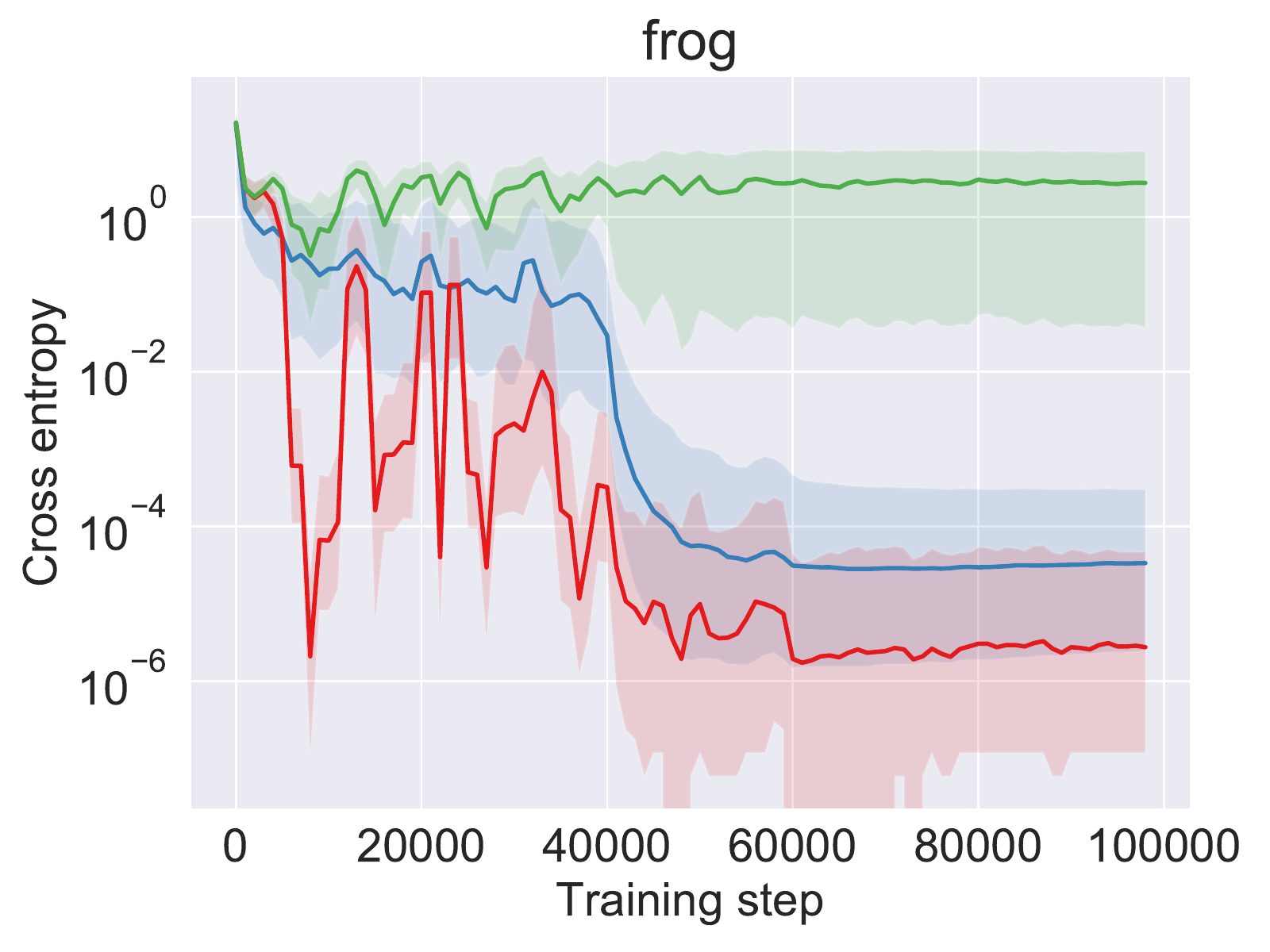}
	\includegraphics[align=c,width=0.3\textwidth]{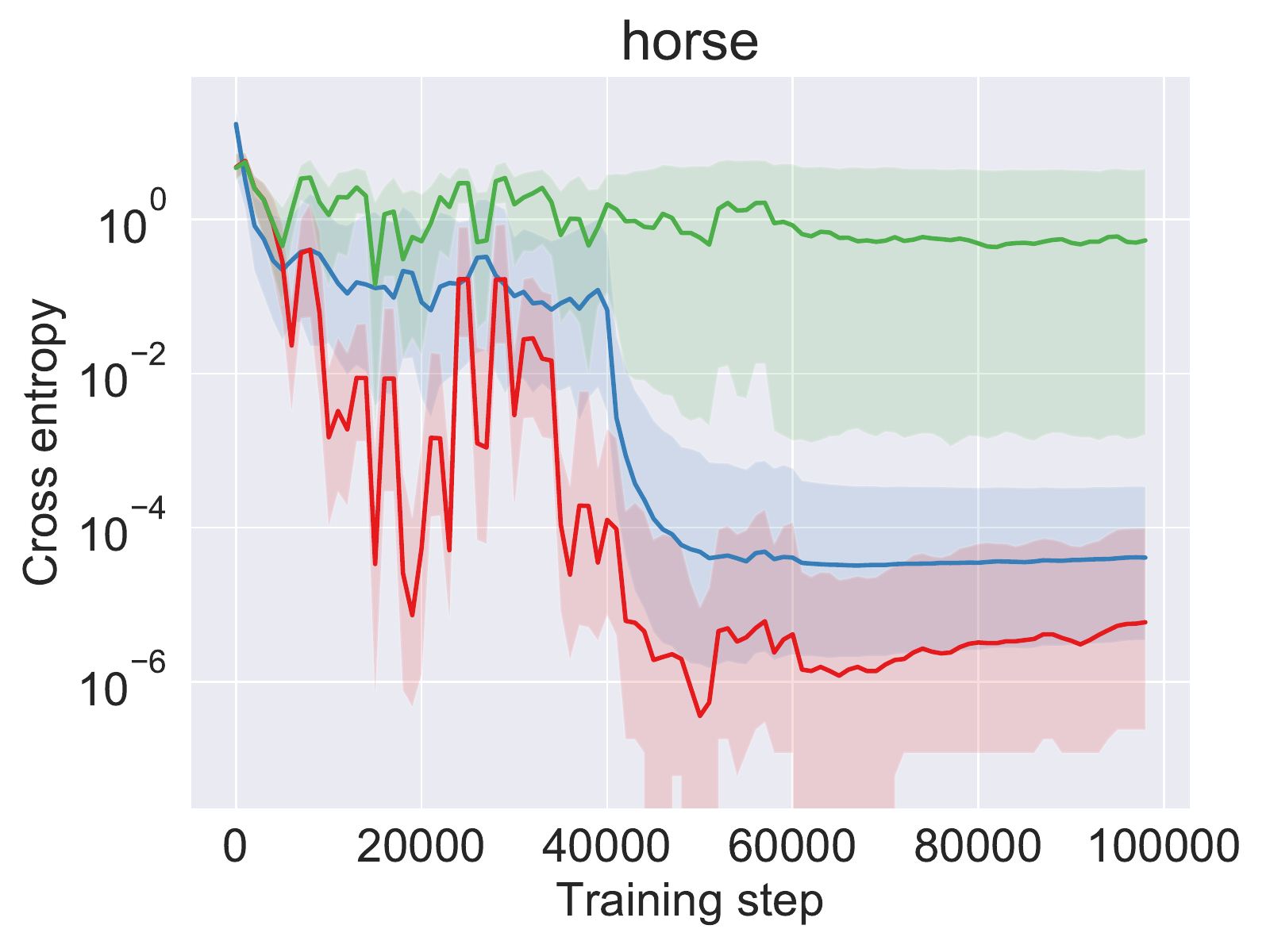}
	\includegraphics[align=c,width=0.3\textwidth]{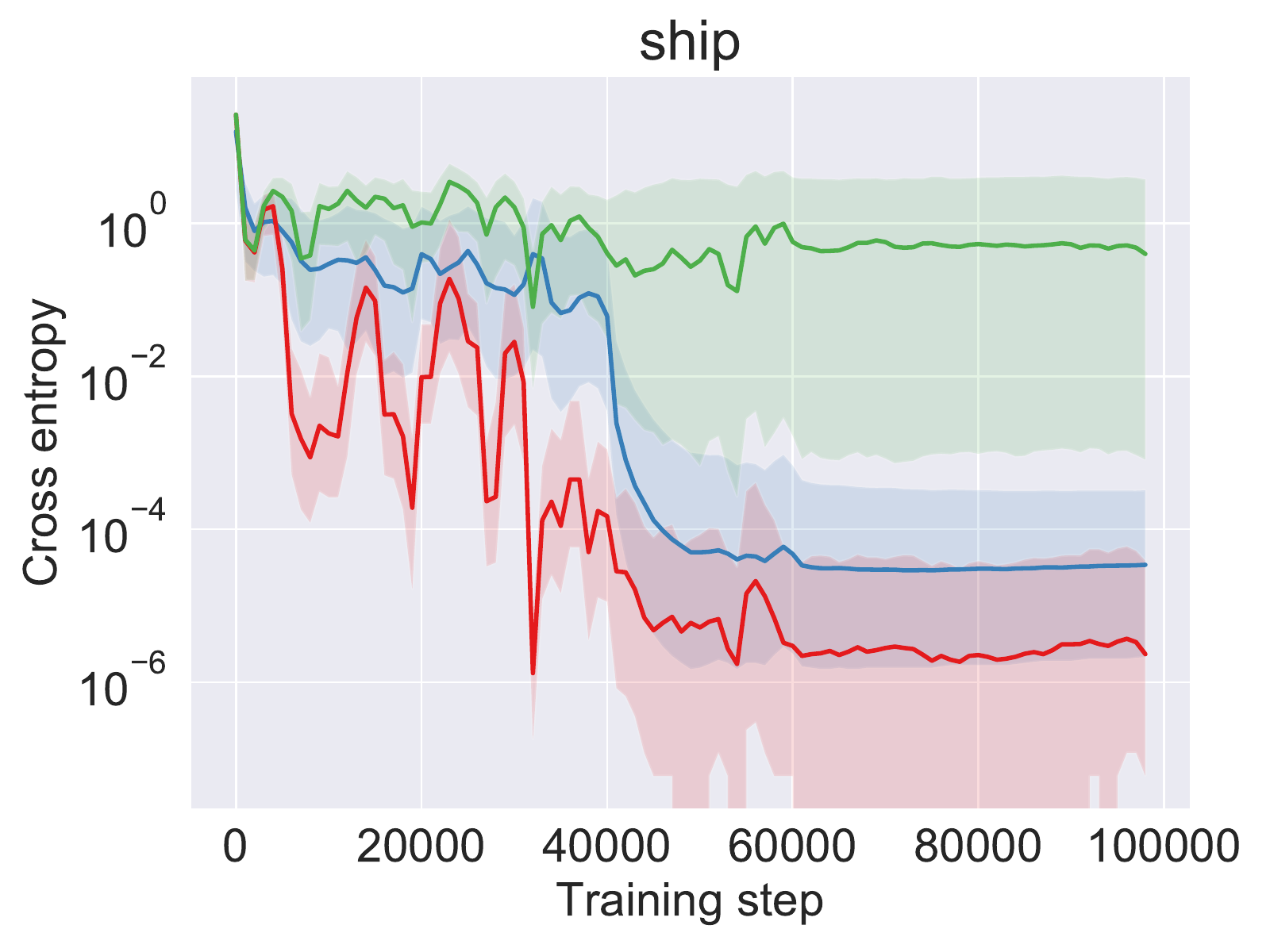}\\
	\includegraphics[align=c,width=0.3\textwidth]{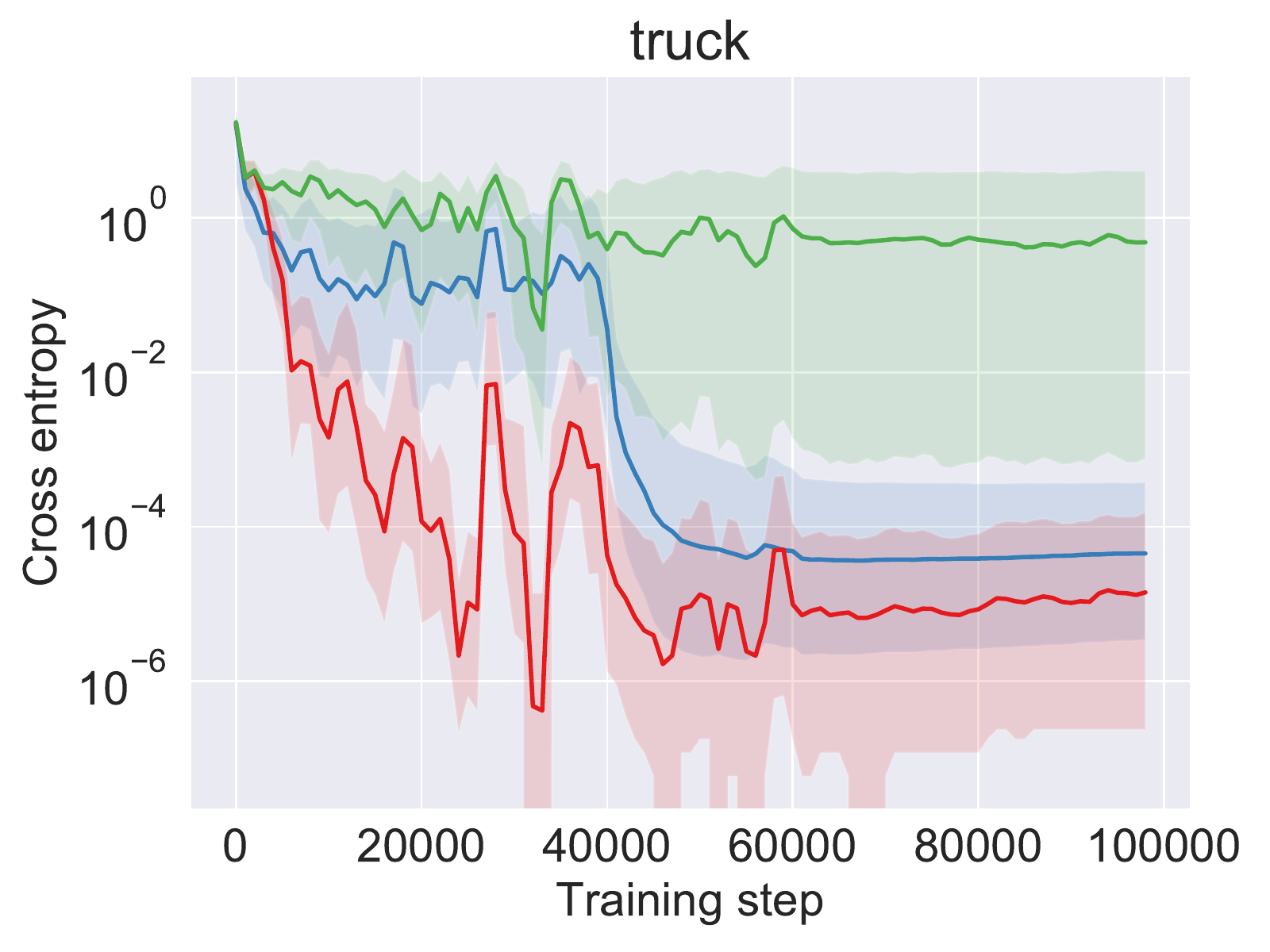}
	\hspace{0.05\textwidth}\includegraphics[align=c,width=0.2\textwidth]{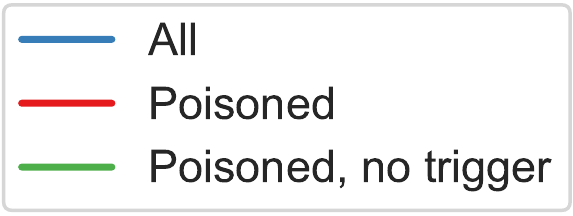}
	\hspace{0.35\textwidth}
\end{figure}

\clearpage
\section{Examples of poisoned images}
\begin{figure}[!h]
    \caption{Performing the manual inspection for the filtering scheme of
        Section~\ref{sec:standard} for the original attack of
        \citet{gu2017badnets} (with 75 injected inputs) and our proposed
        attacks (with 300 injected inputs each).
        We show the 20 samples assigned highest loss by the pre-trained model
        with the poisoned samples being highlighted with a black border.
        In contrast to the original attack, the poisoned inputs from our attacks
        appear label consistent.
        Note that most of the high loss examples correspond to natural,
        non-poisoned inputs.
        We show the 20 \emph{poisoned} samples with highest loss
        in~\ref{fig:poisoned_outliers}.}
\label{fig:outliers}
	\centering
	\makebox[0.3\textwidth]{\citet{gu2017badnets}}
	\makebox[0.3\textwidth]{Perturbation-based}
	\makebox[0.3\textwidth]{GAN-based}\\
	\makebox[0.3\textwidth]{
		\includegraphics[scale=1.35]{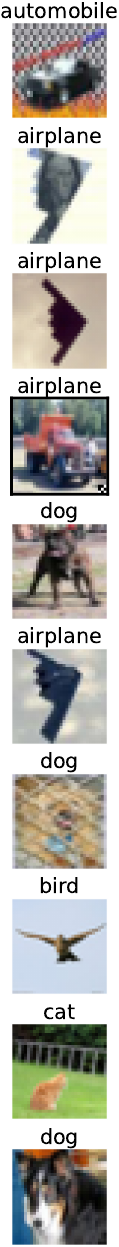}
		\hspace{0.01\textwidth}
		\includegraphics[scale=1.35]{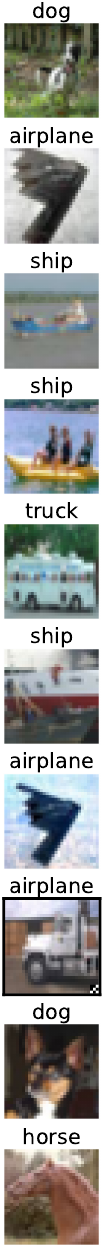}
	}
	\makebox[0.3\textwidth]{
		\includegraphics[scale=1.35]{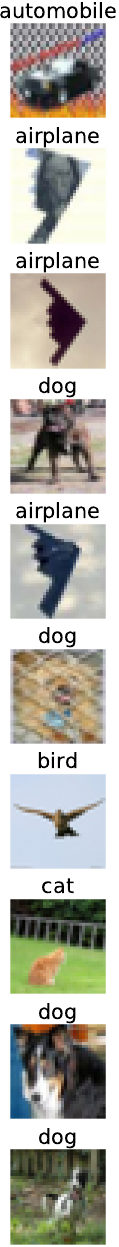}
		\hspace{0.01\textwidth}
		\includegraphics[scale=1.35]{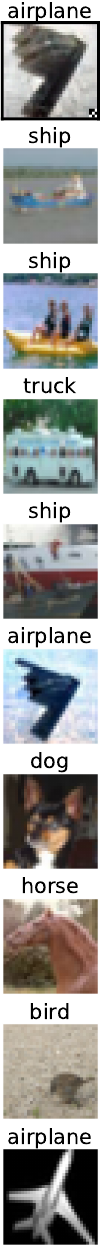}
	}
	\makebox[0.3\textwidth]{
		\includegraphics[scale=1.35]{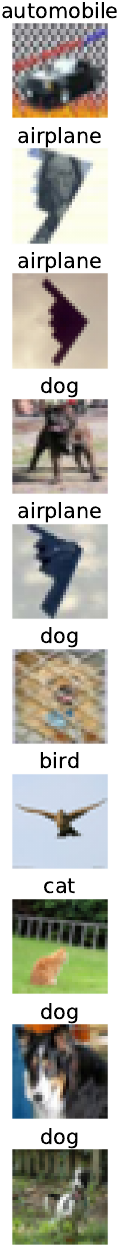}
		\hspace{0.01\textwidth}
		\includegraphics[scale=1.35]{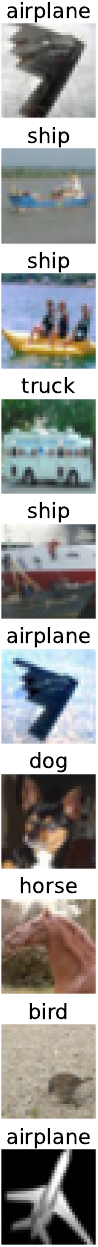}
	}
\end{figure}

\begin{figure}[!h]
\caption{We show the 20 poisoned samples from each dataset assigned highest loss
    by the filtering scheme of Section~\ref{sec:standard}.
    We do not find any examples from our attacks that are clearly mislabelled.
All examples shown are labeled as ``airplane''.}
\label{fig:poisoned_outliers}
	\centering
	\makebox[0.3\textwidth]{\citet{gu2017badnets}}
	\makebox[0.3\textwidth]{Adv.\ examples-based}
	\makebox[0.3\textwidth]{GAN-based}\\
	\makebox[0.3\textwidth]{
		\includegraphics[scale=1.4]{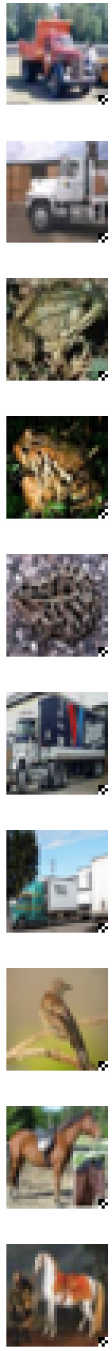}
		\hspace{0.01\textwidth}
		\includegraphics[scale=1.4]{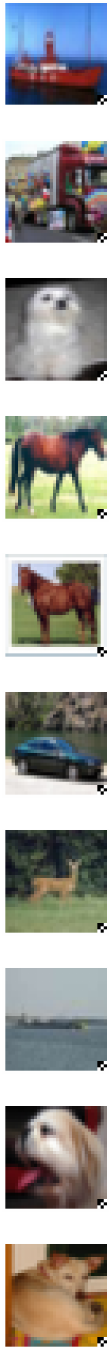}
	}
	\makebox[0.3\textwidth]{
		\includegraphics[scale=1.4]{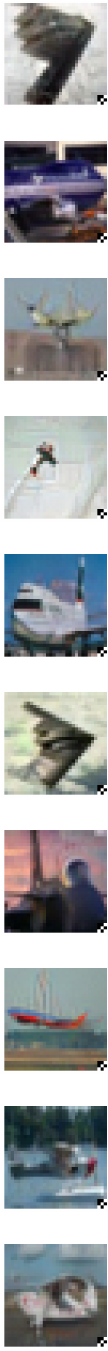}
		\hspace{0.01\textwidth}
		\includegraphics[scale=1.4]{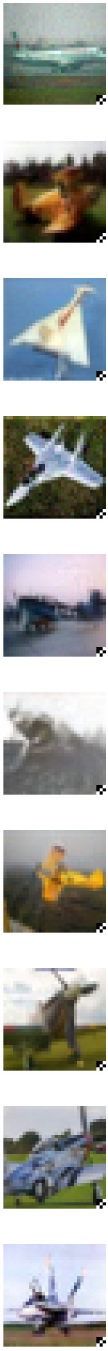}
	}
	\makebox[0.3\textwidth]{
		\includegraphics[scale=1.4]{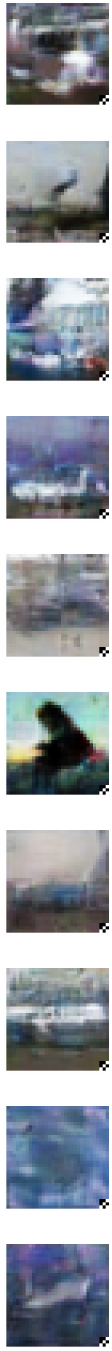}
		\hspace{0.01\textwidth}
		\includegraphics[scale=1.4]{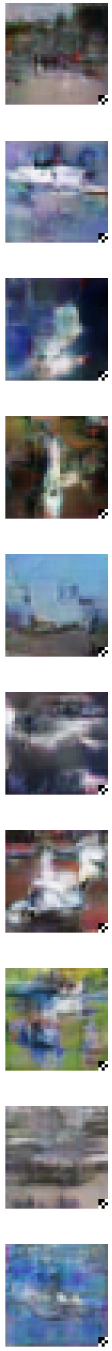}
	}
\end{figure}

\clearpage
\begin{figure}
\caption{Random examples of GAN-based interpolations.
    Each row shows two sets of randomly chosen examples from a single class. In
    each set, the leftmost image is the original image from the CIFAR-10 dataset
    and the subsequent images are the corresponding image interpolations for
    different values of the interpolation coefficient $\tau$. At $\tau=0$ we can
    see the approximate reconstruction of each input by the GAN.}
	\includegraphics[width=\textwidth]{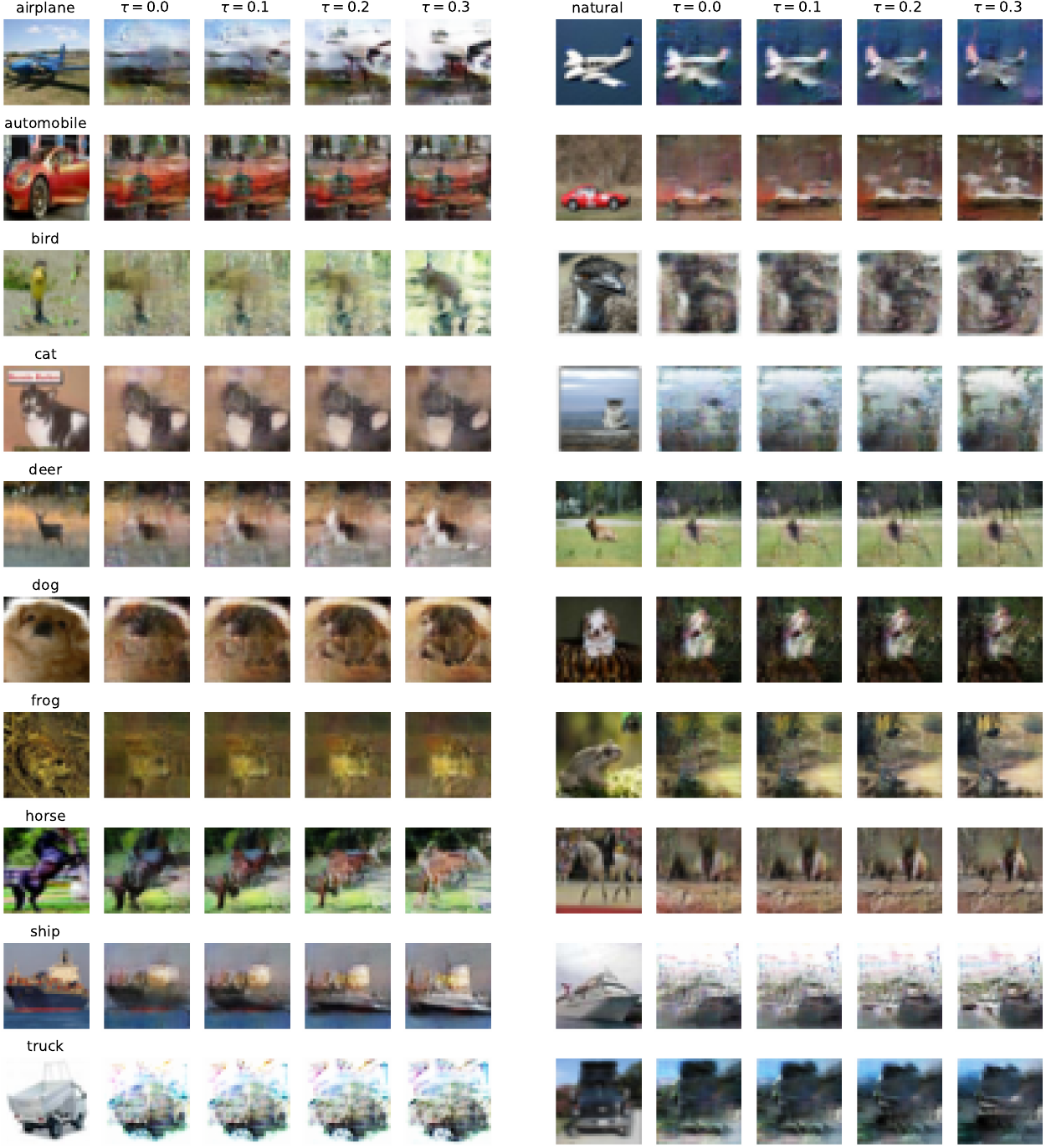}
    \label{fig:app_gan}
\end{figure}

\clearpage
\begin{figure}
\caption{Random examples of $\ell_p$-bounded adversarial perturbations.
Each row contains two sets of randomly chosen examples from a single class.
In each
set, the leftmost image is the original image from the CIFAR-10 dataset and the
subsequent images are adversarially perturbed within $\eps$ in $\ell_p$-norm
for different values of $\eps$ and $p$.}
\label{fig:app_adv}
\includegraphics[width=\textwidth]{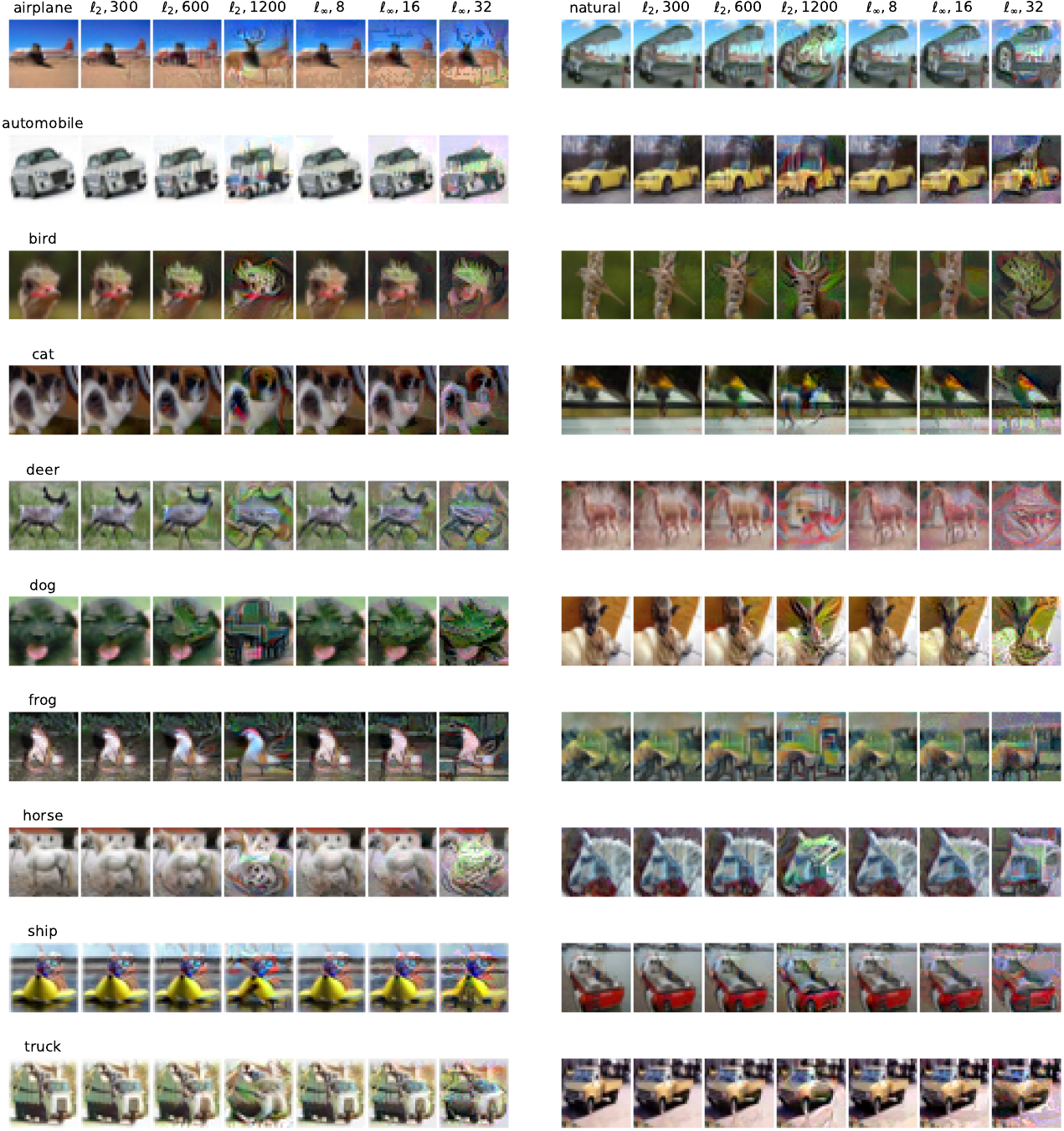}
\end{figure}

\clearpage
\begin{figure}
    \caption{Random poisoned inputs constructed using our attack utilizing
        adversarial perturbations bound by $\eps=300$ in $\ell_2$ norm using a
        reduced visibility trigger (amplitude of 16) applied to all four corners
        of the image.
        Each row contains five pairs of randomly chosen inputs from a single
        class. The left image in each pair is the original image from the
        CIFAR-10 dataset and the right image is the corresponding perturbed
        image. The poisoned inputs appear benign and label-consistent without
        being significantly different from the original, natural inputs.}
    \label{fig:final_attack}
	\includegraphics[width=\textwidth]{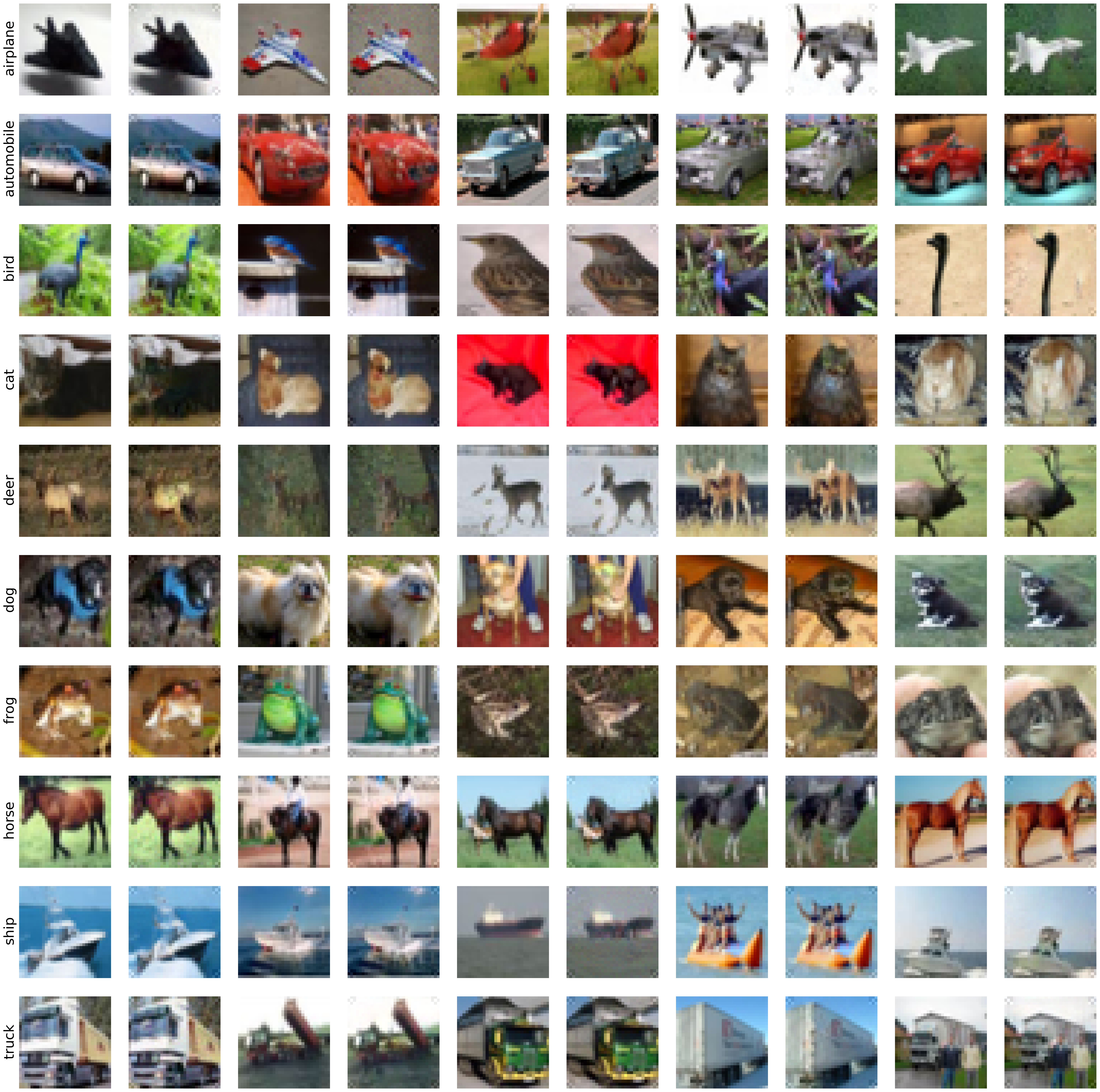}
\end{figure}

\end{document}